%% file: main.tex
\documentclass[preprint,12pt]{elsarticle}

\usepackage{color}
\usepackage{ mathrsfs }

\newcommand{\bb}{\color{black}}

\newcommand{\bl}{\color{black}}

\newcommand{\mycomment}[1]{}

\usepackage{graphicx} 
\usepackage{amssymb}
\usepackage{amsmath}
\usepackage{url}
\usepackage{booktabs}
\usepackage{multirow}
\usepackage{tikz}
\usetikzlibrary{positioning}

\usepackage{subcaption}
\usepackage{bbold}
\usepackage{longtable}
\usepackage{slashbox}
\usepackage{tabularx}
\usepackage{mathrsfs }

\newcommand{\emoji}[1]{\raisebox{-0.4em}{\includegraphics[height=1.5em,trim=2px 2px 2px 2px,clip]{#1}}}

\journal{Computer Methods and Programs in Biomedicine}

\begin{document}
\begin{frontmatter}


\title{Benchmarking Foundation Models and Parameter-Efficient Fine-Tuning for Prognosis Prediction in Medical Imaging}


\author[campus]{Filippo Ruffini} 
\ead{filippo.ruffini@unicampus.it}
\author[campus]{Elena Mulero Ayllon} 
\ead{e.muleroayllon@unicampus.it}
\author[shenzhen]{Linlin Shen}
\ead{llshen@szu.edu.cn}
\author[campus,umea]{Paolo Soda\corref{cor1}}
\ead{p.soda@unicampus.it, paolo.soda@umu.se}
\author[campus]{Valerio Guarrasi}
\ead{valerio.guarrasi@unicampus.it}
\cortext[cor1]{Corresponding author}

\affiliation[campus]{organization={Unit of Artificial Intelligence and Computer Systems, Department of Engineering,
Università Campus Bio-Medico di Roma},
            addressline={Via Álvaro del Portillo, 21}, 
            city={Rome},
            postcode={00128}, 
            country={Italy}}
            
\affiliation[shenzhen]{organization={College of Computer Science and Software Engineering, Shenzhen University},
            addressline={School of Computer and Software, Canghai Campus}, 
            city={Shenzhen},
            country={China}}

\affiliation[umea]{organization={Department of Diagnostics and Intervention, Radiation Physics, Biomedical Engineering, Umeå University},
            city={Umeå},
            postcode={901 87}, 
            country={Sweden}}


\begin{abstract}
\bl
\textbf{Background and Objectives}

Despite the significant potential of Foundation Models (FMs) in medical imaging, their application to prognosis prediction remains challenging due to data scarcity, class imbalance, and task complexity, limiting their clinical adoption.
This study introduces the first structured benchmark to assess the robustness and efficiency of transfer learning strategies for FMs compared with convolutional neural networks (CNNs) in predicting COVID-19 patient outcomes from chest X-rays.
The goal is to systematically compare fine-tuning strategies, classical and parameter-efficient, under realistic clinical constraints related to data scarcity and class imbalance, offering empirical guidance for AI deployment in clinical workflows.

\textbf{Methods} 
Four publicly available COVID-19 chest X-ray datasets were used, covering mortality, severity, and ICU admission, with varying sample sizes and class imbalances.
CNNs pretrained on ImageNet and FMs pretrained on general or biomedical datasets were adapted using full fine-tuning, linear probing, and parameter-efficient methods.
Models were evaluated under full-data and few-shot regimes using Matthews Correlation Coefficient (MCC) and Precision–Recall AUC (PR-AUC) with cross-validation and class-weighted losses.

\textbf{Results} CNNs with full fine-tuning performed robustly on small, imbalanced datasets, while FMs with Parameter-Efficient Fine-Tuning (PEFT), particularly LoRA and BitFit, achieved competitive results on larger datasets.
Severe class imbalance degraded PEFT performance, whereas balanced data mitigated this effect.
In few-shot settings, FMs showed limited generalization, with linear probing yielding the most stable results.
\textbf{Conclusions:} No single fine-tuning strategy proved universally optimal.
CNNs remain dependable for low-resource scenarios, whereas FMs benefit from parameter-efficient methods when data are sufficient.
\bb

\end{abstract}


\end{frontmatter}



\section{Introduction}\label{sec:Intro}
Artificial Intelligence (AI) has become a transformative wave in medical imaging, widely adopted across specialties such as radiology, oncology, and pathology~\cite{poalelungi2023advancing, rana2023machine}. Its effectiveness in disease detection, cancer subtype classification, and early screening~\cite{ibrahim2024pneumonia, aksu2024toward, Chowdhury2020CanAH} was 
exceptionally evident during the COVID-19 pandemic, when the development of AI-based diagnostic models for radiology were essential to address the crisis~\cite{danilov2022automatic, signoroni2021bs, rangarajan2021artificial}, underling the pivotal role of AI in managing large-scale health emergencies~\cite{rangarajan2021artificial}.
In a complementary direction, prognostic models have made significant advancements, incorporating sophisticated AI-based approaches to estimate and anticipate clinical outcomes, as the progression of pathology and mortality risk~\cite{Buttia2023PrognosticMI, MultiDataset_MICCAI2024}. 
However, prognosis tasks remain more complex and less explored than diagnostic tasks, primarily due to data-related challenges such as the inherent temporal structure of the data and the limited availability of high-quality annotations or adequately large cohorts to effectively harness the full potential of AI-driven models~\cite{adamidi2021artificialCOVID, gillman2021automated}.

Among these models, Convolutional Neural Networks (CNNs) have been central to medical imaging, particularly for diagnostic tasks. 
They demonstrate strong performance when first trained in a supervised manner on large-scale natural image datasets (e.g., ImageNet~\cite{ImageNet}) and subsequently fine-tuned on clinical data~\cite{ke2021chextransfer}.
Thanks to their inductive biases and computational efficiency, CNNs are particularly well-suited to medium-sized radiological datasets publicly available for diagnostic tasks. 
However, their effectiveness diminishes in prognostic settings, where datasets are typically smaller and the modeling of temporal dependencies is essential. 
The need for full fine-tuning, combined with their limited capacity to capture temporal structure, reduces their suitability for tasks involving complex prediction objectives or constrained sample availability~\cite{raghu2019transfusion_review, kora2022transfer}.

As an alternative to supervised pre-training, which requires huge labeling effort to acquire the datasets needed during training, Self-Supervised Learning (SSL) has emerged as a scalable paradigm that reduces or eliminates the need for manual annotations~\cite{bommasani2021opportunities, Azizi2021BigSM}. 
Models like DINO~\cite{caron2021emerging} or CLIP~\cite{radford2021learning-cliplarge} symbolize this shift, leveraging large amounts of unlabeled data to learn rich visual representations through contrastive or knowledge distillation objectives.
SSL-pretrained models demonstrated remarkable transferability to unseen downstream tasks and strong robustness in high-variance clinical settings~\cite{anton2022well, wolf2023self}. Building on these advances, FMs emerged, based on the adoption of large transformer-based architectures pretrained on massive unlabeled datasets, extending the principles of self-supervised learning to a broader and more generalist perspective of AI. 
In healthcare, FMs have been extensively developed to align with domain-specific diagnostic and classification tasks~\cite{zhang2024challenges_imaging, moor2023foundation, khan2025comprehensiveFM}.
However, the application of FMs to prognosis prediction remains largely underexplored, despite their presumed generalist adaptability and the expectation that they could transfer effectively across diverse clinical settings. 
The limited deployment of FMs in this domain is mainly due to the inherent challenges of adapting large-scale models to the noisy, heterogeneous, and temporally structured nature of prognostic data, conditions that exacerbate the weaknesses of conventional adaptation methods~\cite{li2024text, shi2023long}.

To address these issues, more effective adaptation strategies are required. Fine-tuning remains the most widely adopted approach; however, its performance is often constrained by domain shifts, limited supervision, and suboptimal representation learning in small target datasets. Despite these drawbacks, it continues to serve as a practical baseline due to its simplicity and demonstrated effectiveness across various clinical prediction tasks.
Yet, given the massive scale and complexity of FMs, the a growing need for more efficient and intelligent approaches to adapt them to new domains. 
Rather than updating all model parameters, PEFT methods modify only a small subset of task-specific components, such as bias terms~\cite{zaken2021bitfit}, low-rank adapters~\cite{hu2021lora}, or scaling vectors~\cite{liu2022-ia3}, while keeping the backbone frozen.
These approaches drastically reduces the computational and memory demands typically associated with full fine-tuning, mitigates overfitting, and enables more flexible adaptation across tasks with limited data availability.
Despite their growing success in diverse domains, the use of PEFT techniques in prognostic modeling remains remarkably underexplored, leaving a significant gap in understanding how such methods perform in temporally structured and highly heterogeneous clinical settings.

Moving in a similar direction, Few-Shot Learning (FSL) has emerged as a complementary paradigm particularly suited to prognosis prediction, where annotated cohorts are often small and imbalanced.
FSL methods aim to learn transferable representations from only a handful of labeled examples~\cite{pachetti2024systematic, woerner2024navigating}, leveraging prior knowledge from pretrained models through meta-learning or prototype-based strategies.
Their ability to generalize from minimal supervision is especially appealing for rare disease scenarios, where data scarcity is intrinsic.
In this context, the integration of PEFT within FSL frameworks represents a compelling avenue, combining the efficiency of parameter-sparse adaptation with the data economy of few-shot learning to advance real-world prognostic modeling~\cite{Bansal2022MetaAdaptersPE}.
\subsection{Contributions}
\bl
This work fills a critical gap in the application of FMs in prognosis prediction tasks, by introducing the first large-scale benchmark of fine-tuning, encompassing Full Fine-Tuning (FFT), Linear Probing (LP), and multiple Parameter-Efficient Fine-Tuning (PEFT) methods.
We systematically assess how dataset size, class imbalance, and model architecture influence adaptability, robustness, and efficiency across both CNNs and FMs.
Our benchmark evaluates fine-tuning methods on a diverse set of pretrained models, including general-purpose architectures (e.g., ResNet, CLIP) and biomedical-specific ones (e.g., MedCLIP, BioMedCLIP), covering both CNN and transformer-based FMs.
We focus on COVID-19 chest X-ray (CXR) prognosis as a clinically meaningful use case, leveraging public datasets with heterogeneous outcome types, sample sizes, and imbalance levels.
In detail, our main contributions can be summarized as follows:
\bl 
\begin{itemize}
\item \textbf{Comprehensive Benchmark of CNNs and FMs.}
We present the first systematic comparison of PEFT techniques against classical fine-tuning strategies (FFT and LP) across CNNs and FMs pretrained on both general and biomedical domains.
The benchmark identifies which combinations of model architecture and fine-tuning method remain most resilient to data scarcity and class imbalance, two central challenges in clinical prognosis modeling.

\item \textbf{PEFT for Prognostic Foundation Models.} 
While CNNs have been widely explored, the application of FMs to prognosis remains largely untested. 
We perform an extensive evaluation of PEFT, LP, and FFT strategies on FMs, quantifying the trade-off between predictive performance, computational efficiency, and model complexity, thereby establishing the first systematic assessment of PEFT effectiveness in medical prognosis tasks.

\item \textbf{Few-Shot Learning under Clinical Constraints.} 
Lastly, we simulate few-shot learning scenarios to evaluate the ability of PEFT-adapted Foundation Models to generalize effectively under extreme data scarcity.
This analysis mirrors real-world conditions such as rare disease modeling, emerging outbreak response, and underrepresented patient cohorts, demonstrating the practical feasibility of PEFT-based fine-tuning for rapid clinical deployment.

\end{itemize}
By addressing these questions, this study establishes a methodological foundation for adapting FMs to prognostic prediction, providing actionable insights for building robust, efficient, and generalizable AI systems in real-world medical environments.
Moreover, it contributes to the broader field of radiomics by advancing the integration of deep foundation models within quantitative imaging workflows, promoting a shift from handcrafted feature engineering toward data-driven, transferable representations for prognostic analysis.
The remainder of this work is organized as follows: Section~\ref{sec:relatedworks} reviews related efforts on FMs and PEFT in medical imaging, emphasizing the lack of systematic studies on prognosis.
Section~\ref{sec:Methods} describes the datasets, methodological framework, model selection criteria, fine-tuning strategies, and experimental setup.
Finally, Section~\ref{sec:RES} presents the results and discussion, structured around the three central research questions introduced above.

\section{Related Works}\label{sec:relatedworks}

To contextualize this benchmark, we review prior research on (i) the use of FMs in medical imaging and (ii) PEFT as an emerging strategy for adapting large pretrained models to medical tasks.
Despite rapid progress in both fields, structured analyses of fine-tuning and PEFT for prognostic prediction remain largely absent, particularly under clinically realistic constraints such as limited data and severe class imbalance.

\subsection{Foundation Models in Medical Imaging}

FMs have redefined medical imaging by transferring large-scale SSL paradigms into clinical domains~\cite{khan2025comprehensiveFM}.
Trained on massive unlabeled datasets, these models enable general-purpose representation learning and have led to biomedical adaptations such as MedCLIP~\cite{wang2022medclip}, BioMedCLIP~\cite{zhang2023biomedclip}, and CheXagent~\cite{chen2024vision}, developed for radiological interpretation and multimodal reasoning.
In segmentation and 3D imaging, extensions like MedSAM~\cite{ma2024segmentMEDSAM} and time-to-event pretraining~\cite{huo2024time} demonstrate the versatility of large vision-language architectures in capturing anatomical or temporal features.

However, most works focus on diagnostic and segmentation tasks, emphasizing classification or localization rather than patient-level outcome prediction. Prognostic modeling introduces additional challenges, small sample sizes, class imbalance, and temporal dependency structures, that remain poorly addressed by current FM adaptation strategies~\cite{guarrasi2025systematic}.
While isolated attempts, such as BEPH~\cite{yang2025foundation}, explore survival prediction, no large-scale benchmark currently evaluates FMs or PEFT approaches in prognosis-oriented tasks.

\subsection{Parameter-Efficient Fine-Tuning in Medical Imaging}

PEFT has emerged as a lightweight alternative to full fine-tuning, allowing large pretrained models to adapt to downstream tasks by updating only a small subset of parameters~\cite{han2024parameter, dutt2023parameter}.
Approaches such as BitFit~\cite{zaken2021bitfit}, LoRA~\cite{hu2021lora}, and VeRA~\cite{kopiczko2023vera} have proven effective in reducing computational cost while maintaining strong predictive performance.
Medical applications have largely focused on diagnostic or segmentation tasks, as demonstrated by FairMedFM~\cite{jin2024fairmedfm}, FairTune~\cite{dutt2023fairtune}, and Med-SA~\cite{wu2023medical}, which show that PEFT can improve model fairness and generalization under low-resource conditions.

In parallel, FSL has emerged as a complementary paradigm for extreme data scarcity~\cite{pachetti2024systematic, woerner2024navigating}. Integrating PEFT into FSL frameworks further enhances efficiency and transferability, as demonstrated in MetaAdapters~\cite{Bansal2022MetaAdaptersPE}.
\bl
\subsection{Limitations of Related Works}

While prior research has established the utility of FMs and PEFT in diagnostic and segmentation applications, several key limitations persist:
\begin{enumerate}
  \item Lack of structured benchmarks assessing fine-tuning and PEFT performance under prognostic conditions;
  \item Limited focus on data imbalance and temporal dynamics, which are intrinsic to clinical outcome prediction;
  \item Insufficient evaluation of biomedical-specific versus general-purpose, leaving unclear which pretraining strategies best generalize to prognosis;
  \item Absence of unified methodological frameworks that compare full fine-tuning, linear probing, and PEFT under controlled, clinically realistic settings.
\end{enumerate}
Our work directly addresses these gaps by introducing the first systematic benchmark of fine-tuning strategies across CNNs and FMs for prognosis prediction, explicitly designed to analyze the interplay between dataset size, class balance, and model architecture.
This framework provides a standardized foundation for evaluating FM adaptability in data-limited and high-impact medical applications.
\bb

\section{Materials \& Methods}\label{sec:Methods}

\begin{figure}[t!]
 \centering \includegraphics[width=1\linewidth,trim=3 3 3 3,clip]{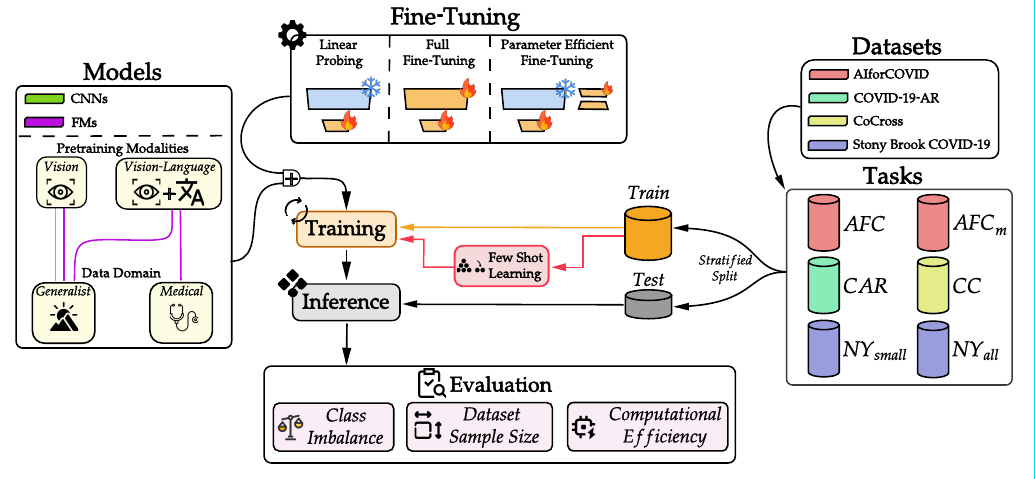}
 \caption{Overview of the methodological framework. The pipeline is structured into five main stages: dataset, models, fine-tuning, training (w/ or w/o FSL) and inference/evaluation. }
 \label{fig:method-overview}
\end{figure}

Our experimental pipeline, illustrated in Fig.~\ref{fig:method-overview}, defines the key methodological framework adopted in this benchmark. 
It is structured into five main stages, ensuring a systematic and reproducible evaluation of fine-tuning strategies and model transferability for clinical prognosis prediction using medical imaging data. 
Each stage has been carefully designed to simulate real-world clinical conditions, balancing methodological rigor with practical applicability.

\begin{description}

\item[\textbf{Dataset Selection.}]
We employ four publicly available COVID-19 CXR datasets annotated with a variety of prognostic outcomes, including mortality, severity, and ICU admission. These datasets cover a wide range of clinical scenarios and data challenges, including varying degrees of class imbalance, outcome heterogeneity, and differences in sample size, ranging from highly underrepresented cases to larger, more balanced datasets. This diversity reflects the complexities typical in prognosis prediction and enables a robust evaluation of model adaptability under different data regimes~\cite{mamandipoor2022covid19, jiao2021prognosticationCOVID19}. Further details of the datasets are provided in Section~\ref{section:DATA}.

\item[\textbf{Model Categorization.}]
We systematically evaluate models grouped into two main architectural families. The first group composed of three CNNs pretrained on ImageNet using supervised learning, as a strong baselines in medical imaging. 
The second group composed entirely by FMs, pretrained through self-supervised or contrastive learning, either on large-scale general image-text datasets or specialized medical datasets. These two group allow us to explore the interplay between architectural design, pretraining paradigms, and domain specificity in prognosis prediction tasks. Further details of the models are provided in Section~\ref{sec:MODELS}.

\item[\textbf{Fine-Tuning Strategies.}]
Each model undergoes adaptation to the target prognostic tasks using one of several fine-tuning strategies. 
These include FFT as a performance upper bound; LP to evaluate the quality of frozen representations; and multiple PEFT techniques, which are designed to minimize computational overhead and improve adaptation efficiency. 
The full range of supported configurations and their applicability to specific architectures are detailed in Section~\ref{sec:Methods}.

\item[\textbf{Training Regimes.}]
To assess the generalization capabilities of models under realistic constraints, we benchmark each configuration under two training regimes. 
In the full-data regime, where models are trained using all available labeled samples. And the FSL regime, where we limit the training to few samples per class, simulating extreme data scarcity common in rare disease contexts or emerging health crises. 
This dual evaluation framework enables a comprehensive analysis of model robustness and the effectiveness of fine-tuning strategies across a spectrum of data availability scenarios.

\item[\textbf{Inference \& Evaluation.}]
We evaluate model performance primarily using the Matthews Correlation Coefficient (MCC), a metric known for its robustness in the presence of class imbalance, which is a prevalent issue in prognostic datasets. This ensures a fair assessment of model predictions across both majority and minority classes. An Additional evaluation metric, Area Under the Precision-Recall Curve (AUC-PR), is employed to provide a broader understanding of the models' behavior.

\end{description}
This framework establishes a rigorous foundation for the analyses presented in the following sections, where we discuss in detail the datasets employed, the models and fine-tuning strategies evaluated, and the experimental design used to systematically address the research questions of this study.

\subsection{Datasets}\label{section:DATA}

\begin{table}[t!]

\begin{center}
\resizebox{\columnwidth}{!}{
\begin{tabular}{l|p{2.5cm}|p{2.5cm}|p{2cm}|p{2cm}|p{2cm}}
\toprule
\textbf{Dataset} 
& \textbf{Task} 
& \textbf{Labels } 
& \textbf{ $\textbf{\#N}$}
& \textbf{ $\textbf{\#P}$} 
& \textbf{ {$\textbf{\#C}$}} \\
\midrule

\multirow{4}{*}{\textbf{AIforCOVID}~\cite{soda2021aiforcovid-afc}} &
Severity Outcome ($\mathbf{AFC}$)& Mild: 839 \newline Severe: 746 & 1585 & 1585 & 6  \\ 
\cmidrule(lr){2-6}
& Mortality ($\mathbf{AFC_{m}}$) & Alive: 1336 \newline Deceased: 249 & 1585 & 1585 & 6 \\ 
\midrule

\multirow{2}{*}{\textbf{COVID-19-AR}~\cite{desai2020chest-car}} &
ICU Admission ($\mathbf{CAR}$) & Yes: 28 \newline No: 71 & 99 & 99 & 1 \\ 
\midrule

\multirow{2}{*}{\textbf{CoCross}~\cite{kilintzis2022cocross-cc}} &
ICU Outcome ($\mathbf{CC}$) & Alive: 243 \newline Deceased: 146 & 389 & 150 & 1 \\ 
\midrule

\multirow{4}{*}{\textbf{Stony Brook COVID-19}~\cite{saltz2021stony-ny}} &
Mortality ($\mathbf{NY_{small}}$) & Alive: 1183 \newline Deceased: 182 & 1365 & 1365 & 1 \\ 
\cmidrule(lr){2-6}
&

Mortality ($\mathbf{NY_{all}}$) & Alive: 8742 \newline Deceased: 4897 & 13639 & 1365 & 1 \\ 
\bottomrule

\end{tabular}
}

\end{center}
\vskip -0.1in

\caption{Detailed overview for the COVID-19 CXR datasets used in this benchmark, among with each prognosis prediction task. \textbf{$\textbf{\#N}$}: total number of samples, \textbf{$\textbf{\#P}$}: patients, \textbf{$\textbf{\#C}$}: medical centers}
\label{tab:dataset-table}
\end{table}

In this study, we evaluate four publicly available COVID-19 chest X-ray (CXR) datasets, chosen for their relevance to prognosis prediction and their diversity in task definition, class distribution, and sample size.
We focus on COVID-19 due to its representativeness and the presence of standardized care protocols established during the pandemic, which ensure consistent outcome definitions across datasets and enable reliable cross-cohort comparisons~\cite{mamandipoor2022covid19,jiao2021prognosticationCOVID19}.
This design allows us to benchmark the selected models within a unified disease framework while still reflecting real-world challenges such as data scarcity, heterogeneous imaging frequency, and variability in clinical outcomes.

Each dataset is annotated with different task, ranging from mortality or severity, to ICU admission, and collectively spanning from balanced to highly imbalanced class distributions and small to large sample sizes. 
Table~\ref{tab:dataset-table} reports the main statistics, including image count, patient numbers, and class distributions of the selected datasets and related tasks.

The AIforCOVID dataset~\cite{soda2021aiforcovid-afc} includes 1,585 patients collected from six Italian hospitals and defines two distinct prognostic tasks.
The first concerns disease severity classification, Mild vs. Severe (\textbf{AFC}), featuring a balanced class distribution ($53–47\%$), while the second focuses on mortality prediction (\textbf{AFC\textsubscript{m}}), characterized by a pronounced class imbalance ($85–15\%$).
Both tasks are derived from the same patient cohort and share identical CXRs examinations.

The CoCross dataset~\cite{kilintzis2022cocross-cc} comprises 387 CXRs from ICU patients monitored longitudinally, annotated with a binary ICU outcome label (\textbf{CC}) that exhibits moderate class imbalance ($63–37\%$).
The COVID-19-AR dataset~\cite{desai2020chest-car}, although limited to 99 CXRs, specifically targets rural and underrepresented populations. Its ICU admission label (\textbf{CAR}) is strongly imbalanced (71–29\%), providing a valuable testbed for assessing model robustness under extreme low-data and class-imbalance conditions.

Lastly, the Stony Brook COVID-19 dataset~\cite{saltz2021stony-ny} is employed for mortality prediction and includes two complementary versions: $\mathbf{NY_{small}}$, consisting of one CXR per patient taken closest to hospital admission (1,365 CXRs; $87–13\%$ class imbalance), and $\mathbf{NY_{all}}$, comprising all CXRs acquired throughout hospitalization (13,639 images; $64–36\%$ class imbalance). 
Both versions share the same patient cohort, enabling controlled investigations of how dataset size and class imbalance affect prognostic model performance.

Collectively, these datasets offer a comprehensive testbed for prognostic evaluation, encompassing key characteristics typical of real-world prognosis tasks, namely, high task complexity, variability in sample size, and differing degrees of class imbalance. The influence of these factors on model behavior will be systematically examined and discussed in relation to the performance of the evaluated Foundation Models across all tasks, with detailed results reported in Section~\ref{sec:RES}.

\subsection{Models}\label{sec:MODELS}
For this study, we evaluate a diverse set of models pretrained under three distinct paradigms: \textit{supervised}, \textit{self-supervised}, and \textit{contrastive language-image} learning.
This setup enables a controlled comparison of how different architectural designs and pretraining strategies influence model transferability to prognostic tasks.
A comprehensive summary of the evaluated models is provided in Table~\ref{tab:model_summary}, detailing their architectural class, pretraining strategy, pretraining dataset, vision backbone, number of vision encoder parameters, and embedding dimension.
The symbol $-$ in the embedding dimension column indicates models that are either non Transformer-based or do not possess a uniform embedding size.

\begin{table}[t!]
\centering
\resizebox{\columnwidth}{!}{
\begin{tabular}{l|c|c|c|c|c|c}
\toprule
\textbf{Model} & \textbf{Pretraining}  & \textbf{Data} & \textbf{Vision Backbone} & \textbf{\#Params (M)} & \textbf{Embed Dim} & \textbf{Category} \\
\midrule
ResNet18~\cite{he2016deep-resnet}       & Supervised & ImageNet~\cite{ImageNet}        & ResNet18     & 11.7   &   -   & \textbf{CNNs} \\
ResNet50~\cite{he2016deep-resnet}       & Supervised & ImageNet~\cite{ImageNet}        & ResNet50   & 23.5   & -     & \textbf{CNNs} \\
DenseNet121~\cite{huang2017densely-densenet}    & Supervised & ImageNet~\cite{ImageNet}        & DenseNet121    & 7.9    & -     & \textbf{CNNs} \\
\midrule
DINOv2$_s$~\cite{oquab2023dinov2}     & Self-supervised & LVD-$142$M~\cite{oquab2023dinov2}  & ViT-S/14         & 21     & 384   & \textbf{FMs}  \\
DINOv2$_b$~\cite{oquab2023dinov2}     & Self-supervised & LVD-$142$M~\cite{oquab2023dinov2}   & ViT-B/14        & 86    & 512  & \textbf{FMs} \\
DINOv2$_l$~\cite{oquab2023dinov2}     & Self-supervised & LVD-$142$M~\cite{oquab2023dinov2}    & ViT-L/14       & 300   & 768  & \textbf{FMs} \\
\midrule
CLIP-Large~\cite{radford2021learning-cliplarge}     & Contrastive & LAION-400M    & ViT-L/14    & 300    & 768   & \textbf{FMs} \\
MedCLIP$_c$~\cite{wang2022medclip}    & Contrastive & CheXpert$^{\dagger}$ + MIMIC$^{\dagger}$~\cite{johnson2023mimic} & ResNet50    & 23   & -     & \textbf{FMs} \\
MedCLIP$_v$~\cite{wang2022medclip}    & Contrastive & CheXpert$^{\dagger}$ + MIMIC$^{\dagger}$~\cite{johnson2023mimic} & Swin-T      & 27 & -   & \textbf{FMs} \\
PubMedCLIP~\cite{eslami-etal-2023-pubmedclip}     & Contrastive & ROCO$^{\dagger}$ ~\cite{ruckert2024rocov2}         & ViT-B/14       & 86     & 512   & \textbf{FMs} \\
BioMedCLIP~\cite{zhang2023biomedclip}  & Contrastive & PMC-15M$^{\dagger}$~\cite{zhang2023biomedclip}        & ViT-B/14       & 86     & 512   & \textbf{FMs} \\
\bottomrule 
\end{tabular}}
\caption{Summary of the selected models, detailing their architectural category, pretraining strategy, pretraining dataset (with $\dagger$ indicating medical data), vision backbone type, number of vision encoder parameters (in millions), and embedding dimension.}
\label{tab:model_summary}
\end{table}
We include both \textbf{ResNet18} and \textbf{ResNet50} as representative CNN architectures, given their well-established success in transfer learning for medical imaging, particularly when pretrained on large-scale natural image datasets such as ImageNet~\cite{ke2021chextransfer, raghu2019transfusion_review}.
Their relatively low parameter count, strong inductive biases, and efficient training dynamics make them especially suitable for low-data scenarios, where overfitting and computational constraints are major challenges.

For the self-supervised learning (SSL) paradigm, we adopt \textbf{DINOv2}~\cite{oquab2023dinov2}, a state-of-the-art Vision Transformer (ViT) model that has demonstrated competitive performance in medical imaging applications~\cite{huang2024comparative, jin2024fairmedfm}.
Its capacity to learn rich and transferable visual representations from unlabeled data makes it particularly valuable for prognostic tasks in resource-limited settings, where annotated data are often scarce.
Lastly, we include \textbf{CLIP-Large}~\cite{radford2021learning-cliplarge} to evaluate the generalization potential of large-scale contrastive language–image pretraining for prognosis prediction.
For the contrastive pretraining, we selected also a set of models pretrained on large-scale multimodal biomedical datasets: \textbf{MedCLIP}~\cite{wang2022medclip}, \textbf{PubMedCLIP}~\cite{eslami-etal-2023-pubmedclip}, and \textbf{BioMedCLIP}~\cite{zhang2023biomedclip}.

For the multimodal architectures (i.e., MedCLIP), we extracted the vision encoder from the whole architecture, and later adding a task-specific linear adapter layer designed for each prognostic prediction task.
This diverse selection, spanning architectural types, pretraining strategies, and domain alignments, enables a systematic evaluation of adaptation strategies and model robustness in extreme low-data prognostic settings.

\subsection{PEFT, Linear Probing and Full Fine-Tuning}\label{sec:FT}

\begin{table}[ht!]
\centering
\renewcommand{\arraystretch}{0.85} 
\resizebox{\columnwidth}{!}{
\begin{tabular}{c|c|c|c|c|c|c|c}
\toprule
\textbf{Family} &\backslashbox{\textbf{Model}}{\textbf{FT}} & \textbf{LoRA} & \textbf{VeRA} & \textbf{\textit{IA}$^{\mathbf{3}}$} & \textbf{BitFit} & \textbf{LP} & \textbf{FFT} \\
\midrule
\multirow{3}{*}{\textbf{CNNs}} 
& ResNet18 & \checkmark & & & \checkmark & \checkmark & \checkmark \\
& ResNet50 & \checkmark & & & \checkmark & \checkmark & \checkmark \\
& DenseNet121 & \checkmark & & & \checkmark & \checkmark & \checkmark \\

\midrule
\multirow{6}{*}{\textbf{FMs}} 
& DINOv2$_{s,b,l}$ & \checkmark & \checkmark & \checkmark & \checkmark & \checkmark & \checkmark \\
& CLIP-Large & \checkmark & \checkmark & \checkmark & \checkmark & \checkmark & \checkmark \\
& MedCLIP$_{c}$ & \checkmark & & & \checkmark & \checkmark & \checkmark \\
& MedCLIP$_{v}$ & \checkmark & & \checkmark & \checkmark & \checkmark & \checkmark \\
& PubMedCLIP & \checkmark & \checkmark & \checkmark & \checkmark & \checkmark & \checkmark \\
& BioMedCLIP & \checkmark & \checkmark & \checkmark & \checkmark & \checkmark & \checkmark \\

\bottomrule
\end{tabular}
}
\caption{Overview of model architectures and their compatible PEFT strategies. \checkmark indicates the PEFT method is applicable to the corresponding model.}
\label{tab:peft-table}
\end{table}

In the context of medical imaging prognosis, adapting deep learning models to clinical outcome prediction remains a challenging task that requires balancing computational efficiency with effective knowledge transfer. 
Given the characteristics of our datasets, we evaluate several PEFT techniques against standard LP and FFT, across models pretrained with diverse learning paradigms and datasets (Section~\ref{sec:MODELS}). Table~\ref{tab:peft-table} summarizes the compatibility of each PEFT method with the considered architectures. Detailed mathematical formulations of all methods are provided in Appendix~\ref{app:math_peft}.

\textbf{Low-Rank Adaptation} (\textbf{LoRA})~\cite{hu2021lora} introduces trainable low-rank matrices while keeping the pretrained weights frozen. By decomposing weight updates into two smaller matrices, it significantly reduces the number of trainable parameters while preserving the performance of the original model. This is particularly beneficial in prognosis tasks, where data scarcity makes overfitting a risk. Although commonly applied to transformers~\cite{he2021towards, han2024parameter}, we also extend its evaluation to CNNs to assess its versatility across architectural types.

\textbf{Vector-based Random Matrix Adaptation} (\textbf{VeRA})~\cite{kopiczko2023vera} builds upon LoRA by employing fixed random projections shared across layers, modulated by trainable scaling vectors. This further reduces the parameter footprint while retaining competitive performance. Due to its reliance on a uniform feature space, VeRA is only applied to ViT-based models, excluding CNNs and the hierarchical Swin-based MedCLIP$_v$.

\textbf{BitFit}~\cite{zaken2021bitfit} is a sparse fine-tuning method that updates only the bias terms of the model. Despite modifying fewer than $0.1\%$ of the parameters, it often yields strong results, making it particularly attractive for prognosis applications where training resources and data availability are limited. Moreover, it offers insights into how far inductive biases encoded in pretrained models can be selectively activated through minimal adaptation.

\textbf{Infused Adapter by Inhibiting and Amplifying Inner Activations} (\textbf{IA}$^{3}$)~\cite{liu2022-ia3} adapts a pretrained model by inserting multiplicative gates that modulate intermediate activations. These scaling vectors act on attention and feed-forward layers without altering the model’s main parameters. Initialized to preserve the model’s original behavior, IA$^{3}$ enables stable and efficient adaptation, especially effective in low-data scenarios typical of prognosis tasks.

In addition to PEFT techniques, we include \textbf{FFT} and \textbf{LP} as reference fine-tuning settings. FFT updates all parameters during training, and serves as the upper-bound useful to evaluate how full model adaptability perform to prognosis downstream tasks, while LP freezes the backbone and optimizes only a linear classification head, offering a lower-bound that evaluates the quality of pretrained features~\cite{radford2021learning-cliplarge, he2020momentum}. 
Using both benchmarks establishes a performance spectrum, within which we assess the efficiency and expressiveness of PEFT strategies in real-world clinical settings.

\subsection{Few-Shot Learning Analysis}\label{sec:FSL}

FSL refers the problem setting in which a model must learn to generalize from only a few labeled examples per class \cite{pachetti2024systematic}. This paradigm is particularly relevant in medical imaging, where annotated data is often scarce due to high labeling costs and limited expert availability.
To evaluate the robustness and adaptability of PEFT methods under such constraints, we design a systematic few-shot learning benchmark. Let $ \mathcal{Y} = \{1, \dots, C\} $ denote the set of class labels. For a fixed number of shots $ k $, the few-shot training set is defined as follows:

\begin{equation}
\mathcal{D}_{\text{train}}^{(k)} = \left\{ (x_i, y_i) \ \middle|\ y_i \in \mathcal{Y},\ 1 \leq i \leq kC \right\}
\label{eq:FSL}
\end{equation}
where exactly $ k \in \mathbb{N}$, represents the samples drawn for each class, leading to a total of $ k C $ training examples per experiment. We evaluate all PEFT strategies listed in Table~\ref{tab:peft-table}, excluding FFT due to its tendency to overfit in low-data regimes. The training protocol follows a geometric progression over the number of shots: $ k \in \{2, 4, 8, 16, 32\} $. This allows us to systematically characterize the scaling behavior of each method as the number of labeled samples increases. 
Importantly, while training sets are artificially balanced, test sets preserve the natural class imbalance of each dataset. This imbalance poses a realistic challenge: models learn under balanced supervision but must generalize to clinically imbalanced conditions. 
By exploring this setup, we aim to determine the extent to which PEFT approaches can effectively adapt to prognosis prediction tasks with extremely limited data.

\subsection{Experimental Setup}\label{section:exp_stp}

In this section, we detail the experimental setup used to evaluate all models, covering training configurations, loss functions, evaluation metrics, and architectural considerations. 

All experiments followed a uniform configuration for the fine-tuning. Optimization was performed using Stochastic Gradient Descent (SGD) with a momentum of $ 0.9 $, weight decay of $ 10^{-4} $, and an initial learning rate of $ 10^{-3} $. Training proceeded for a maximum of 1000 epochs, with a 50-epoch warm-up phase, and employed early stopping with a patience of 50 epochs to mitigate overfitting. We used a batch size of 128 for larger tasks (AFC, AFC\textsubscript{m}, NY\textsubscript{small}, and NY\textsubscript{all}), and 32 for smaller datasets (CAR, CC) to match their lower dimensionality. 
\bl
To ensure strict experimental determinism and reproducibility, random seeds were fixed for all stages of experimentation, including data splitting, model initialization, and data loading/batching.
\bb

For model validation, we opted for a 5-fold cross-validation for all datasets, with the exception of the multi-centric AFC and AFC\textsubscript{m} datasets. In these cases, we employed Leave-One-Center-Out (LOCO) validation, which better reflects real-world deployment scenarios by accounting for inter-center variability, a critical factor in multi-institutional medical imaging studies. 
For the low-rank adaptation methods (LoRA and VeRA), we systematically varied the rank parameter $ r \in \{4, 8, 16\} $ to balance efficiency and expressiveness. 
Lower ranks reduce the number of trainable parameters, while higher ranks allow more task-specific adaptation. 

This variation helps assess the sensitivity of each method respect dimensionality of the adaptation space and how it is impacting the adaptation on the downstream tasks performance.

All models employed in this study were equipped with a task-specific classification head, appended to the output of the vision backbone. Feature representations were extracted from the final block of each model and passed through a lightweight linear layer to produce the final logits. In the case of Vision Transformer (ViT) and foundation model (FM) architectures, we consistently used the class token as the global feature descriptor. This design choice ensures structural and dimensional homogeneity across all architectures, facilitating fair comparisons across PEFT strategies and model types.

Due to the high class imbalance and limited data, a class-weighted Binary Cross-Entropy loss was computed on top of each classification head to ensure robustness across all the tasks. The loss is defined as:

\begin{equation}
\mathcal{L}_{BCE} = -\frac{1}{N}\sum_{i=1}^{N} w_i \left[ y_i \log(\hat{y_i}) + (1 - y_i) \log(1 - \hat{y_i}) \right],
\label{eq:WBCE}
\end{equation}

where $ w_i $ denotes the inverse class-frequency weight for sample $ i $, and $ \hat{y_i} $ is the predicted probability. The loss function in Eq.~\ref{eq:WBCE} was applied uniformly across all experimental settings, including every architectural variant, fine-tuning strategy, and FSL scenarios. This ensures consistency and comparability in training dynamics and evaluation.

Regarding evaluation metrics, we adopted the Matthews Correlation Coefficient (MCC) as our primary metric due to its robustness in extreme class imbalance conditions~\cite{chicco2023matthews}. 
The metric ranges from $\text{-}1 $to $1$, where $1$ indicates perfect prediction, $0$ corresponds to random guessing, and $\text{-}1$ denotes total disagreement between predictions and ground truth. 
Indeed, it considers true and false positives and negatives simultaneously, being well-suited for prognosis tasks where the minority class is typically of greater clinical interest yet severely underrepresented in the data (e.g., progression or adverse outcome).
To complement MCC, we also report the Precision-Recall Area Under the Curve (PR-AUC), which is more sensitive than ROC-AUC to the performance on the positive (minority) class and is thus better aligned with the requirements of our benchmark's tasks \cite{PRAUC-1}. 

\bl
Experiments were conducted on a high-performance computing cluster equipped with 10 NVIDIA TESLA A40 GPUs (48GB each). The total compute required for the full benchmark was approximately 3,200 GPU-hours, corresponding to an approximated average of 6 GPU-hours per model-dataset combination. 
For all the models presented in this benchmark, we used publicly released pretrained weights available on their official repositories or \textit{Hugging Face} hub. The corresponding links to the official repositories are reported in the appendix Table~\ref{tab:foundation_models_links}.
The complete code-base is publicly available at \url{https://github.com/fruffini/PEFT_Prognosis}.
\bb

\section{Results and Discussion}\label{sec:RES}
\begin{figure}[!t]
  \centering
  \resizebox{\columnwidth}{!}{
  \includegraphics[width=0.95\textwidth, trim=2.0cm 2.4cm 1cm 1cm, clip]{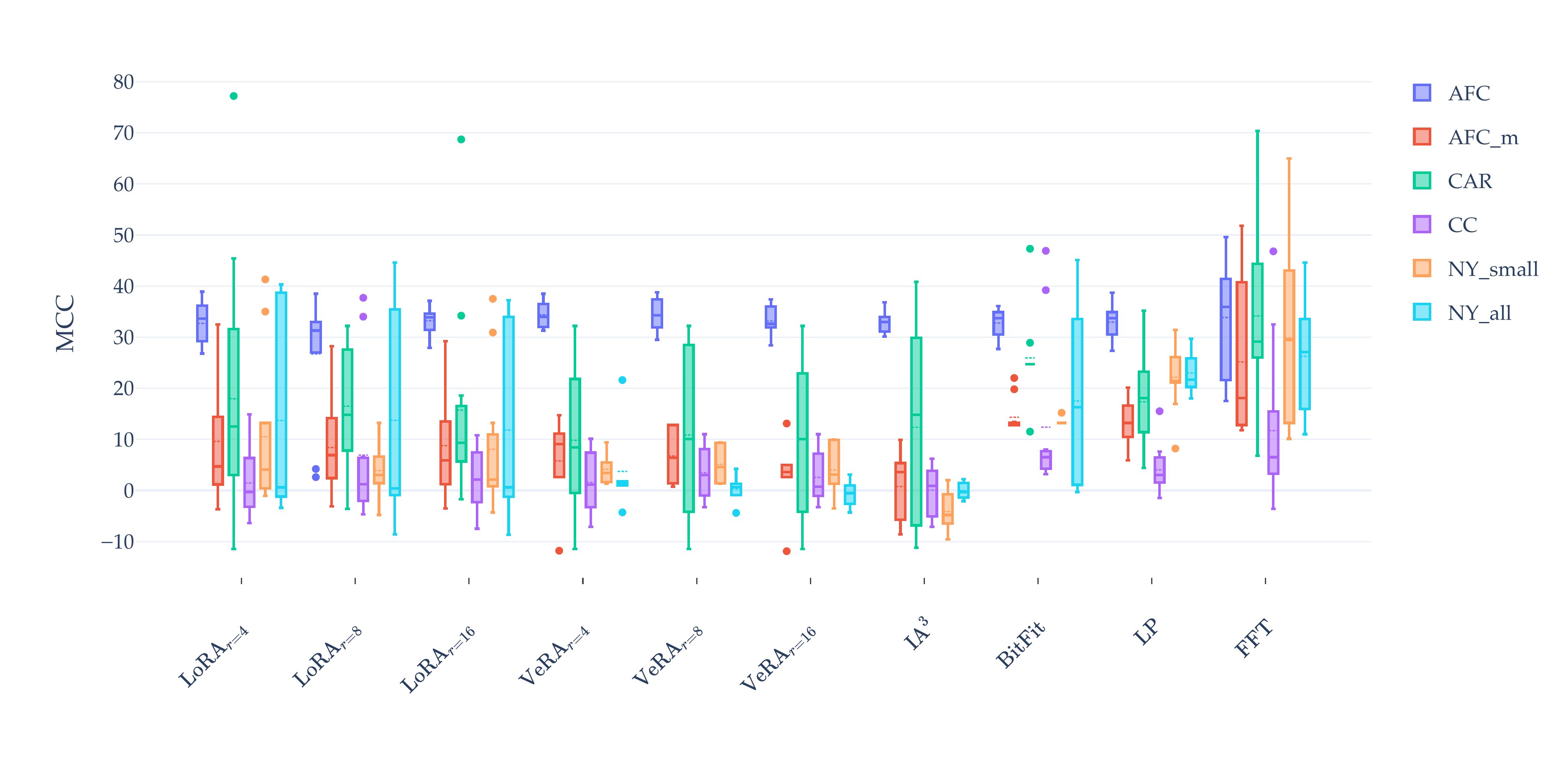}
  }
\caption{
Distribution of performance scores across fine-tuning methods and datasets. 
Each box plot summarizes the mean performance of all models fine-tuned with a given technique on a specific dataset. 
Boxes indicate the inter-quartile range for all the mean result obtained from all the models fine-tuned with a specific technique on a specific dataset (divided by color), while the central line represents the median and whiskers extend to the minimum and maximum values.
}

  \label{fig:finetuning_comparison}

\end{figure}
\bl

This section presents and discusses the main experimental results obtained following the experimental methodology described in Section~\ref{sec:Methods}. 
All reported results are evaluated using the MCC and PR-AUC scores, which serve as the primary performance metrics. 
A complete summary is presented in Table~\ref{tab:results_MCC} and Table~\ref{tab:results_AUC}, which report all experimental outcomes and highlight the top- and second-performing model–technique pairs for each dataset, determined according to the highest mean value of the MCC and PR-AUC metrics, respectively. 
From the MCC score in Table~\ref{tab:results_MCC}, we further derived Figure~\ref{fig:finetuning_comparison}, displaying the inter-quartile range of the mean results obtained from models fine-tuned with each specific technique on each dataset.
\medskip

In addition, to formally support comparative claims, all methods were further assessed through a non-parametric statistical analysis. 
Specifically, we employed the one-tailed Wilcoxon signed-rank test on the rank distributions derived from the mean MCC values of the models across all datasets. This approach yields a pairwise significance matrix (shown in Table~\ref{tab:CNN_vs_FM_stat} and Table~\ref{tab:FM_stat}), which reports, for each method pair, the number of datasets in which the method on a given row outperforms the method on the corresponding column and the eventual statistical significance in terms of $p$-values, annotated using the conventional significance levels “*” for $p < 0.05$. This non-parametric framework provides a reliable indication of whether observed differences are statistically meaningful rather than arising from random variation.
\medskip

The following results are organized into two main parts. The \textbf{Setup} paragraph outlines the specific experimental configuration, including the models and PEFT methods analyzed, while the \textbf{Results} paragraph provides an in-depth analysis of the results obtained through the combination of models, datasets, and fine-tuning strategies described in Section~\ref{sec:Methods}. 

\bb


\input{tables/results_MCC}

\input{tables/results_AUC}

\subsection{What are the most effective fine-tuning strategies for adapting CNN and FM architectures to prognosis tasks?}\label{Q:FT_cnn_fm}

Understanding the effectiveness of fine-tuning strategies across different architectural families, CNNs and FMs, is essential for building deployable and scalable systems in medical imaging prognosis. While CNNs have been widely adopted due to their simplicity and inductive biases, recent advancements in large-scale pretrained vision transformers (e.g., ViT, CLIP, DINO) have positioned FMs as highly competitive alternatives. However, it remains unclear how these architectures respond under constrained regimes typical of prognosis data settings, characterized by strong class imbalance and reduced samples sizes, and whether PEFT can consistently bridge the adaptation gap across architectures. Moreover, we extend our analysis to include extreme fine-tuning configurations, such as LP and FFT, to further characterize the performance boundaries of minimal versus maximal adaptation strategies.

\begin{figure}[!t]
    \centering
    \resizebox{\textwidth}{!}{
    \begin{tikzpicture}
        \node[label={[font=\Large, yshift=.5cm]below: (a) AFC$_m$}] (A) at (0,0) {\includegraphics[width=\textwidth]{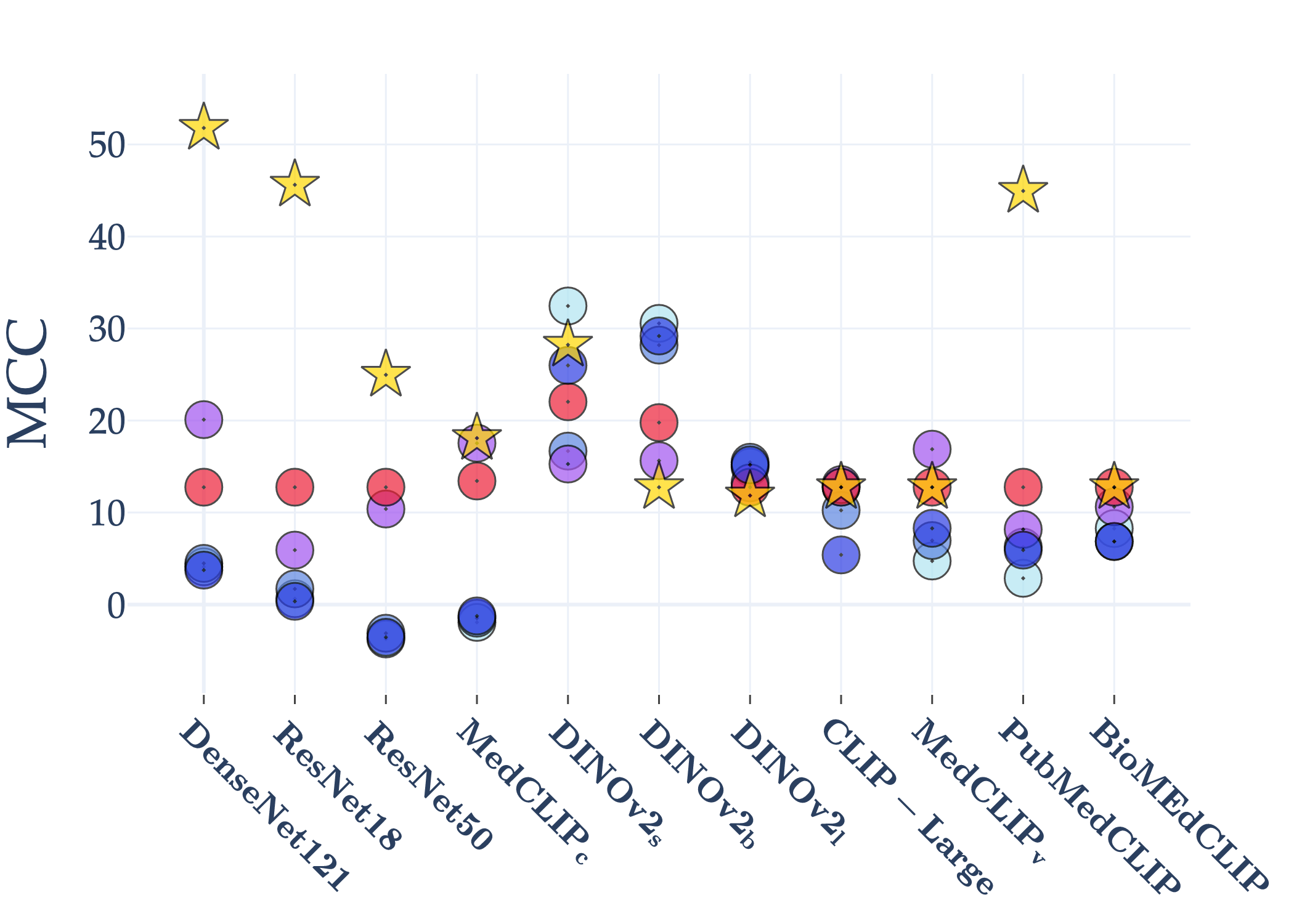}};
        
        \node[label={[font=\Large, yshift=.5cm]below: (b) AFC}] (B) [right=0.5cm of A] {\includegraphics[width=\textwidth]{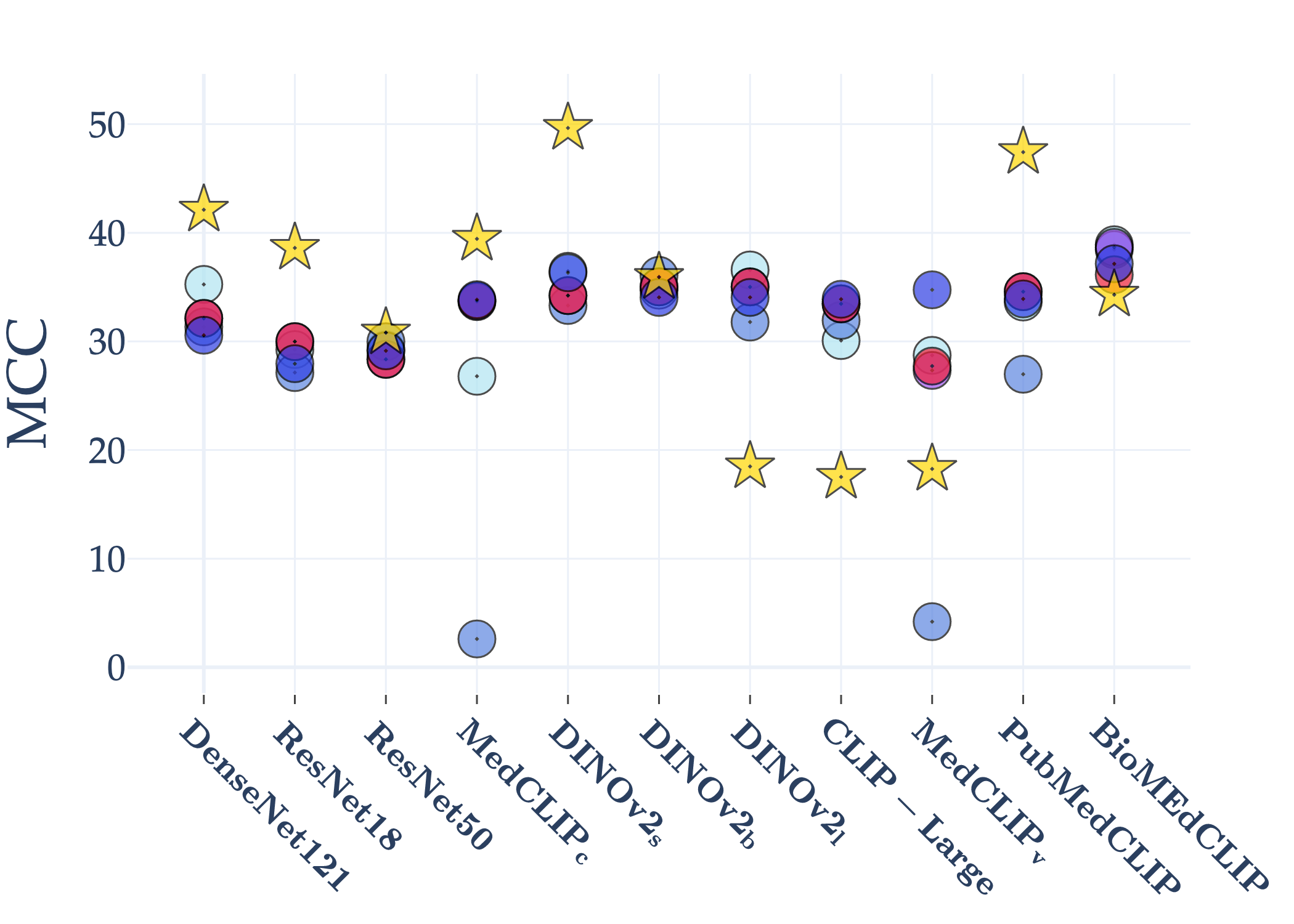}};
        
        \node[label={[font=\Large, yshift=.5cm]below: (c) CAR}] (C) [below=0.2cm of A] {\includegraphics[width=\textwidth]{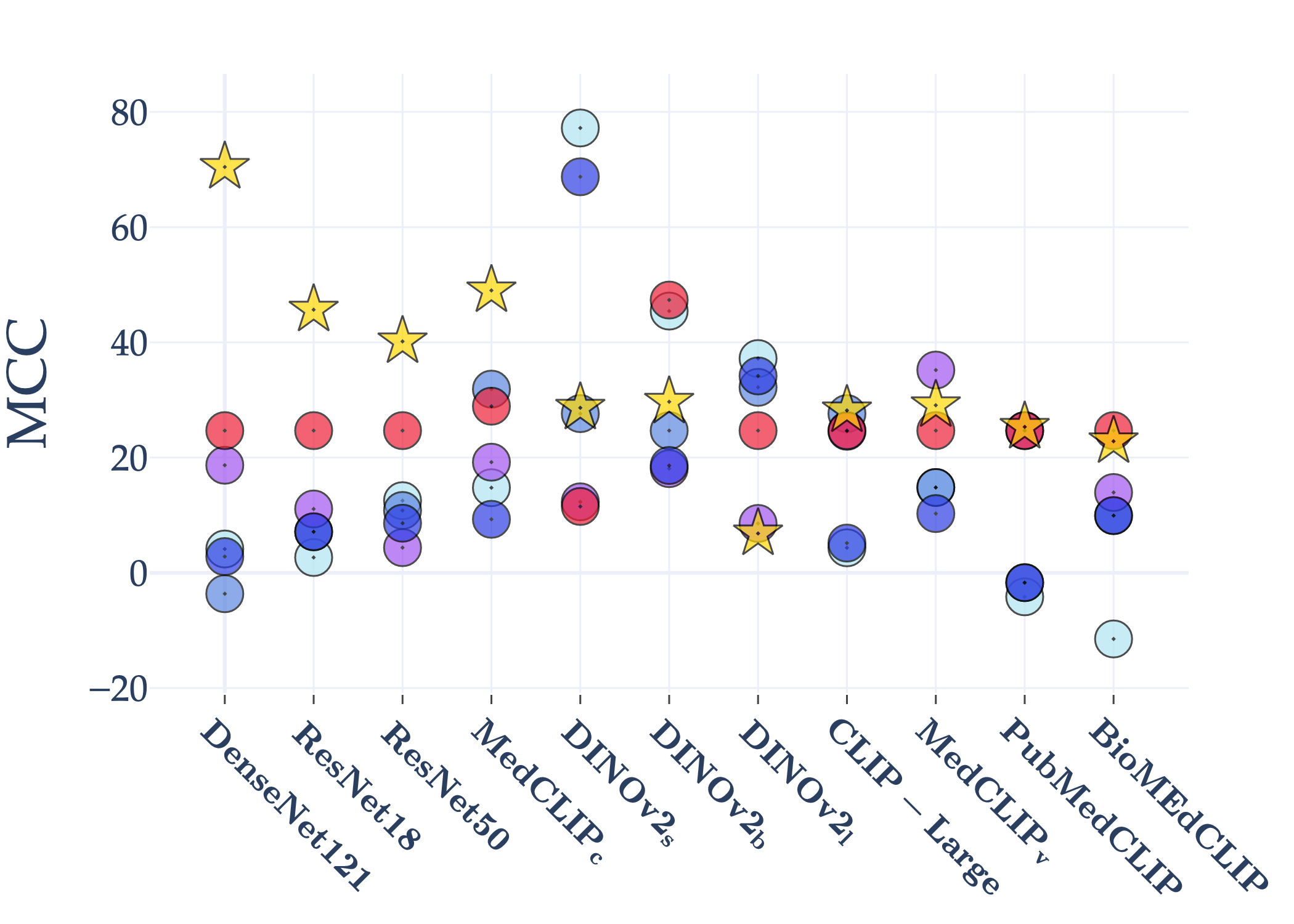}};
        
        \node[label={[font=\Large, yshift=.5cm]below: (d) CC}] (D) [right=0.5cm of C] {\includegraphics[width=\textwidth]{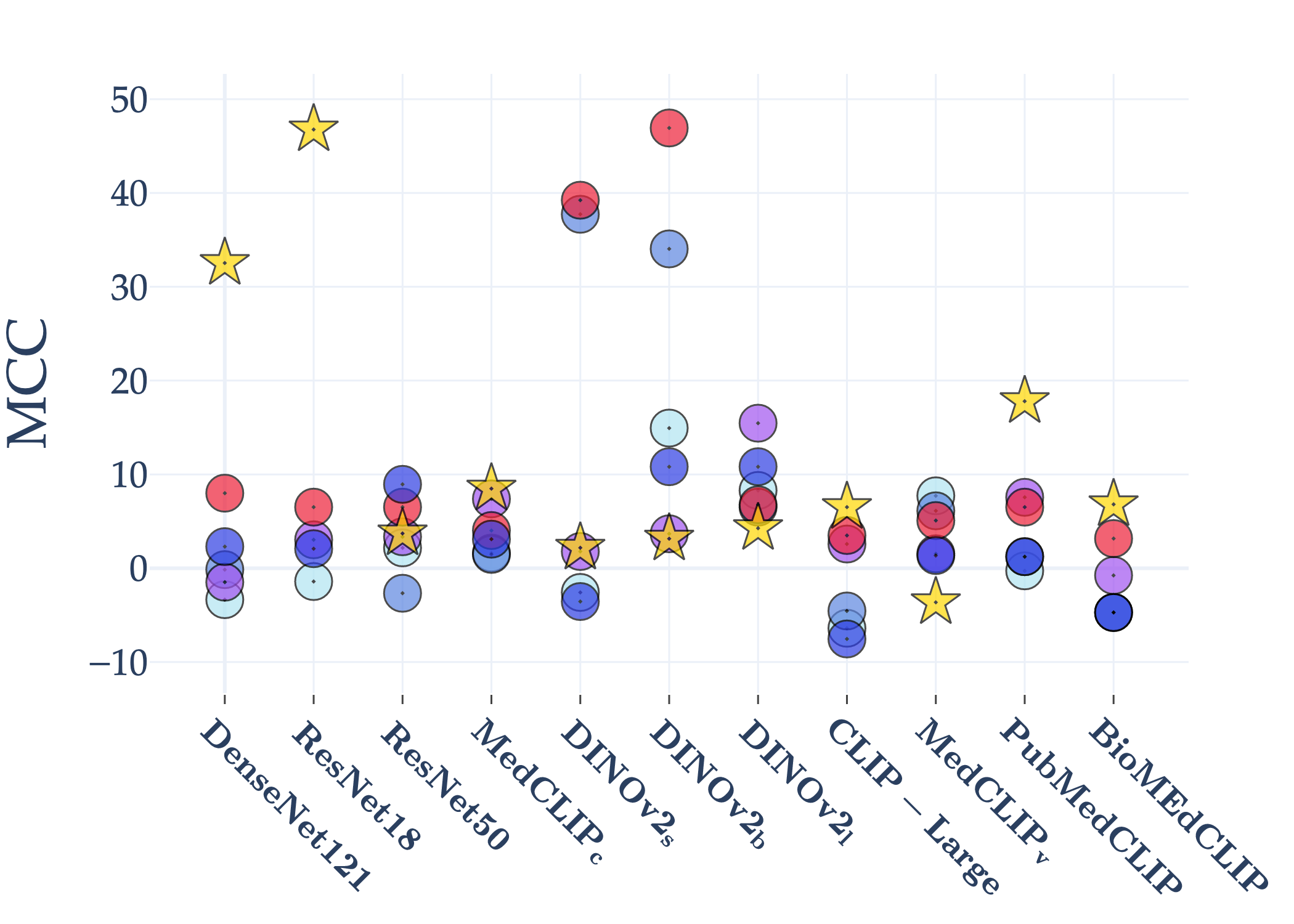}};
        
        \node[label={[font=\Large, yshift=.5cm]below: (e) NY\textsubscript{small}}] (E) [below=0.2cm of C] {\includegraphics[width=\textwidth]{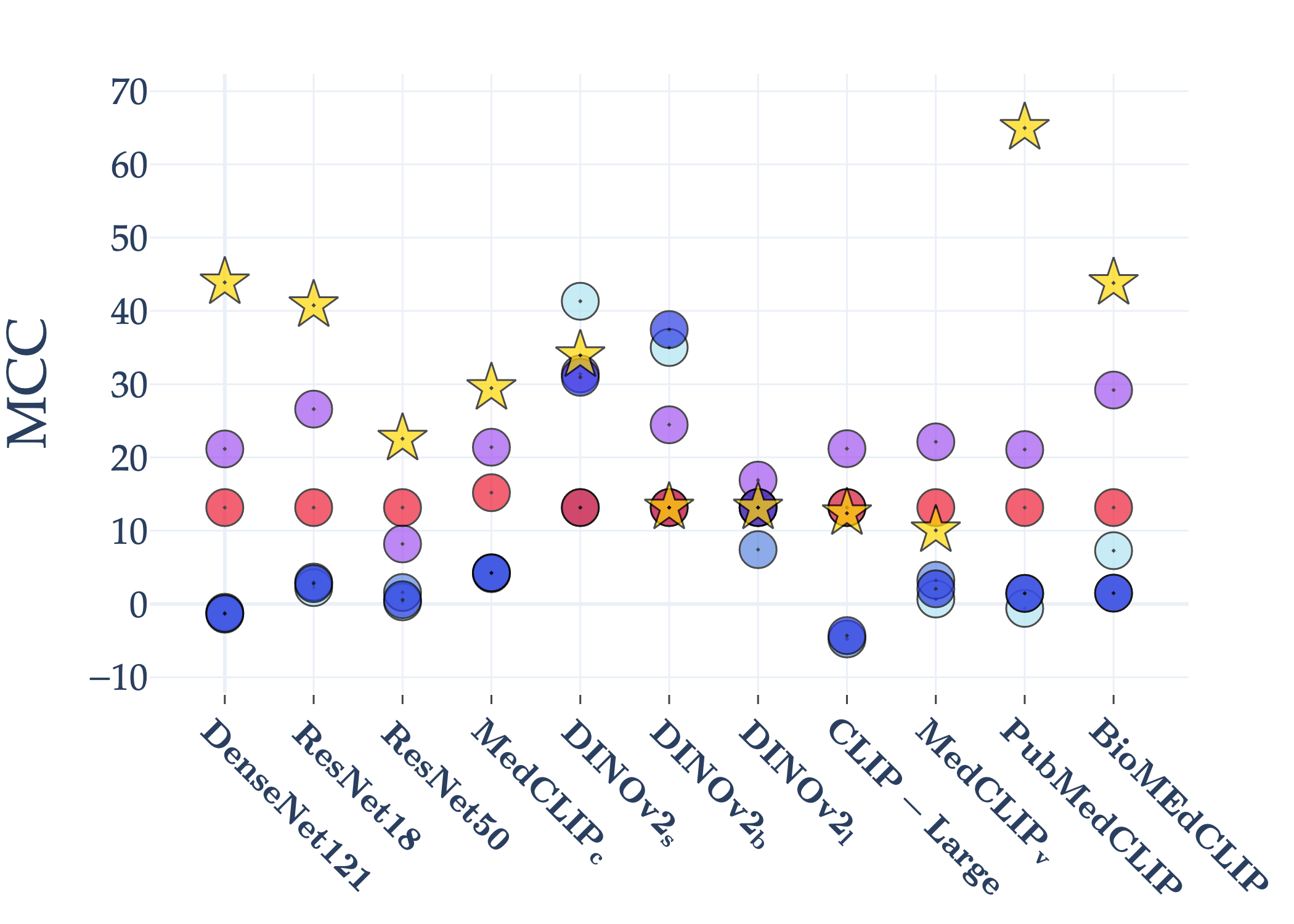}};
        
        \node[label={[font=\Large, yshift=.5cm]below: (f) NY\textsubscript{all}}] (F) [right=0.5cm of E] {\includegraphics[width=\textwidth]{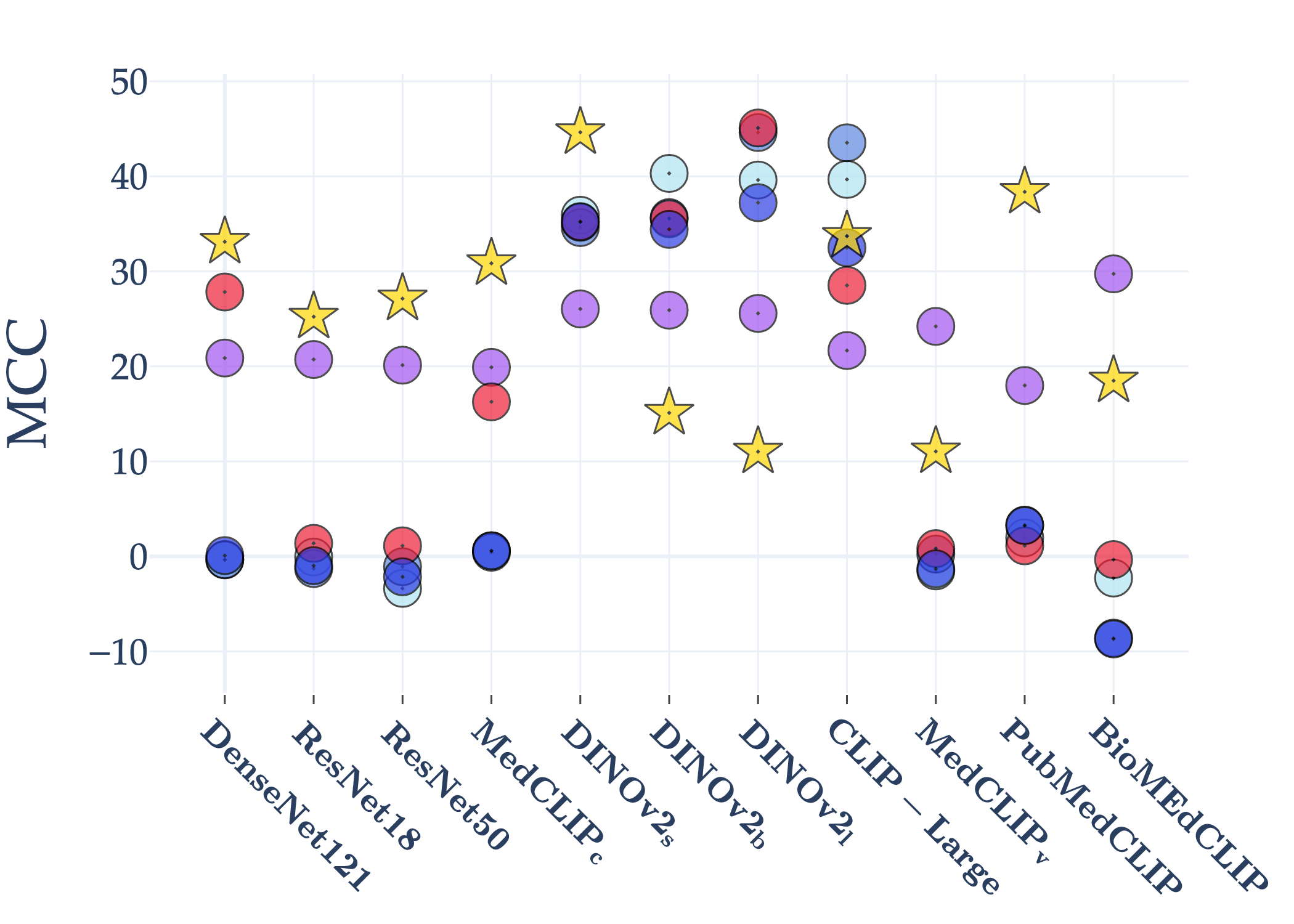}};

    \end{tikzpicture}
    }
    \caption{This figure includes only the fine-tuning techniques applicable to both $CNN$ and $FM$ architecture families. The mean performance over the test set folds for all fine-tuning methods is represented by the \emoji{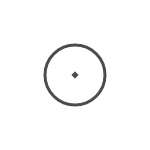} symbol, following the color scheme:
\textcolor[HTML]{b1e5f2}{$\mathbf{LoRA_{r=4}}$}, \textcolor[HTML]{5c86e0}{$\mathbf{LoRA_{r=8}}$}, \textcolor[HTML]{2d3de3}{$\mathbf{LoRA_{r=16}}$};
\textcolor[HTML]{ed1f37}{$\mathbf{BitFit}$}, \textcolor[HTML]{a254f0}{$\mathbf{LP}$}. The FFT method is uniquely represented by the \emoji{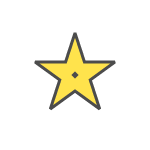} symbol.}

\label{fig:CNN_vs_FM}
\end{figure}

\begin{figure}[ht]
    \centering
    \includegraphics[width=0.75\linewidth]{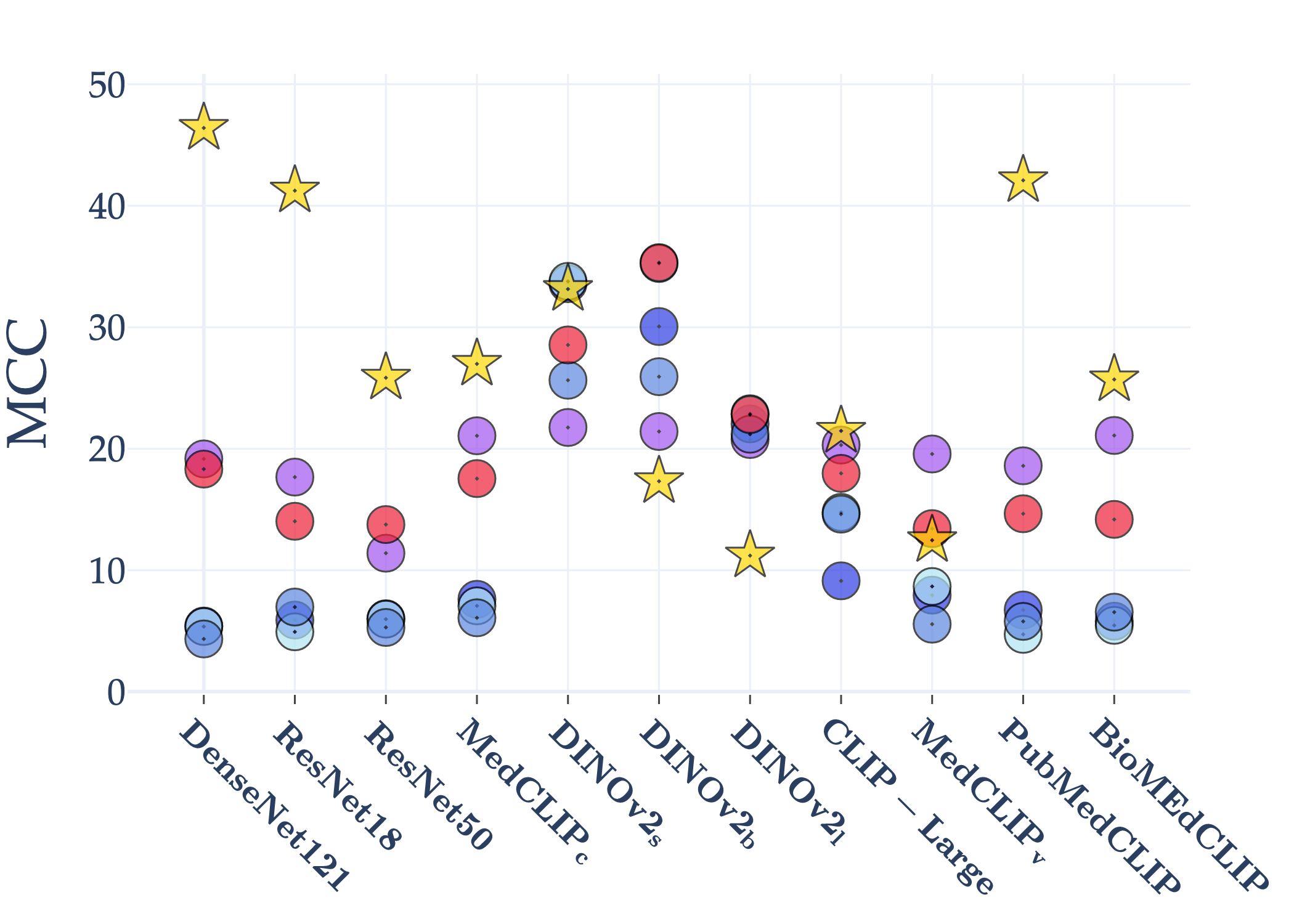}
    \caption{This figure displays the mean over all datasets performances respect only the fine-tuning techniques applicable to both $CNN$ and $FM$ architecture families. The overall mean for the fine-tuning methods is represented by the \emoji{figures/clip-large_symbol_.png} symbol, following the color scheme:
\textcolor[HTML]{b1e5f2}{$\mathbf{LoRA_{r=4}}$}, \textcolor[HTML]{5c86e0}{$\mathbf{LoRA_{r=8}}$}, \textcolor[HTML]{2d3de3}{$\mathbf{LoRA_{r=16}}$};
\textcolor[HTML]{ed1f37}{$\mathbf{BitFit}$}, \textcolor[HTML]{a254f0}{$\mathbf{LP}$}. The FFT method is uniquely represented by the \emoji{figures/CNN_FM_symbol_star_.png} symbol.}
    \label{fig:CNN_vs_FM_overall}
\end{figure}

\subsubsection{Setup.} 
We fine-tuned all the models introduced in Section~\ref{sec:MODELS} on the CXR prognosis tasks described in Section~\ref{section:DATA}. 
To ensure a fair comparison with Table~\ref{tab:peft-table}, we included only the fine-tuning techniques applicable to both architecture families, that are LoRA$_{r\in {4,8,16}}$, BitFit, LP and FFT. 
\bl
The final results obtained are subsequently compared numerically and statistically, where the first analysis is supported by Figure~\ref{fig:CNN_vs_FM}, and the second by Table~\ref{tab:CNN_vs_FM_stat}.
\bb

\input{tables/CCN_vs_FM_stat}

\subsubsection{Results.} Figure~\ref{fig:CNN_vs_FM} presents the aggregated performance, obtained averaging the results across of the fine tuning techniques by dataset and by model.
Starting from the dataset-level comparison.
In the AFC and AFC$_m$ tasks, PEFT fine-tuning in AFC (Figure~\ref{fig:CNN_vs_FM}.a) yields consistent performance across architectures, indicating that dataset balance mitigates architectural variability.
At the same time DINOv2$_s$ and PubMedCLIP achieve the best performance, along with CNNs fine-tuned using FFT. 
These results suggest that PEFT techniques can successfully leverage pretrained representations in balanced class distribution datasets, without being influenced by model choice~\cite{Wang2025}.
However, in AFC$_m$ (Figure~\ref{fig:CNN_vs_FM}.b), where class distribution is highly imbalanced, PEFT performance drops significantly, yielding results around $10\%$ in MCC. 
Interestingly, PEFT methods remain effective only on DINOv2 variants, while FFT consistently outperforms other approaches across all models. 
These results confirm the strong sensitivity of PEFT techniques to data imbalance, where models struggle to generalize effectively under skewed class distributions.
Lastly, a comparative analysis between NY\textsubscript{small} and NY\textsubscript{all} (Figures~\ref{fig:CNN_vs_FM}.e,~\ref{fig:CNN_vs_FM}.f) highlights dataset size as a critical factor in PEFT effectiveness. 
In NY\textsubscript{small}, most PEFT methods fail to achieve meaningful improvements, with only certain DINOv2$_s$ variants showing moderate success, while FFT on PubMedCLIP emerges as the top performer. 
In NY\textsubscript{all}, PEFT techniques exhibit significantly higher performance, in some cases matching or even competing with FFT, particularly on DINOv2 and CLIP-Large. 
This confirms that PEFT methods require a sufficiently large dataset to effectively adapt and optimize their representations. 

In general, the overall findings suggest that CNNs remain more resilient for extremely small datasets, likely due to their lower parameter count and the efficiency conferred by their inductive biases, even if not derived from pretraining the model on medical domain datasets. 
In contrast, FMs exhibit reduced performance under by the imbalance that characterize prognosis datasets. FMs, as DINOv2 large, fail to generalize in this context, likely due to their elevated parameter count, disproportionate respect the limited dimensionality and scale of the available training data.
Despite this, given we obtain better results with the same model architecture (DINOv2) but using less parametrized variants. These less parametrized variants demonstrate that FMs combined with PEFT can extract meaningful representations even in extreme low-data regimes. 
\bl

The results reveal that the best-performing PEFT strategies are found at the two extremes of parameter efficiency: FFT and the most lightweight ones, LP and also BitFit, consistently outperform more efficiency-oriented methods as LoRA. 
This outcome suggests that both highly expressive and highly constrained adaptation schemes can effectively leverage the representational capacity of CNN and FMs, whereas mid-range configurations may struggle to strike a suitable balance between adaptation flexibility and regularization strength.
These observations are corroborated by the statistical analysis reported in Table \ref{tab:FM_stat}, where FFT, LP, and BitFit achieve superior performance in most datasets, showing significant advantages over LoRA-based methods. This pattern is further reflected in the box-plot trends shown in Figure \ref{fig:finetuning_comparison}, which confirm the consistent dominance of these approaches across tasks.
\bb

In conclusion, PEFT methods are highly sensitive to class imbalance and dataset size. Severe imbalance and small sample sizes negatively impact their performance, whereas larger datasets, even with moderate imbalance, result in significant improvements. DINOv2$_s$ exhibit an interesting trend in FFT setting: smaller variants tend to outperform larger ones, as seen in Figure~\ref{fig:CNN_vs_FM_overall}. This suggests that excessive model complexity may not always translate to better fine-tuning outcomes in low-resource settings. 
Furthermore, in the comparison between MedCLIP$_{c}$ and its supervised pretrained equivalent ResNet50, it is evident that despite MedCLIP$_{c}$ was pretrained using SSL, the performance gains are lower than expected. Across most fine-tuning techniques, it achieves only a marginal improvement, with increases of just a few percentage points in the metric score. 
\bl
Despite this, all the methods suffer an elevated variance across folds, as reported in Table~\ref{tab:results_MCC} and Table~\ref{tab:results_AUC}.
Such variability reflects the intrinsic instability of model optimization in low-sample regimes, where limited data dimensionality amplifies stochastic effects.
\bb

Given the importance of PEFT techniques applied to FMs, we henceforth concentrate our analysis on this paradigm. 
This decision aligns with established knowledge in the literature, notably Dutt et al.~\cite{dutt2023parameter}, who demonstrated that the advantages of PEFT typically grow with decreasing data volume and increasing model size, particularly benefiting small- to medium-sized datasets common in medical imaging. 
However, our we extend existing understanding by highlighting important limitations:  
we show that CNN architectures fine-tuned with FFT maintain superior mean performance in extremely small or severely imbalanced datasets. 
In the following section, we analyze the results obtained from all PEFT techniques applied exclusively to foundation models, aiming to quantify the relative importance and effectiveness of these strategies.

\subsection{Which fine-tuning strategy provides the best trade-off between efficiency and effectiveness when applied to medical prognosis tasks using FMs?}\label{Q:FM}

The design of an effective fine-tuning strategy is a key decision point in adapting large pretrained models to prognosis tasks. Despite FMs ignited new opportunities in medical imaging, the optimal fine-tuning approach remains unclear. These techniques promise reduced computational and memory costs by updating only a small subset of the model's parameters, yet their ability to retain performance under medical data constraints is still underexplored. 
\begin{figure}[!ht]
    \centering
    \resizebox{\textwidth}{!}{
    \begin{tikzpicture}
        \node[label={[font=\Large, yshift=.5cm]below: (a) AFC$_m$}] (A) at (0,0) {\includegraphics[width=\textwidth]{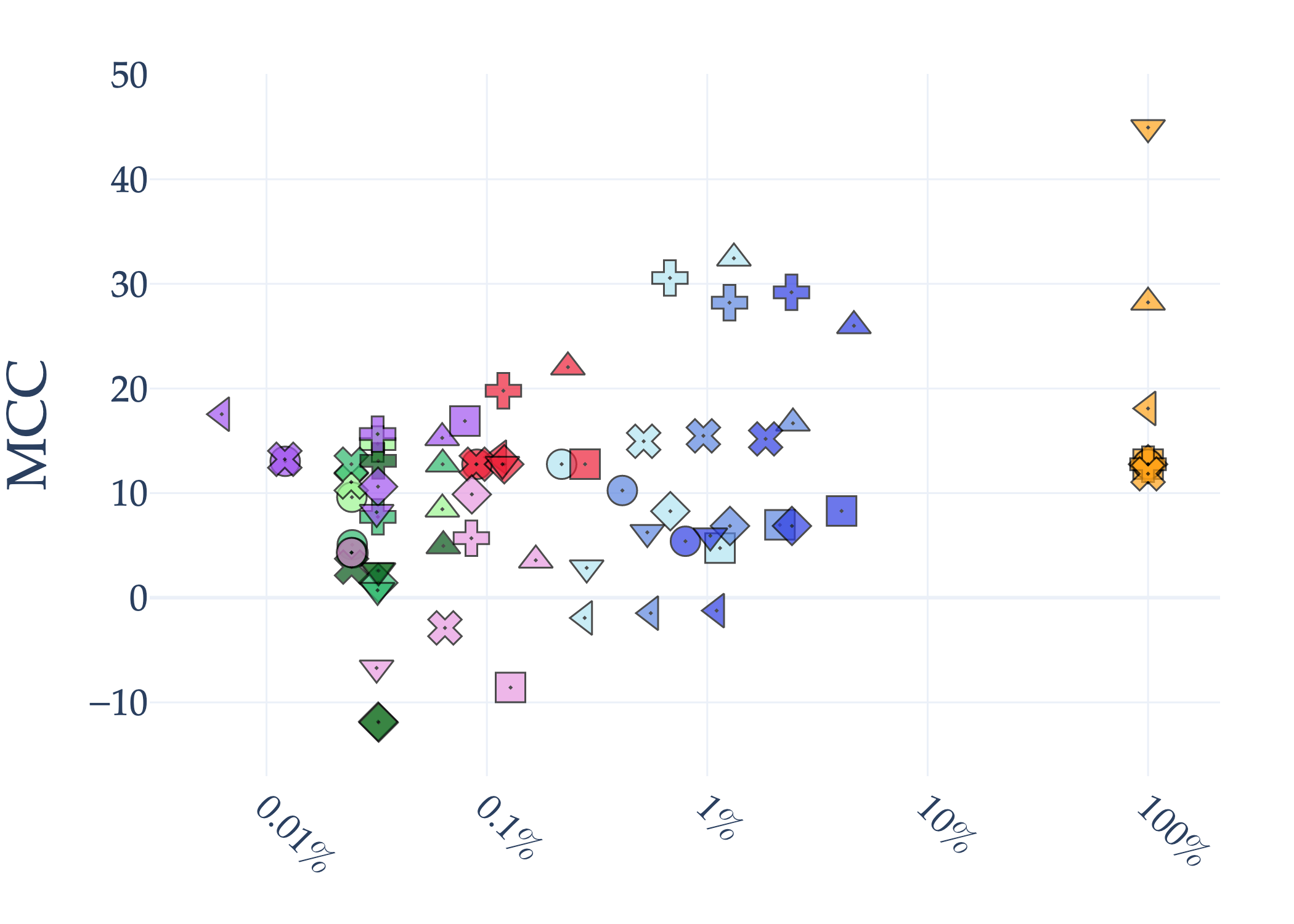}};
        \node[label={[font=\Large, yshift=.5cm]below: (b) AFC}] (B) [right=0.5cm of A] {\includegraphics[width=\textwidth]{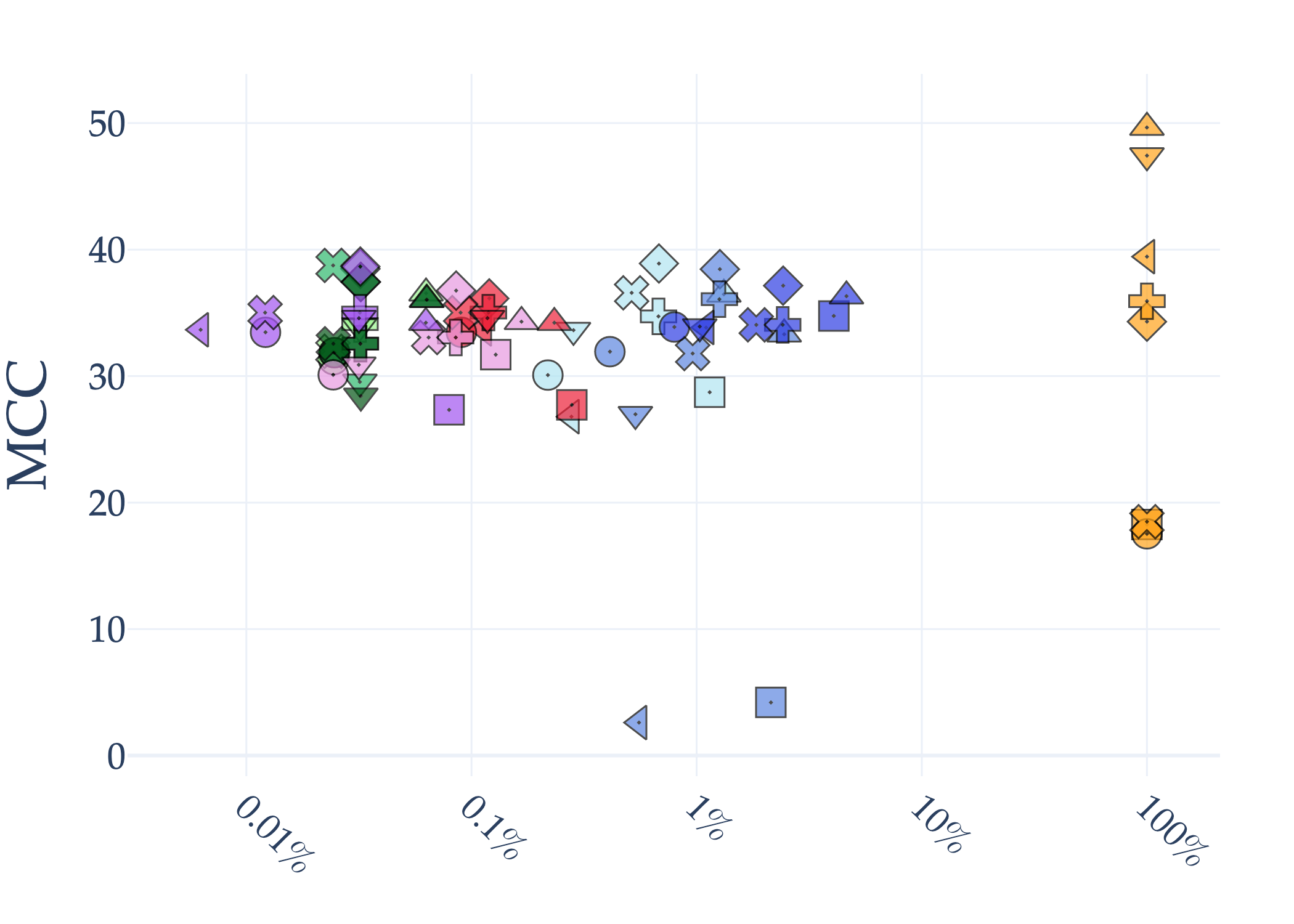}};
        \node[label={[font=\Large, yshift=.5cm]below: (c) CAR}] (C) [below=0.2cm of A] {\includegraphics[width=\textwidth]{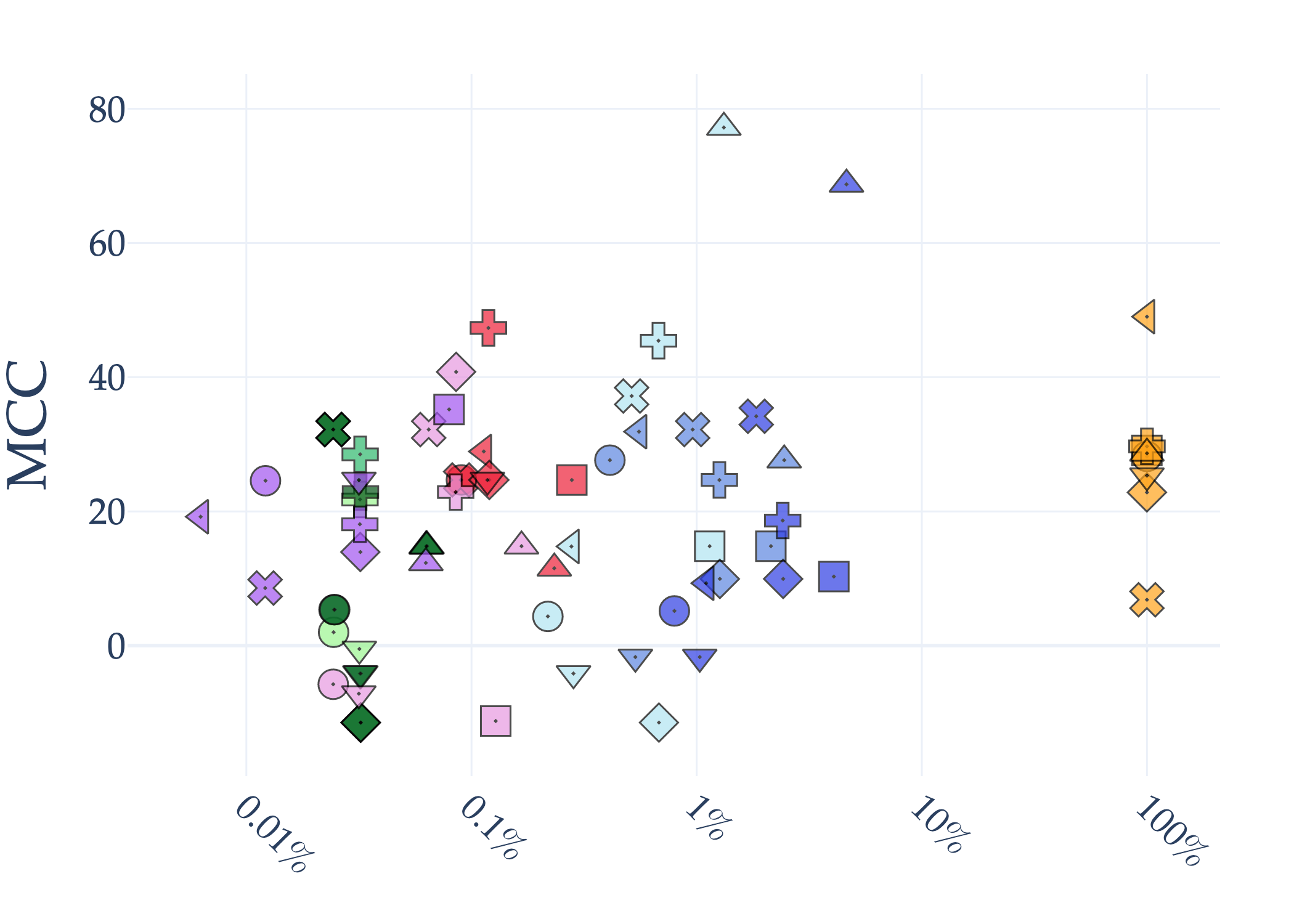}};
        \node[label={[font=\Large, yshift=.5cm]below: (d) CC}] (D) [right=0.5cm of C] {\includegraphics[width=\textwidth]{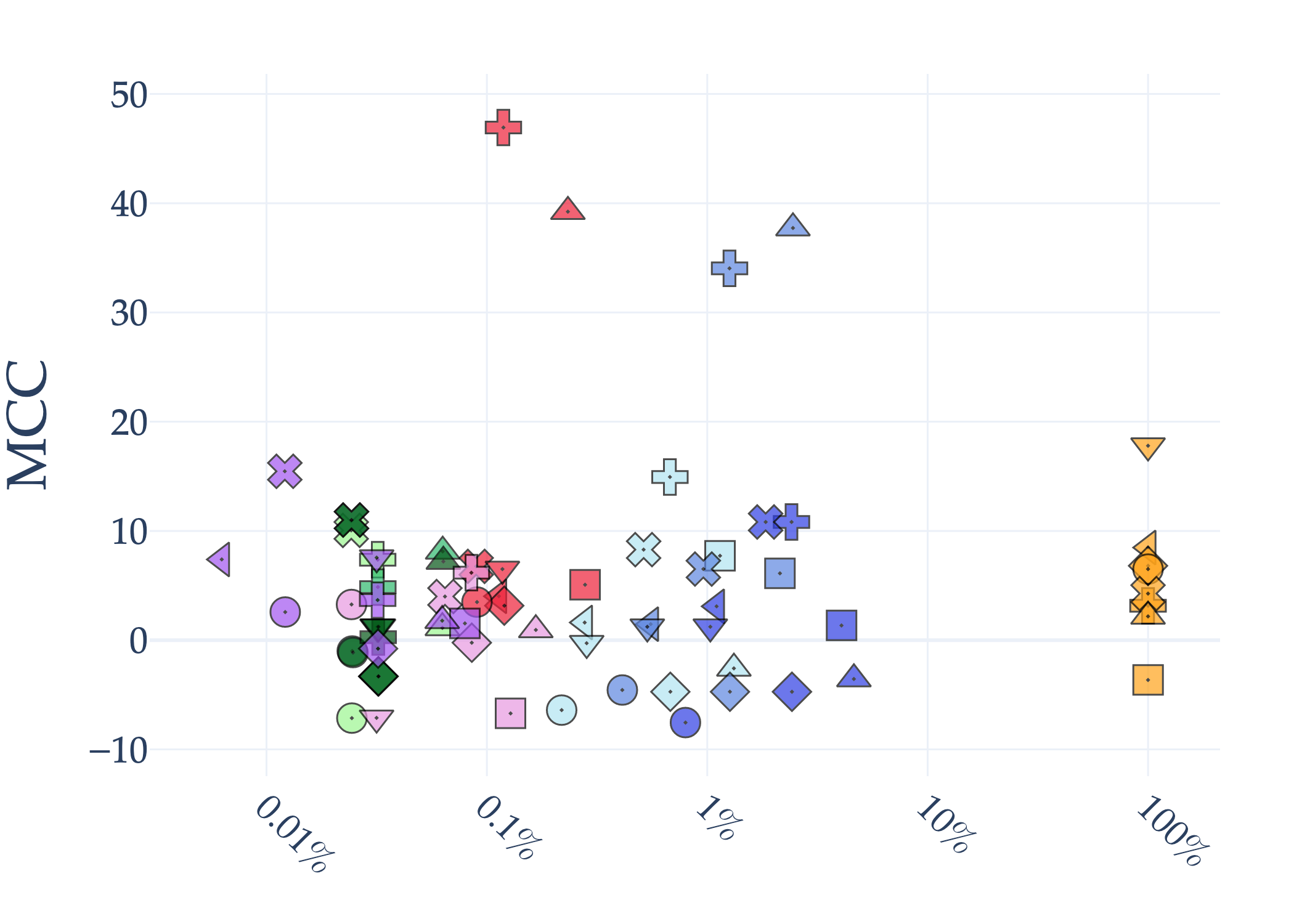}};
        \node[label={[font=\Large, yshift=.5cm]below: (e) NY\textsubscript{small}}] (E) [below=0.2cm of C] {\includegraphics[width=\textwidth]{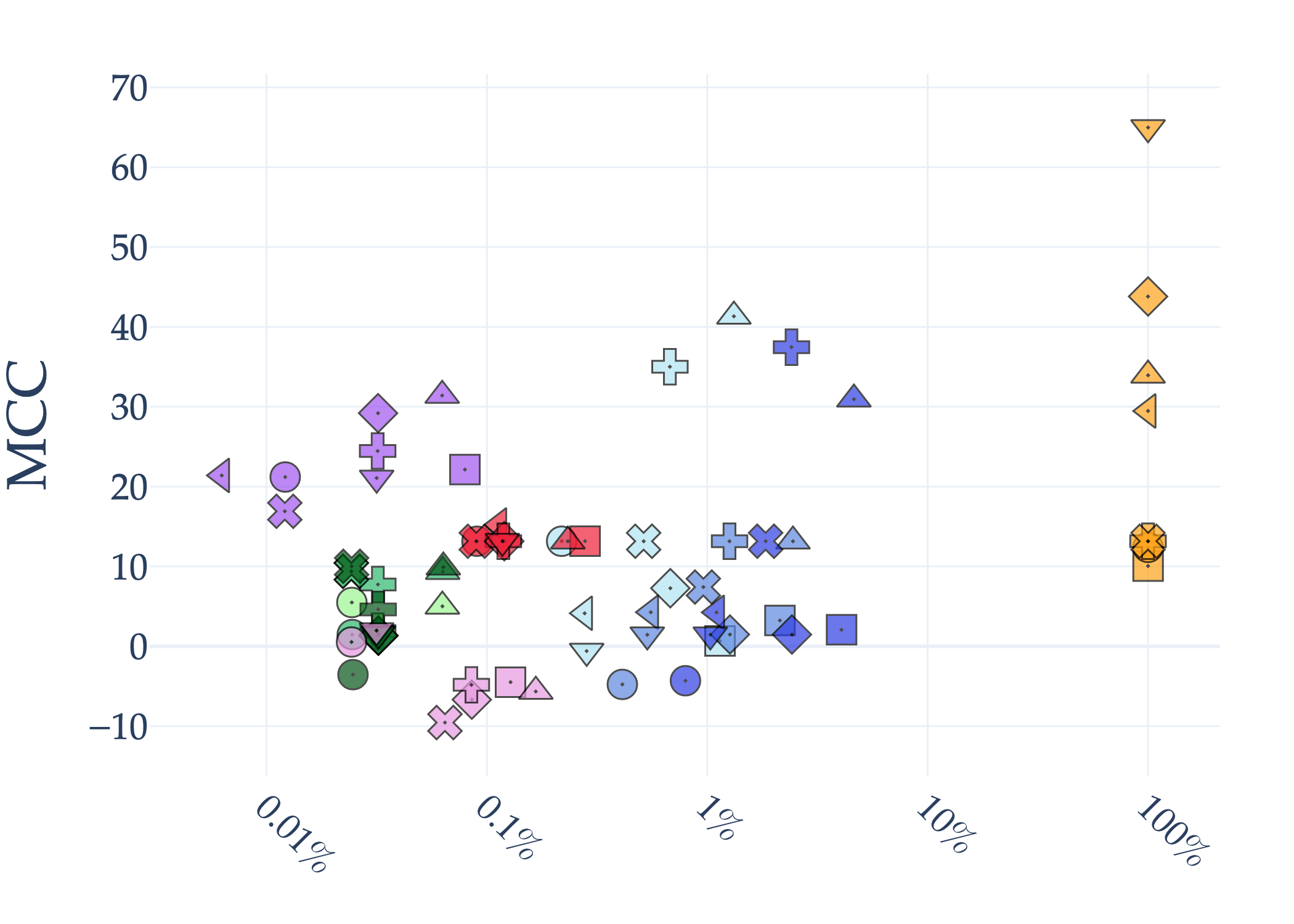}};
        \node[label={[font=\Large, yshift=.5cm]below: (f) NY\textsubscript{all}}] (F) [right=0.5cm of E] {\includegraphics[width=\textwidth]{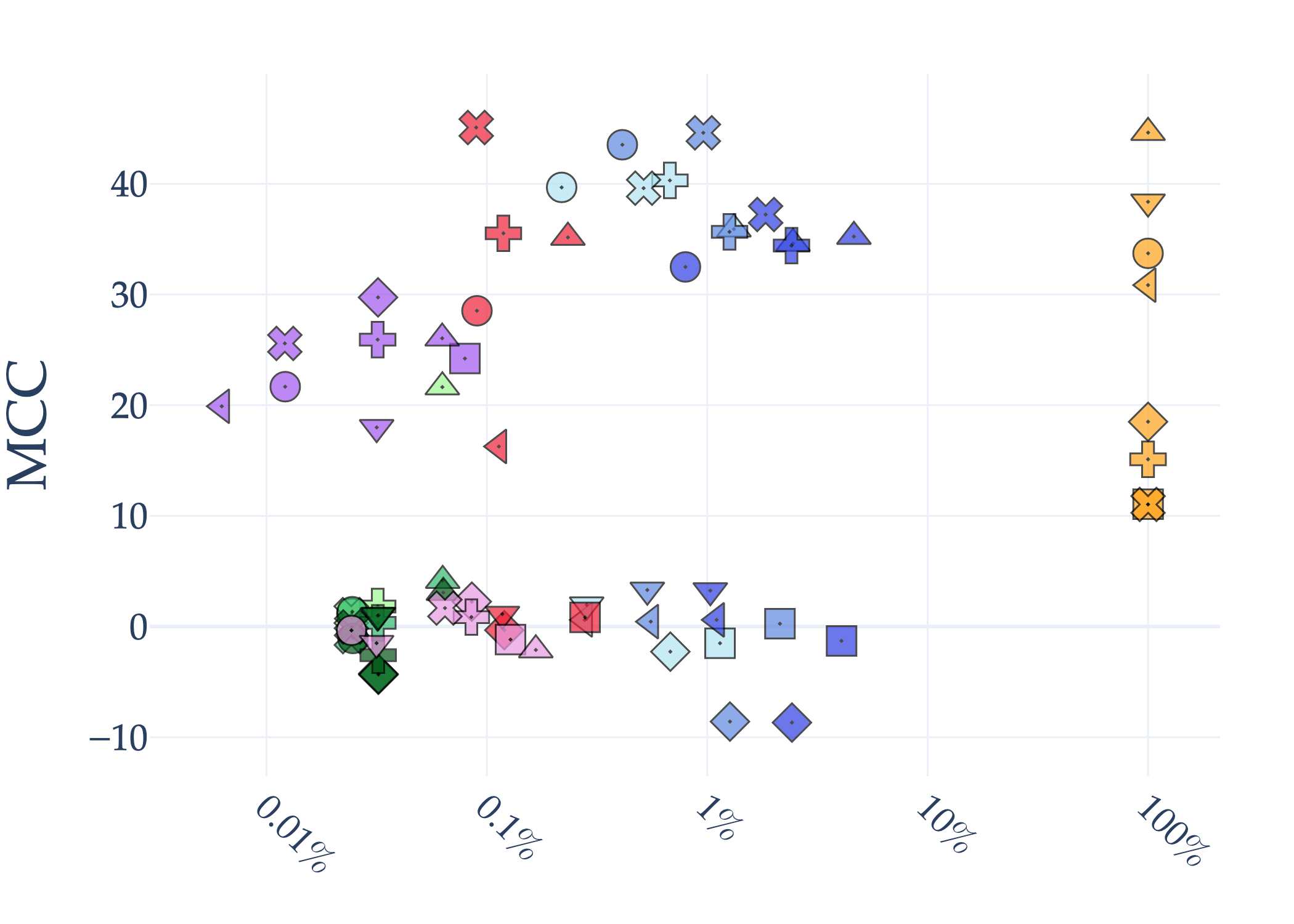}};
    \end{tikzpicture}
    }
    \caption{This figure presents one subplot for each dataset, illustrating on the $y$ axis the MCC results and on the $x$ axis the percentage of models' parameters trained over the total. Fine tuning techniques and models have a double encoding systems, one based on colors as \textcolor[HTML]{b1e5f2}{$\mathbf{LoRA_{r=4}}$}, \textcolor[HTML]{5c86e0}{$\mathbf{LoRA_{r=8}}$}, \textcolor[HTML]{2d3de3}{$\mathbf{LoRA_{r=16}}$}, \textcolor[HTML]{9cf590}{$\mathbf{VeRA_{r=4}}$}, \textcolor[HTML]{2eb86c}{$\mathbf{VeRA_{r=8}}$}, \textcolor[HTML]{025415}{$\mathbf{VeRA_{r=16}}$}, \textcolor[HTML]{ed1f37}{$\mathbf{BitFit}$}, \textcolor[HTML]{e899e0}{$\mathbf{IA^3}$}, \textcolor[HTML]{a254f0}{$\mathbf{LP}$}, and \textcolor[HTML]{ffa319}{$\mathbf{FFT}$}; and one based on the model \emoji{figures/clip-large_symbol_.png}($CLIP\text{-}Large$), \emoji{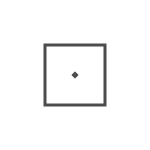}(\textit{MedCLIP$_{v}$}), \emoji{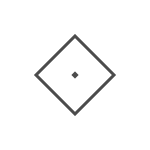} ($BioMEdCLIP$), \emoji{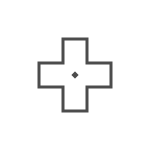} (\textit{DINOv2$_b$}), \emoji{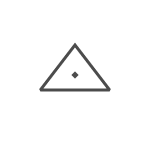}(DINOv2$_s$),
    \emoji{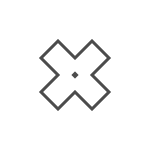} (DINOv2$_l$), \emoji{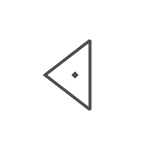} (MedCLIP$_{c}$), and \emoji{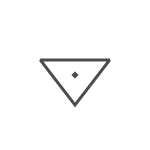} (PubMedCLIP).}
    \label{fig:FM_all_results_by_dataset}
\end{figure}

\subsubsection{Setup.}
In this section, we analyze various fine-tuning techniques applied exclusively to FMs, with the combination of PEFT methods and models following Table~\ref{tab:peft-table}. To maintain consistency, all models were trained using the same experimental settings as discussed in Section~\ref{section:exp_stp}.
This analysis aims to provide insights into FMs as a pretraining paradigm, rather than being tied to a specific model architecture. Figure~\ref{fig:FM_all_results_by_dataset} presents the results, where the x-axis represents the percentage of trained FM parameters over the total, and the y-axis indicates the average metric value obtained on the test sets.
\bl
To complement this visual analysis, Table~\ref{tab:FM_stat} reports the corresponding pairwise statistical comparison among all fine-tuning strategies, quantifying their relative effectiveness and highlighting statistically significant differences across fine tuning strategies.
\bb
\input{tables/FMs_stat}

\subsubsection{Results.}We conducted a comprehensive analysis of the results presented in Figure~\ref{fig:FM_all_results_by_dataset}, examining performance with respect to different dataset characteristics, including sample size and class distribution. Furthermore, we evaluated PEFT efficiency by investigating the relationship between model performance and the proportion of model parameters adapted during fine-tuning. This analysis aimed to quantify the effectiveness of PEFT methods in achieving task-specific adaptation while modifying only a small fraction of the original model parameters.

In smaller datasets such as CAR and CC (Figures~\ref{fig:FM_all_results_by_dataset}.c and~\ref{fig:FM_all_results_by_dataset}.d), most fine-tuning techniques fail to deliver significant improvements, suggesting that despite their parameter efficiency, these methods struggle to adapt and generalize under data constraints. This is particularly evident in Figure~\ref{fig:FM_all_results_by_dataset}.d, where most techniques plateau around $10\%$ MCC, except for LoRA$_{r=8}$ and BitFit, which perform notably better when applied to DINOv2$_l$ and DINOv2$_s$.

For CAR, results are generally more favorable, with several experiments reaching up to $30\%$ MCC. Notably, LoRA${r=4}$ and LoRA${r=16}$ applied to DINOv2$_s$ achieve performance levels approaching $80\%$, demonstrating that certain configurations can still extract meaningful representations even under data scarcity. However, across both datasets, FFT consistently underperforms, likely due to the limited sample size, which prevents effective optimization of all parameters.

These results indicate that the most parameter-efficient techniques are not necessarily the most effective in low-data regimes. The efficiency–effectiveness trade-off, illustrated in Figure~\ref{fig:CNN_vs_FM}, shows that performance is influenced not only by the number of trainable parameters but also by the interplay between dataset complexity, class imbalance, and the quality of pretrained representations. In particular, methods such as LoRA and BitFit perform well on DINOv2 variants because their adaptation mechanisms align closely with the model’s pretrained feature organization, allowing efficient transfer of domain-relevant knowledge. Conversely, methods involving full fine-tuning (FFT) are more sensitive to limited data, as their larger parameter space increases overfitting risk. Hence, there is no simple correlation between parameter count and performance; rather, success depends on how effectively each PEFT method exploits pretrained knowledge to capture features discriminant for prognosis prediction.

Another key finding concerns the impact of task imbalance on performance. 
This aspect is investigated by comparing AFC and AFC$_m$ (Figures~\ref{fig:FM_all_results_by_dataset}.a and~\ref{fig:FM_all_results_by_dataset}.b), which share the same number of samples but differ in class distribution. In the balanced dataset AFC (Figure~\ref{fig:FM_all_results_by_dataset}.a), all fine-tuning methods achieve stable results around $\sim30\%$ mean of MCC score, with FFT performing best for DINOv2$_s$ and PubMedCLIP. In contrast, the imbalanced task AFC$_m$ (Figure~\ref{fig:FM_all_results_by_dataset}.b) exhibits a substantial drop in performance despite identical data volume. 
Notably, LoRA and BitFit still reach $35\%$ mean of MCC score, indicating some resilience to imbalance. This highlights that dataset properties, beyond size alone, critically influence fine-tuning outcomes. When adapting FMs, class balance must therefore be carefully considered, as intrinsic model biases can lead to misleadingly poor results if not properly accounted for.

A similar pattern emerges when comparing NY\textsubscript{small} and NY\textsubscript{all} (Figures~\ref{fig:FM_all_results_by_dataset}.e and~\ref{fig:FM_all_results_by_dataset}.f), which represent the same task but differ in dataset size by an order of magnitude. 
Several fine-tuning techniques remain near $\sim10\%$ mean of MCC score, but DINOv2$_s$ and \textit{DINOv2$_b$} achieve the best results with LoRA, BitFit, and FFT on NY\textsubscript{all}, particularly for DINOv2 and CLIP-Large. 
In contrast, only LoRA performs well on NY\textsubscript{small}. 
Interestingly, FFT achieves the highest performance on NY\textsubscript{small}, mirroring the behavior observed in AFC$_m$. However, on NY\textsubscript{all}, its performance converges with that of PEFT methods that adapt fewer parameters, suggesting that larger datasets enhance representation adaptation, reducing the need for extensive parameter updates.

Fine-tuning effectiveness is strongly influenced by the pretrained model’s intrinsic biases. 
In general, FFT achieves the best performance for PubMedCLIP and DINOv2 variants, indicating that for efficient adaptation, generalist models like DINOv2, despite lacking domain-specific pretraining, can transfer effectively across diverse tasks. 
Conversely, biomedical-specific models such as PubMedCLIP benefit most from FFT, emphasizing that optimal fine-tuning strategies depend jointly on model architecture and dataset properties.

Importantly, all performance comparisons reported here are based on average MCC values computed across folds, as shown in Tables~\ref{tab:results_MCC} and \ref{tab:results_AUC}. 
However, the associated standard deviations remain high across most settings, reflecting the considerable variability intrinsic to low-data regimes. 
This elevated variance highlights the stochastic nature of fine-tuning under constrained sample conditions, reinforcing that the reported trends represent general tendencies rather than absolute performance guarantees.

As additional analysis, the overall statistical comparison reported in Table~\ref{tab:FM_stat} provides a quantitative confirmation of the observed trends. FFT, LP, and BitFit outperform many of the LoRA- and VeRA-based approaches, with several differences reaching statistical significance ($p<0.05$). 
This outcome further supports the conclusion that both the most expressive (FFT, LP) and the most lightweight (BitFit) strategies offer superior adaptability across heterogeneous clinical tasks.
Conversely, intermediate PEFT configurations (e.g., LoRA and VeRA) exhibit lower robustness, suggesting that neither extreme parameter efficiency nor unrestricted fine-tuning alone explains performance; instead, effective adaptation depends on the capacity of the method to align with the pretrained model’s representational structure and the target dataset’s statistical properties. 
Combined with the high variance observed across folds, these findings underscore the complex, data-dependent trade-off that governs PEFT effectiveness in foundation models for medical imaging.
\bb
Given these results, in the following sections we evaluate PEFT techniques under FSL conditions. 
These scenarios are particularly relevant in real-world clinical practice, where annotations are limited and new tasks must be adapted quickly with minimal supervision. 
By isolating PEFT in this extreme regime, we aim to better understand their practical utility and identify configurations that offer robust generalization despite severe training constraints.
\subsection{Can PEFT strategies offer robust and generalizable adaptation in Few-Shot Learning scenarios involving FMs, under prognosis data constraints?}\label{Q:FSL}

In real-world medical imaging applications, particularly in prognosis prediction, annotated datasets are often extremely limited. 
As a result, models are frequently required to adapt from only a few training samples per class. 
This constraint defines FSL setting, a particularly challenging yet clinically realistic scenario. In such contexts, large-scale full fine-tuning is impractical and prone to overfitting, motivating the need for lightweight and efficient adaptation strategies.
PEFT techniques have emerged as a promising direction to address these challenges. However, their effectiveness in extreme low-resource regimes, particularly when applied to high-capacity FMs, remains poorly understood. 
For this reason, in the following sections we discuss the analysis on the performance obtained from FMs on our COVID-19 prognostic datasets. 

\begin{figure}[!ht]
    \centering
    \resizebox{\textwidth}{!}{
    \begin{tikzpicture}
        \node[label={[font=\Large, yshift=.5cm]below: (a) AFC$_m$}] (A) at (0,0) {\includegraphics[width=\textwidth]{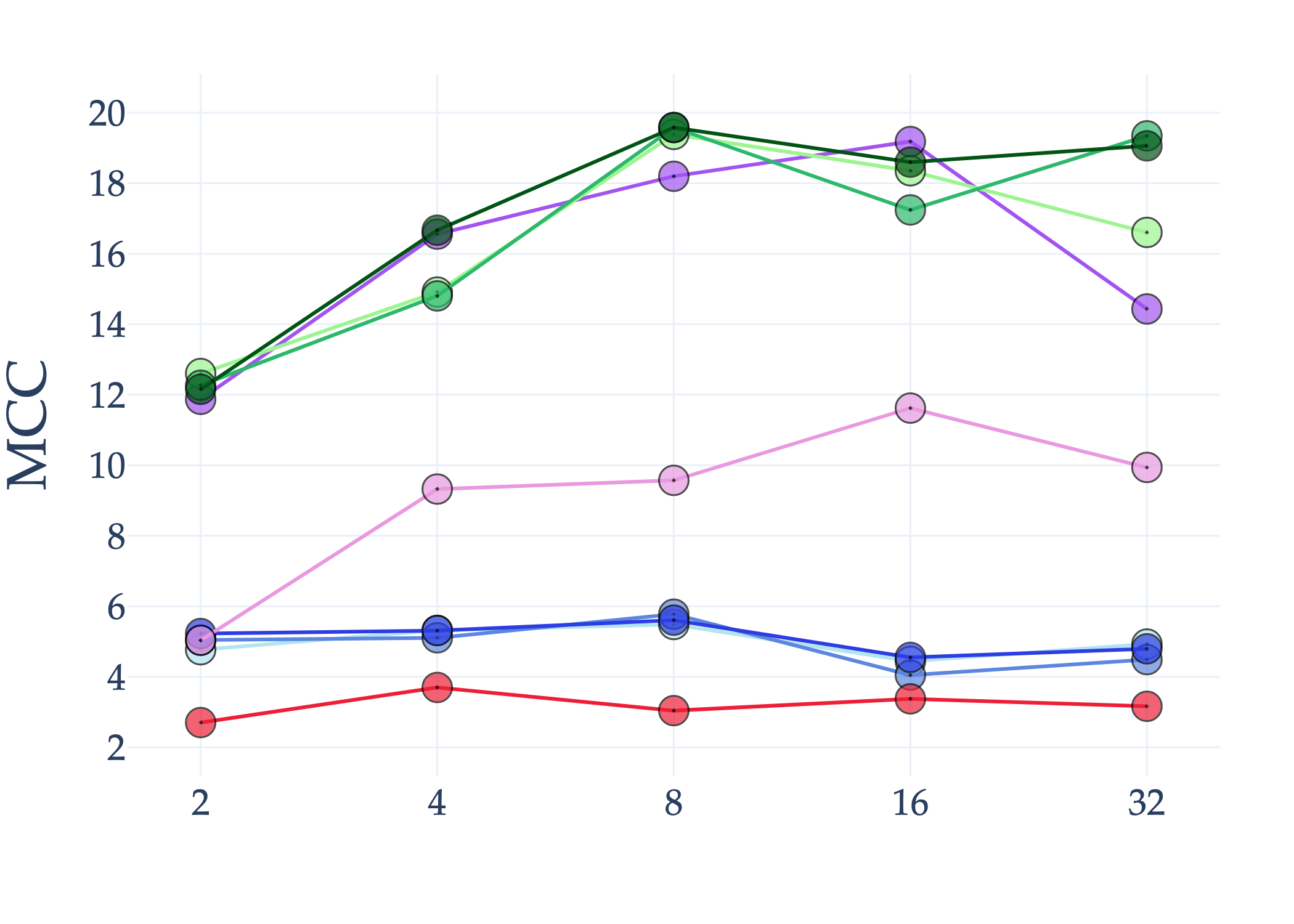}};
        \node[label={[font=\Large, yshift=.5cm]below: (b) AFC}] (B) [right=0.5cm of A] {\includegraphics[width=\textwidth]{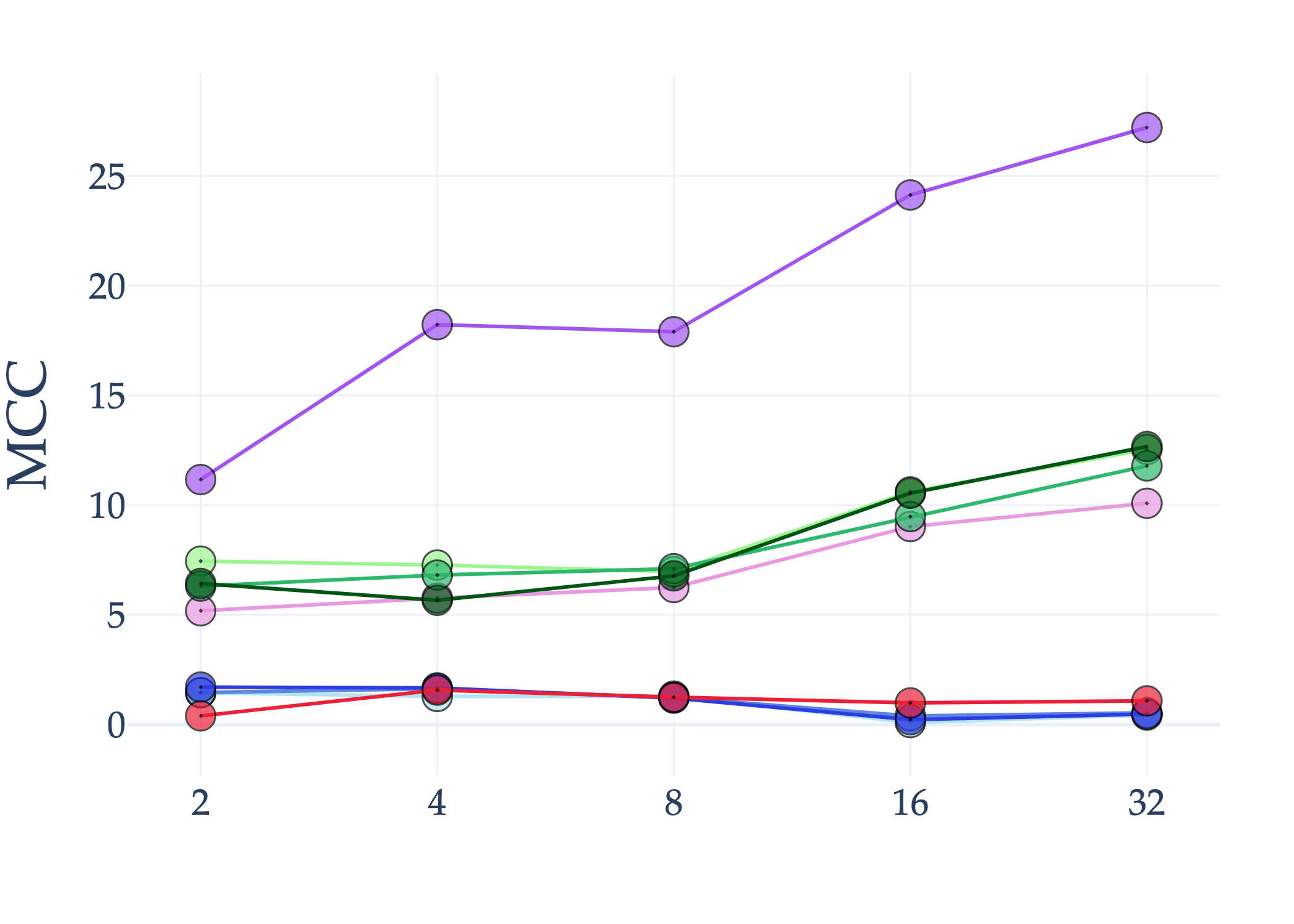}};
        \node[label={[font=\Large, yshift=.5cm]below: (c) CAR}] (C) [below=0.2cm of A] {\includegraphics[width=\textwidth]{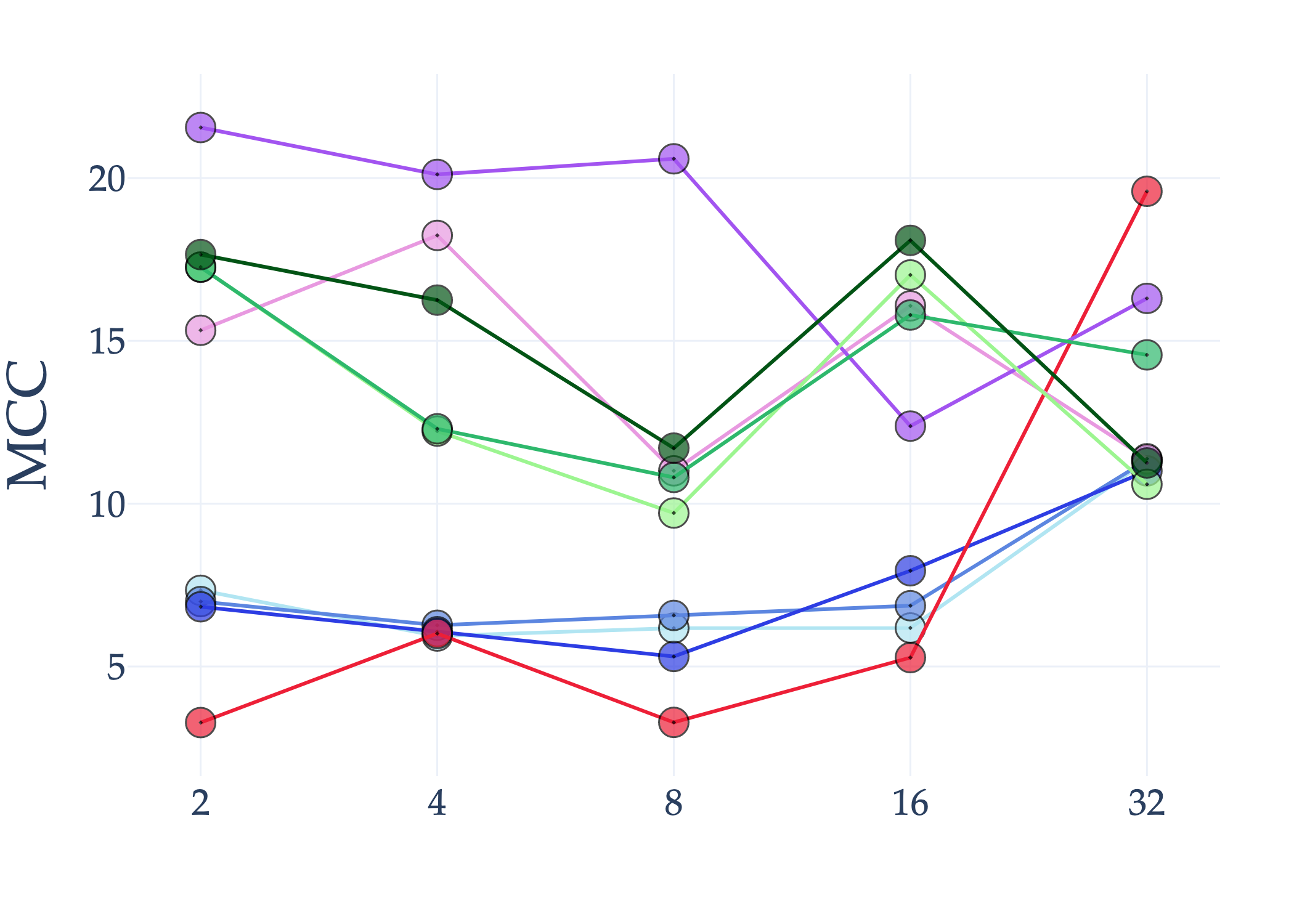}};
        \node[label={[font=\Large, yshift=.5cm]below: (d) CC}] (D) [right=0.5cm of C] {\includegraphics[width=\textwidth]{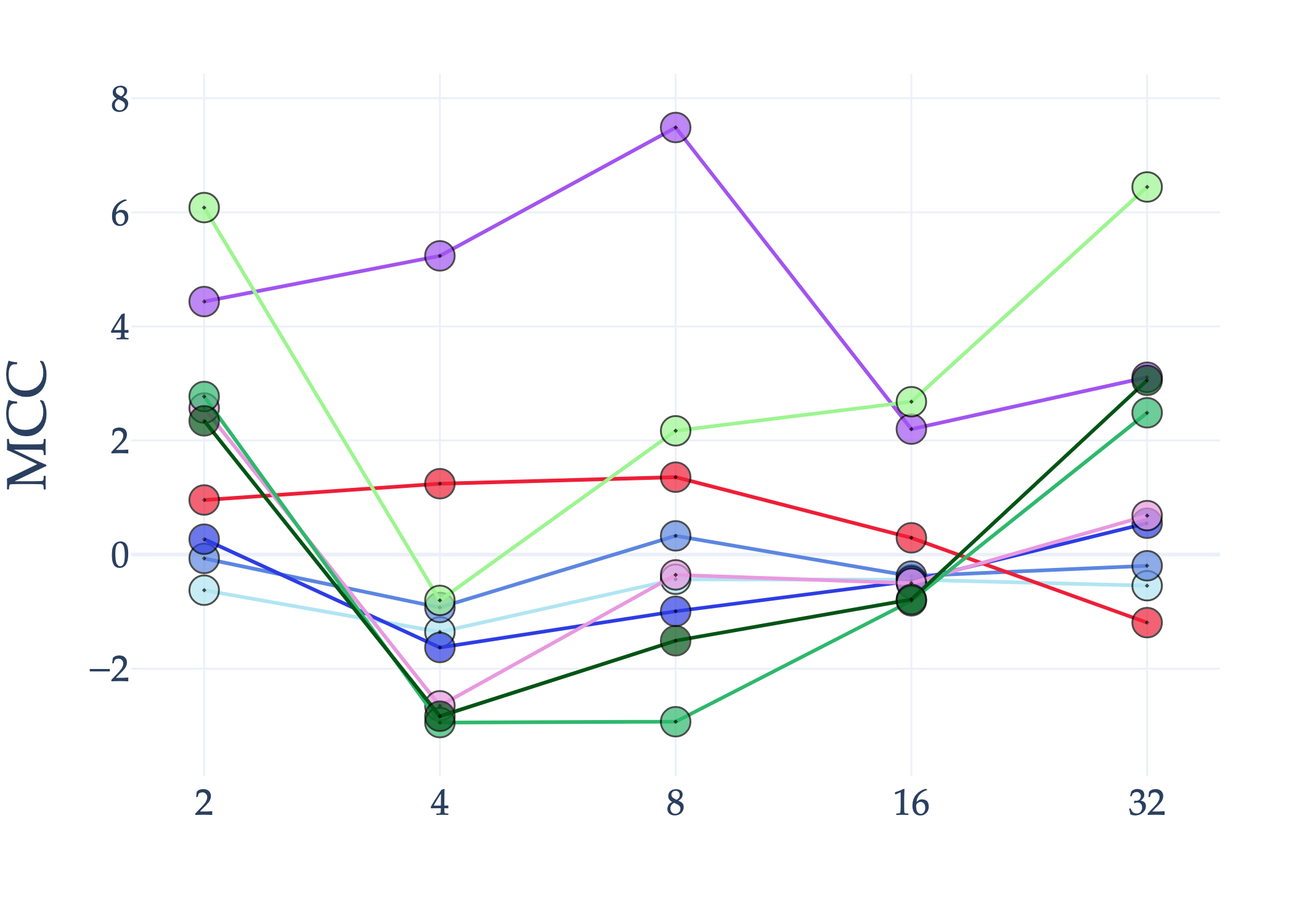}};
        \node[label={[font=\Large, yshift=.5cm]below: (e) NY\textsubscript{small}}] (E) [below=0.2cm of C] {\includegraphics[width=\textwidth]{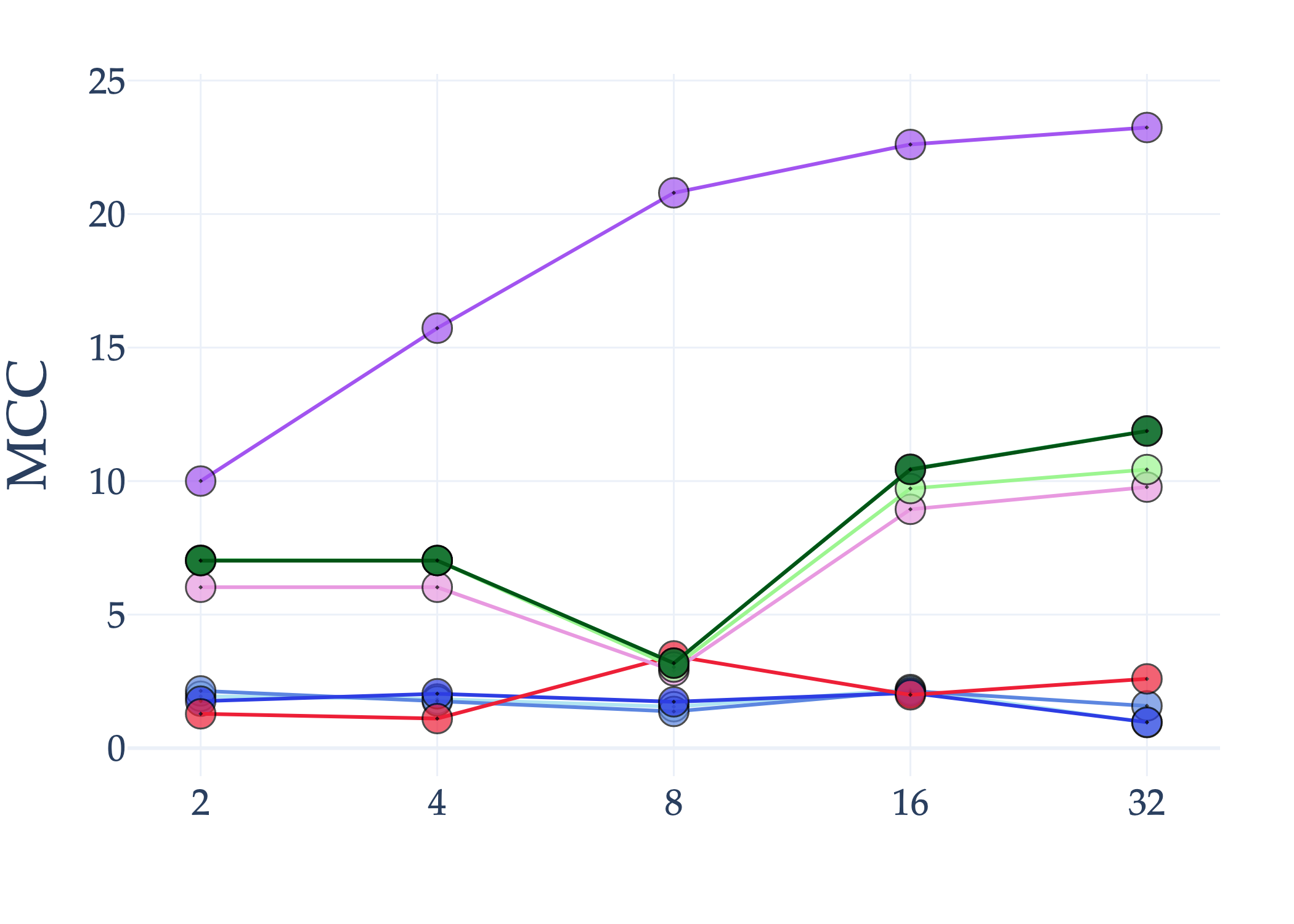}};
        \node[label={[font=\Large, yshift=.5cm]below: (f) \textit{Overall}}] (F) [right=0.5cm of E] {\includegraphics[width=\textwidth]{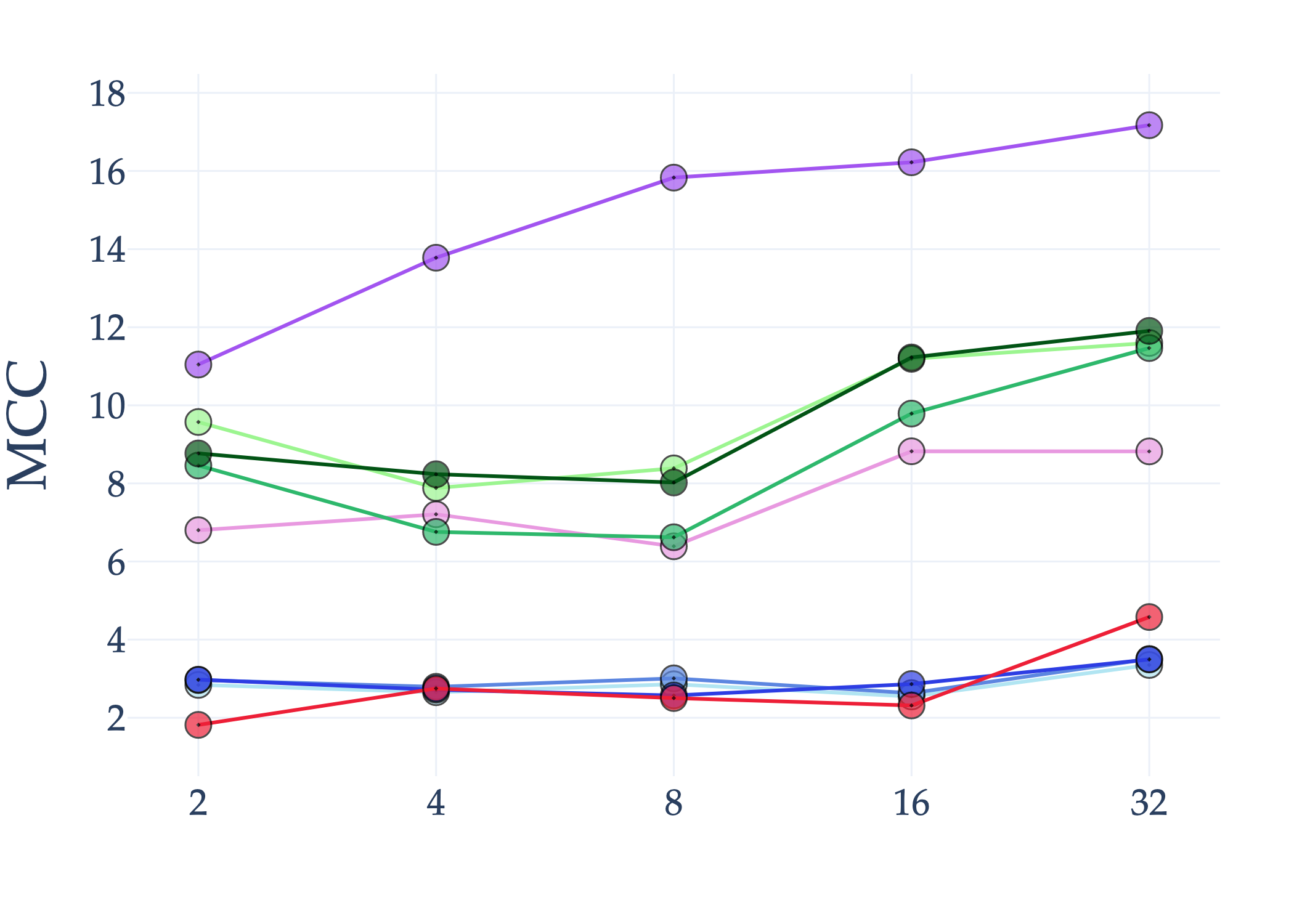}};
    \end{tikzpicture}
    }
    \caption{This figure illustrates the average performance of PEFT techniques across different datasets. The x-axis represents the number of shots (examples) per class, while the y-axis reports the corresponding performance metrics. The final subplot provides an aggregated view, displaying the overall average performance of each technique across all datasets for a more comprehensive comparison. PEFT techniques have a single encoding system based on colors: \textcolor[HTML]{b1e5f2}{$\mathbf{LoRA_{r=4}}$}, \textcolor[HTML]{5c86e0}{$\mathbf{LoRA_{r=8}}$}, \textcolor[HTML]{2d3de3}{$\mathbf{LoRA_{r=16}}$},
    \textcolor[HTML]{9cf590}{$\mathbf{VeRA_{r=4}}$}, \textcolor[HTML]{2eb86c}{$\mathbf{VeRA_{r=8}}$}, \textcolor[HTML]{025415}{$\mathbf{VeRA_{r=16}}$}, \textcolor[HTML]{ed1f37}{$\mathbf{BitFit}$}, \textcolor[HTML]{e899e0}{$\mathbf{IA^3}$}, \textcolor[HTML]{a254f0}{$\mathbf{LP}$}.
    }
    \label{fig:FSL_average_PEFT_by_dataset}
\end{figure}

\subsubsection{Setup.} We also analyzed the efficacy of PEFT in low-resource settings, following the few-shot learning approach described in Section~\ref{sec:FSL}. 
All results are averaged across FM architectures, so the final plots represent the overall behavior of each technique on a specific dataset. We included all the datasets listed in Table~\ref{tab:dataset-table}, except for NY\textsubscript{all}. 
Since the task remains the same while the sample size is reduced, we only evaluate performance on NY\textsubscript{small}. 
Given the FSL setting, the number of training samples is identical for both datasets, meaning the additional samples in NY\textsubscript{all} do not impact the training phase. 
Therefore, we focus on NY\textsubscript{small} for evaluation, as it provides the most relevant insights under the constrained training conditions. 
The performance used for evaluation remains consistent with previous experimental setups, as the test sets are the same across validation strategies. 
The only difference is the training sample size, which is deliberately limited in the FSL setting as described in Section~\ref{sec:FSL}. 
Figure~\ref{fig:FSL_average_PEFT_by_dataset} presents the average performances obtained, where the x-axis represents the number of shots (examples) per class, and the y-axis reports performance metrics. 
The last subplot also shows the overall average performance of each technique across all datasets, providing a comprehensive overview of the results.

\subsubsection{Results.} We analyze the results presented in Figure~\ref{fig:FSL_average_PEFT_by_dataset}. The first key observation is that, across all datasets, performance remains substantially lower than expected.
However, an interesting pattern emerges when comparing the AFC dataset in its balanced and imbalanced versions (Figures~\ref{fig:FSL_average_PEFT_by_dataset}.a and~\ref{fig:FSL_average_PEFT_by_dataset}.b). 
In AFC$_m$, the best results were obtained with VeRA, one of the most parameter-efficient approaches, suggesting that efficiency plays a role in this specific setting. 
However, the overall performance across all datasets appears highly inconsistent, showing no clear evidence of a robust and repeatable trend. In AFC and NY\textsubscript{small} (Figure~\ref{fig:FSL_average_PEFT_by_dataset}.e), LP achieves the best absolute performance, contrasting with the results from AFC, where LP alone performed best.
This discrepancy may be linked to the parameter space, as the number of available samples is highly constrained. It is likely that the limited dataset size affects the model’s ability to effectively predict the task. Similar trends were observed in the fine-tuning settings explored in Section~\ref{Q:FM}, where most methodologies failed to achieve satisfactory performance. 
Several factors could contribute to this, including the adaptability of foundation models, the efficacy of fine-tuning methods, or the complexity of the tasks themselves.

Given these observations, we strongly believe that under prognosis-related data settings, achieving robust performance remains challenging. FMs, despite their strong representation capabilities, struggle in FSL settings due to their high parametrization and reliance on large-scale data for effective adaptation. 
The inherent complexity of prognosis tasks further amplify these challenges, as subtle visual cues correlated with clinical outcomes require a fine balance between feature extraction and model adaptability. 
The major issue is that achieving a sufficiently generalizable network in FSL settings is highly difficult when the test set maintains the natural class imbalance inherent in the dataset distribution, while the training sets are artificially rebalanced using the geometric progression adopted for this experimental setup (Section~\ref{sec:FSL}).

Nevertheless, LP demonstrates to be the most promising approach in extreme low-resource conditions (Figure~\ref{fig:FSL_average_PEFT_by_dataset}.f). 
Additionally, even with stronger imbalances, LP performs well when the number of samples is sufficient, as seen in NY\textsubscript{small} and AFC$_m$. This suggests that, while FMs struggle in highly constrained data regimes, certain fine-tuning strategies can still enable meaningful adaptation.

A critical statistical consideration arises from the imbalance between training and test set distributions in our FSL setup. 
Specifically, while training subsets were artificially rebalanced using the geometric progression described in Section~\ref{sec:FSL} to ensure that all classes were represented during adaptation, the test sets preserved their natural, imbalanced class distributions. This mismatch introduces a potential bias: models may appear to generalize effectively on the balanced training splits but struggle to maintain stable performance when evaluated on the naturally skewed test data. 
Consequently, the apparent improvements observed during training might overestimate generalization capability, especially for methods like VeRA or LoRA that rely on compact parameter adaptation and can overfit balanced subsets.

\section{Conclusions}\label{sec:Conclusion}
\bl
In this work, we conducted a comprehensive benchmark to assess the effectiveness of fine-tuning strategies, ranging from FFT and LP to several state-of-the-art PEFT methods, across a diverse set of models, including both CNNs and FMs. 
Focusing on COVID-19 prognosis tasks derived from publicly available chest X-ray datasets, we systematically explored how different fine-tuning approaches perform under varying data constraints, with particular attention to dataset size, class imbalance, and FSL scenarios.

Our findings lead to several key conclusions:

\begin{itemize}

\item \textbf{CNNs Remain Reliable in Extreme Low-Data Regimes:}
CNNs, especially when fine-tuned with FFT, consistently outperform FMs on very small or imbalanced datasets. Their compact architecture and strong inductive biases reduce overfitting, making them suitable for data-scarce clinical contexts such as rare disease modelling or emergency outbreak response.

\item \textbf{FMs with PEFT Excel as Data Availability Increases:}
While FMs struggle in extreme low-data settings, they demonstrate strong scalability as dataset size grows. PEFT methods such as LoRA and BitFit enable efficient adaptation of large pretrained models with minimal parameter updates, providing competitive performance at a fraction of the computational cost of FFT.
These characteristics make them ideal for established clinical pipelines with sufficient annotated data, where deployment efficiency and model reuse are critical.

\item \textbf{PEFT Methods Are Sensitive to Class Imbalance:}
Across experiments, the performance of PEFT techniques declines under severe class imbalance, especially when minority classes correspond to critical outcomes (e.g., ICU admission, mortality). FFT remains more stable in these cases, though moderate re-balancing restores PEFT competitiveness, suggesting potential for adaptive sampling or hybrid fine-tuning schemes.

\item \textbf{Selective PEFT Configurations Show Promise in FSL:}
In few-shot settings, smaller FM variants (e.g., DINOv2$_s$) combined with PEFT methods, particularly LP and BitFit, can extract meaningful representations even with limited supervision.
LP, in particular, offers stable and computationally inexpensive adaptation, making it a pragmatic option for rapidly evolving or low-annotation scenarios such as rare disease outbreaks or early-stage clinical trials.

\item \textbf{No Single Fine-Tuning Strategy is Universally Optimal:}
The effectiveness of a fine-tuning approach depends on the interplay between model architecture, dataset scale, and task complexity. CNNs remain robust and interpretable for low-resource scenarios, whereas FMs combined with PEFT techniques offer scalable and efficient solutions in data-rich clinical contexts.
\end{itemize}

\paragraph{Study Strengths}
Overall, our results indicate that the success of efficient methods such as LoRA and BitFit is driven not only by parameter efficiency but also by their alignment with the model’s pretrained feature hierarchy and representational quality. These factors govern the adaptability of each method to specific tasks, as observed in the efficiency–effectiveness trade-offs of Figures~\ref{fig:CNN_vs_FM}.

These findings also translate into actionable insights for real-world deployment. The stable performance of LP in few-shot settings for FMs provides a practical solution for emerging rare diseases, where rapid adaptation with minimal labeled data is essential. 
Similarly, CNNs remain the preferred option in extreme low-data or highly imbalanced conditions, such as early-stage outbreak response or localized clinical studies with limited annotation resources. 
Conversely, FMs fine-tuned with LoRA or BitFit are best suited for established clinical workflows, where larger and better-balanced datasets enable efficient model reuse and scalable deployment. 
Together, these results offer a road map for selecting fine-tuning strategies according to data availability, clinical context, and computational constraints, enhancing the translational impact of AI in medical prognosis.

\paragraph{Limitations.}
This study presents several limitations that should be acknowledged.
First, it focuses exclusively on COVID-19 chest X-ray prognosis tasks, which, while clinically relevant, may not generalize to other imaging modalities or disease contexts. Chest X-rays inherently provide limited soft-tissue contrast and spatial detail compared to modalities such as CT or MRI, constraining their capacity to capture early or subtle prognostic markers.

Second, the analysis is restricted to single-image features, without explicitly modeling the temporal evolution of disease. Although some datasets (e.g., NY\textsubscript{all}, Co-Cross) contain longitudinal chest X-rays collected during hospitalization, our current methodology does not integrate temporal information. 
Future work should therefore explore temporal and multimodal fusion strategies to better represent disease progression dynamics and improve prognostic accuracy.

Third, despite their growing promise, FMs still exhibit substantial challenges in data-constrained medical scenarios. Their large parameterization and reliance on abundant supervision limit their effectiveness under data scarcity and severe class imbalance, conditions typical of real-world clinical datasets. While PEFT strategies mitigate these issues to some extent, our results indicate that FMs remain less robust than CNNs in extreme low-data regimes, highlighting the need for more adaptive and data-efficient FM adaptation methods.

Fourth, the anticipated benefits of biomedical-specific pretraining were not consistently observed. Generalist FMs such as DINOv2 often performed comparably, or even better, than biomedical-pretrained counterparts like PubMedCLIP and MedCLIP$_c$. This suggests that current biomedical FMs may not yet fully exploit domain knowledge for prognosis tasks, or that existing PEFT mechanisms do not optimally leverage such representations.

Finally, although we evaluated a broad spectrum of fine-tuning strategies, the search space remains limited in terms of PEFT configurations and hyperparameter exploration.
Nevertheless, the breadth and standardization of our benchmark provide a consistent and informative basis for assessing fine-tuning strategies across heterogeneous prognosis tasks, while leaving the integration of temporal modeling, multimodal data, and domain-specific foundation models to future research.
\bb

\subsubsection{Acknowledgments}
Filippo Ruffini is a PhD student enrolled in the National PhD in Artificial Intelligence, XXXVIII cycle, course on Health and life sciences, organized by Università Campus Bio-Medico di Roma. 
This work was partially supported by: i) the Italian Ministry of Foreign Affairs and International Cooperation, grant number PGR01156, ii) PNRR MUR project PE0000013-FAIR, iii) PNRR – DM 118/2023.
Resources are provided by the \textit{National Academic Infrastructure for Supercomputing in Sweden} (NAISS) and the \textit{Swedish National Infrastructure for Computing} (SNIC) at Alvis @ C3SE, partially funded by the Swedish Research Council through grant agreements no. 2022-06725 and no. 2018-05973.

\bl
\bibliographystyle{elsarticle-num-names} 
\bibliography{main}

\appendix
\section{Mathematical Formulation of PEFT Methods}\label{app:math_peft}

This appendix presents the mathematical foundations of the PEFT strategies evaluated in this study. Each method is formalized with its update rule and adaptation mechanism applied to a pretrained weight matrix defined in the following mathematical formulation as $W_0 \in \mathbb{R}^{n\times n}$.

\subsection{Low-Rank Adaptation (LoRA)}

LoRA~\cite{hu2021lora} assumes that the optimal weight update lies in a low-dimensional subspace. Given a $W_0$, the adaptation introduces two trainable low-rank matrices $A \in \mathbb{R}^{r\times n}$ and $B \in \mathbb{R}^{n\times r}$, with $r \ll n$. The modified layer output for an input $x$ is:
\begin{equation}
h = (W_0 + BA)x = W_0x + BAx .
\label{eq:lora_app}
\end{equation}
Here, $h \in \mathbb{R}^{n}$ denotes the adapted feature representation of the input $x \in \mathbb{R}^{n}$, $A,B$ are low-rank matrices whose product $\Delta W = B A$ represents the task-specific update. By restricting updates to a rank-$r$ subspace, LoRA preserves the pretrained knowledge while enabling efficient adaptation, particularly advantageous in low-data medical imaging settings.

\subsection{Vector-based Random Matrix Adaptation (VeRA)}

VeRA~\cite{kopiczko2023vera} extends LoRA by fixing the low-rank matrices $A$ and $B$ as random projections shared across layers, and learning only two scaling vectors $b,d \in \mathbb{R}^{1\times r}$ represented as diagonal matrices $\Lambda_b$ and $\Lambda_d$. The resulting output is:
\begin{equation}
h = W_0x + \Lambda_b B \Lambda_d A x .
\label{eq:vera_app}
\end{equation}
Where $\Lambda_b, \Lambda_d \in \mathbb{R}^{r\times r}$ are diagonal matrices whose entries are the elements of the scaling vectors $b$ and $d$, respectively. The terms $A$ and $B$ denotes the fixed random projection matrices, while $\Lambda_b$ and $\Lambda_d$ control its magnitude and direction across channels. This formulation drastically reduces the number of trainable parameters while maintaining flexibility. 

\subsection{Bias-Only Fine-Tuning (BitFit)}

BitFit~\cite{zaken2021bitfit} restricts training to the bias terms of a pretrained network $f(\theta_0,x)$ with parameters $\theta_0=\{W_i,b_i\}_{i=1}^N$. During adaptation, only $\theta_b=\{b_i\}_{i=1}^N$ are optimized:
\begin{equation}
\theta^\ast = \arg\min_{\theta_b} \mathcal{L}(f(\theta_b;x),y)
\quad\text{s.t.}\quad W_i=\text{const},\ \forall i \in \{1,...,N\}.
\label{eq:bitfit_app}
\end{equation}
Where $\mathcal{L}$ is the task-specific loss function, $x$ and $y$ are input–label pairs, and $W_i,b_i$ denote the weight and bias parameters at layer $i$ for every layer of the model $f$. The \textit{BitFit} approach dispose that only the biases $b_i$ are updated during training, while all $W_i$ remain frozen. This strategy typically modifies a small fraction of the total parameters of the model.

\subsection{Infused Adapter by Inhibiting and Amplifying Inner Activations (IA$^{3}$)}

IA$^{3}$~\cite{liu2022-ia3} introduces multiplicative gates that modulate activations in attention and feed-forward layers without changing the any original pretrained weight matrix (i.e, $W_0$). In a transformer block with query, key, and value projections functions $Q=xW_Q$, $K=xW_K$, and $V=xW_V$ for the input data $x$, the adapted attention function in IA$^{3}$ method is defined as:
\begin{equation}
\text{Attn}(Q,K,V)
= \text{softmax}\!\left(\frac{Q(\boldsymbol{\ell}_k\odot K)^{\top}}{\sqrt{d_k}}\right)
(\boldsymbol{\ell}_v\odot V),
\label{eq:ia3_attn}
\end{equation}
where $W_Q, W_K, W_V \in \mathbb{R}^{d\times d_k}$ are the projection matrices, $d_k$ is the key dimension, $\boldsymbol{\ell}_k,\boldsymbol{\ell}_v\in\mathbb{R}^{d_k}$ are trainable scaling vectors, and $\odot$ denotes element-wise multiplication.  

For the feed-forward block with weights $W_1\in\mathbb{R}^{d\times d_{\text{ff}}}$ and $W_2\in\mathbb{R}^{d_{\text{ff}}\times d}$, the adapted formulation is:
\begin{equation}
\text{FFN}(x)=W_2\!\left(\boldsymbol{\ell}_{\mathrm{ff}}\odot\sigma(W_1x)\right),
\label{eq:ia3_ffn}
\end{equation}
where $\sigma(\cdot)$ is a non-linear activation (e.g., GELU) and $\boldsymbol{\ell}_{\mathrm{ff}}\in\mathbb{R}^{d_{\mathrm{ff}}}$ is a learned scaling vector applied channel-wise.  
All scaling vectors are initialized to ones, ensuring that the pretrained behavior of the model is preserved at initialization.

\subsection{Full Fine-Tuning (FFT) and Linear Probing (LP)}

For completeness, two reference strategies are considered.  
Full Fine-Tuning (FFT) updates all model parameters:
\begin{equation}
\theta^\ast = \arg\min_{\theta} \mathcal{L}(f(\theta;x),y),
\end{equation}
where $\theta$ denotes all trainable parameters of the model $f(\cdot)$.  
Linear Probing (LP) freezes the backbone and trains only a linear classifier $g_\phi$:
\begin{equation}
\phi^\ast = \arg\min_{\phi} \mathcal{L}(g_\phi(f(x)),y),
\end{equation}
where $\phi$ represents the parameters of the classification head and $f(x)$ is the frozen feature extractor output.  
FFT represents the upper bound of adaptability, whereas LP serves as a lower bound that evaluates the separability of pretrained representations.

\section{Experiments Plot with $95\%$ confidence interval}
\begin{figure*}[ht]
    \centering
    
    \includegraphics[width=\textwidth]{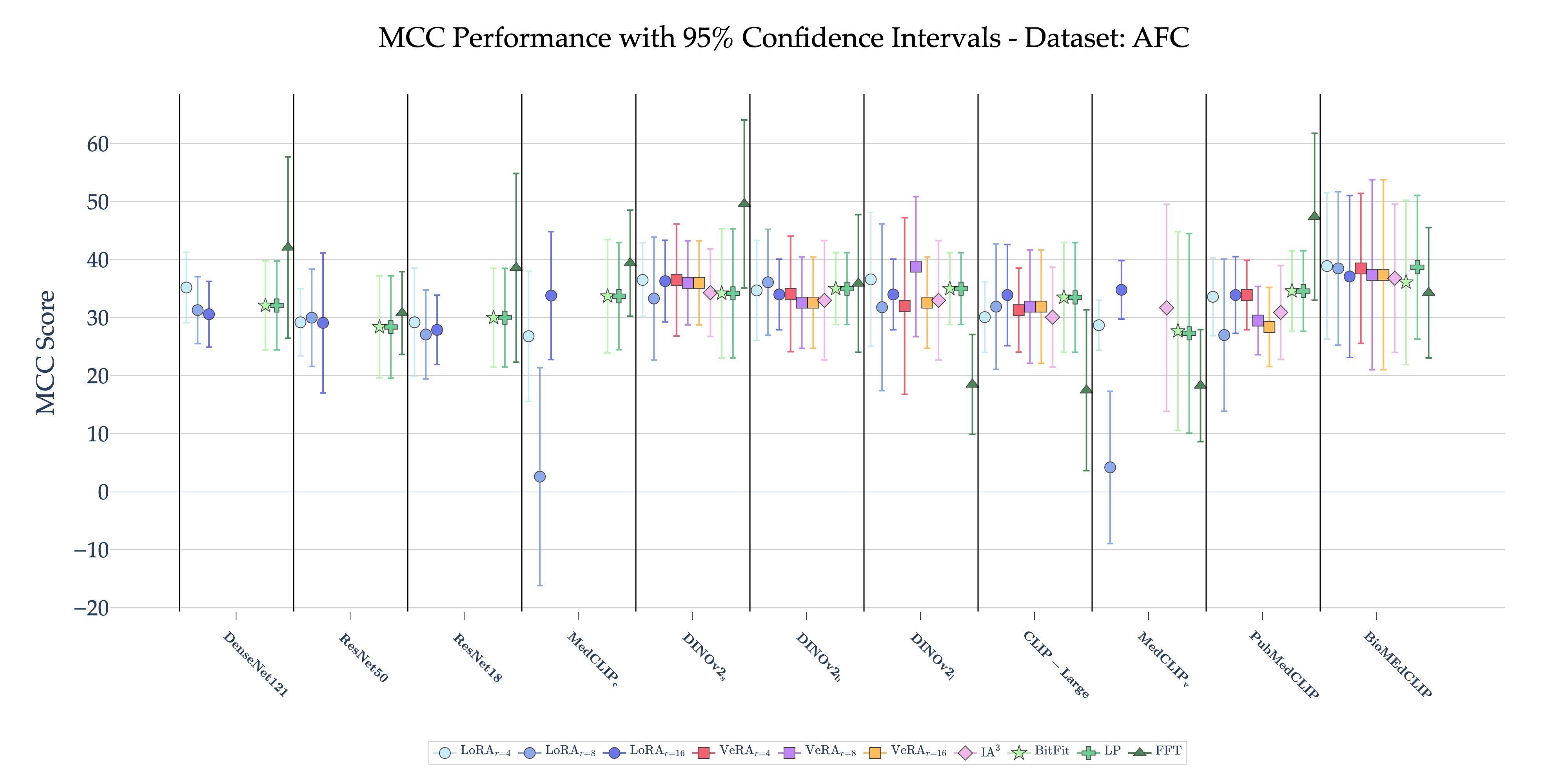}
    \caption{
        \textbf{MCC performance with 95\% confidence intervals. Task: AFC.}
        Each point represents the Matthews correlation coefficient (MCC) obtained for a fine-tuning method across foundation models.
        Error bars indicate 95\% confidence intervals.
    }
    \label{fig:mcc_ci_afc}

\end{figure*}

\begin{figure*}[htbp]
    \centering
    \includegraphics[width=\textwidth]{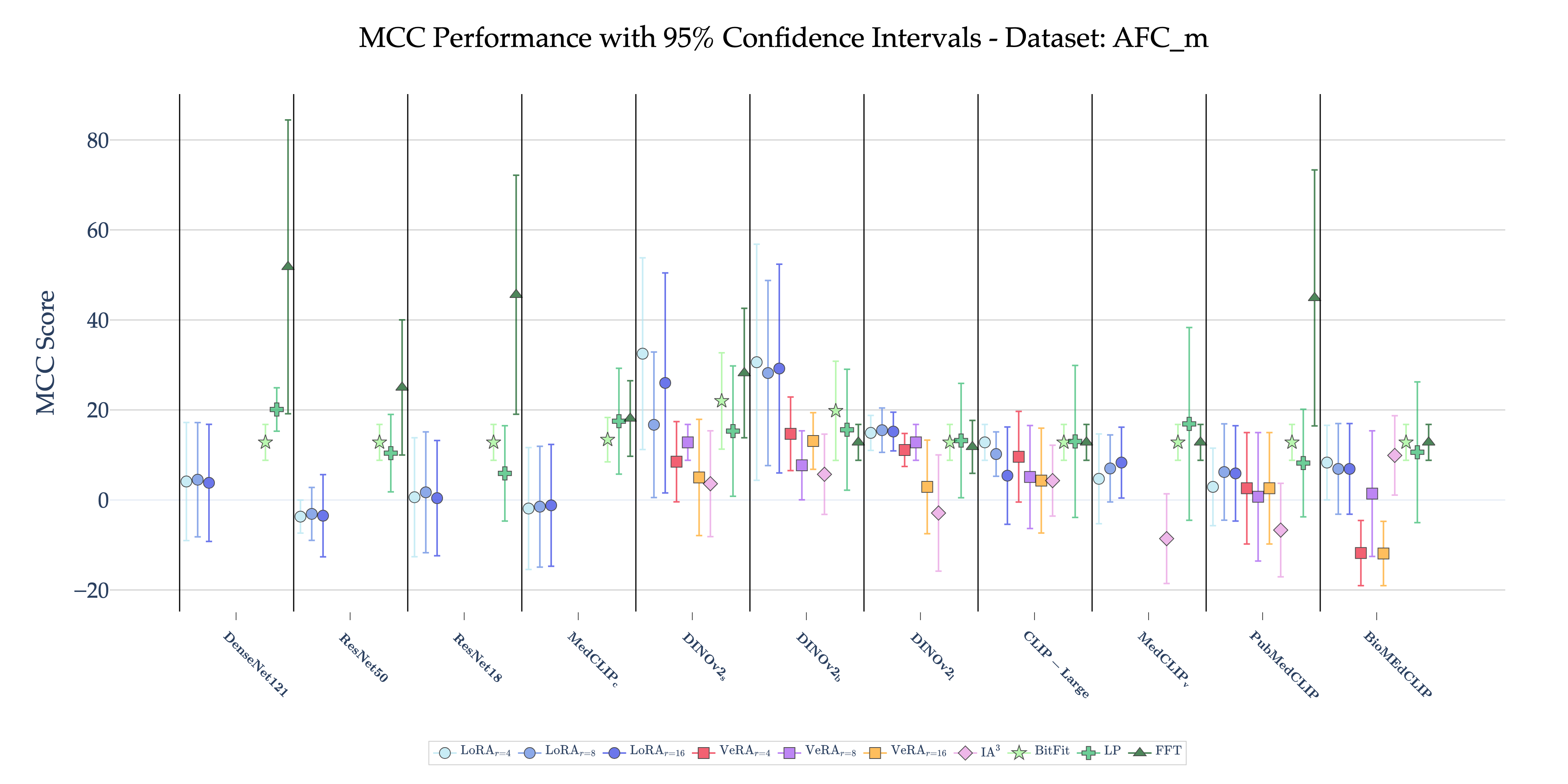}
    \caption{
        \textbf{MCC performance with 95\% confidence intervals. Task: AFC\textsubscript{m}.}
        Comparison of fine-tuning strategies across backbone models.
        Error bars represent 95\% confidence intervals of MCC values across folds.
    \label{fig:mcc_ci_afc_m}
    }
\end{figure*}

\begin{figure*}[htbp]
    \centering
    \includegraphics[width=\textwidth]{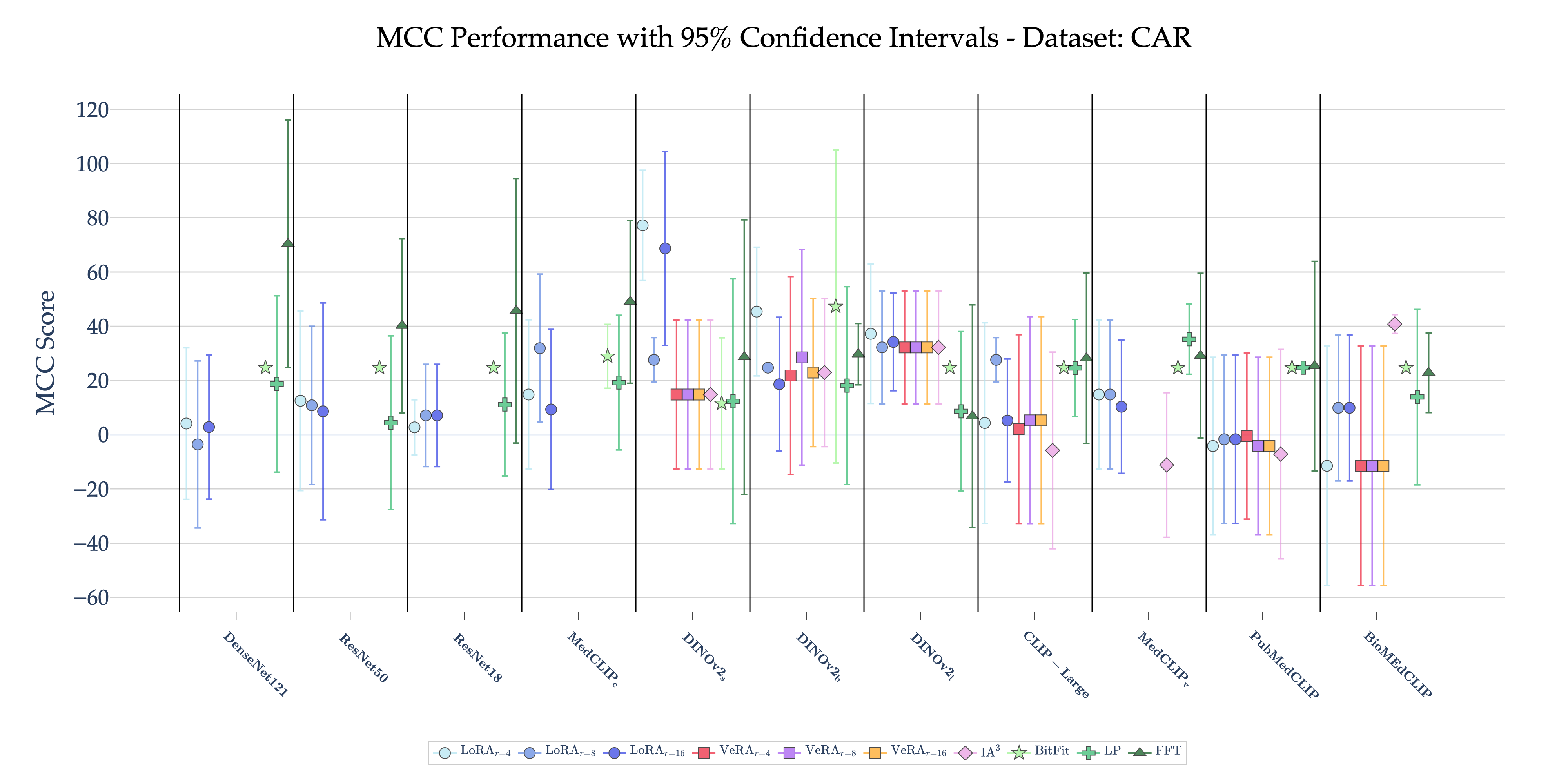}
    \caption{
        \textbf{MCC performance with 95\% confidence intervals. Task: CAR.}
        Performance of different fine-tuning methods across foundation models with 95\% confidence intervals.
    }
    \label{fig:mcc_ci_car}
\end{figure*}

\begin{figure*}[htbp]
    \centering
    \includegraphics[width=\textwidth]{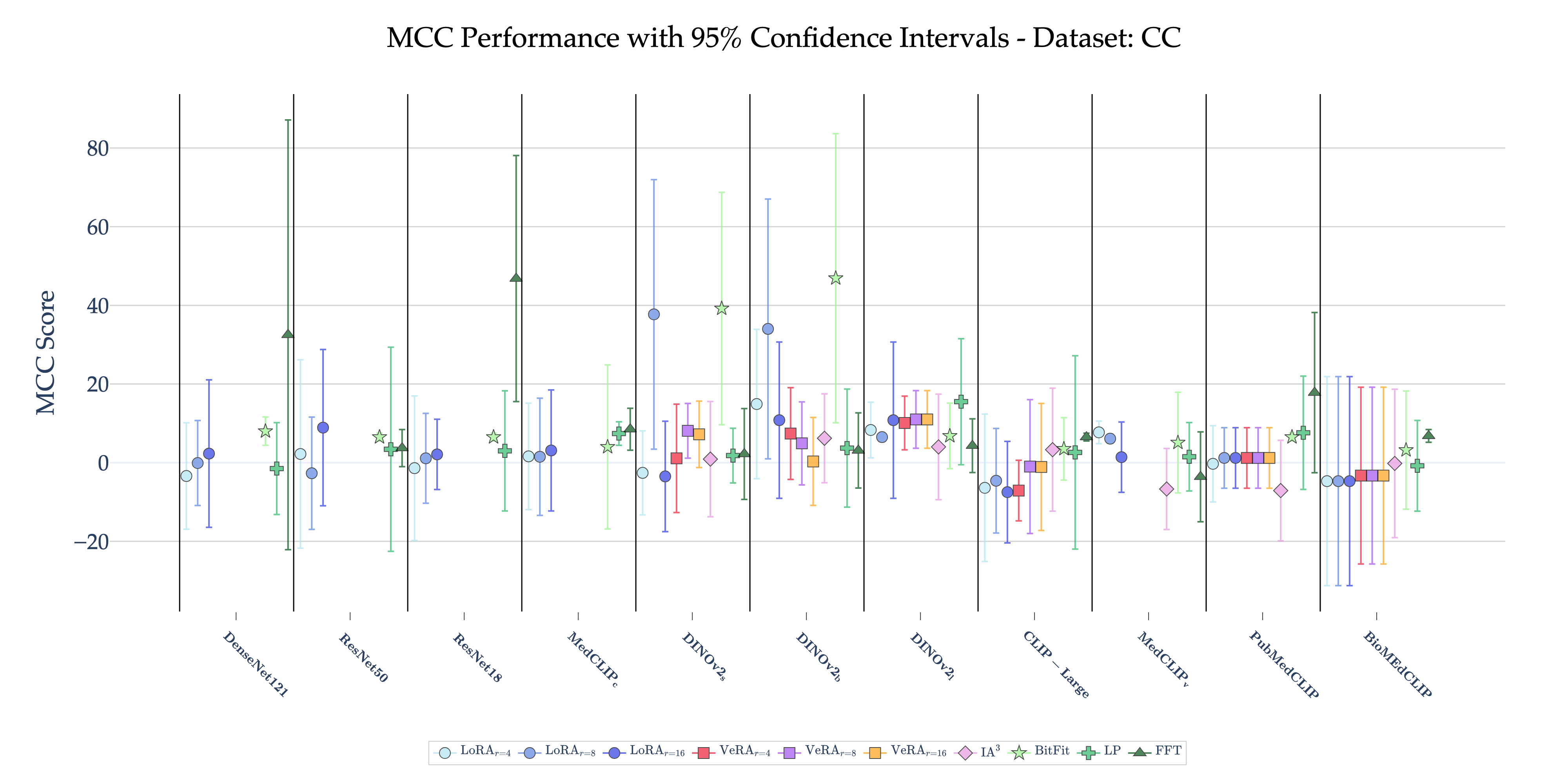}
    \caption{
        \textbf{MCC performance with 95\% confidence intervals. Task: CC.}
        Error bars denote 95\% confidence intervals of the MCC metric for each fine-tuning configuration.
    }
    \label{fig:mcc_ci_cc}
\end{figure*}

\begin{figure*}[htbp]
    \centering
    \includegraphics[width=\textwidth]{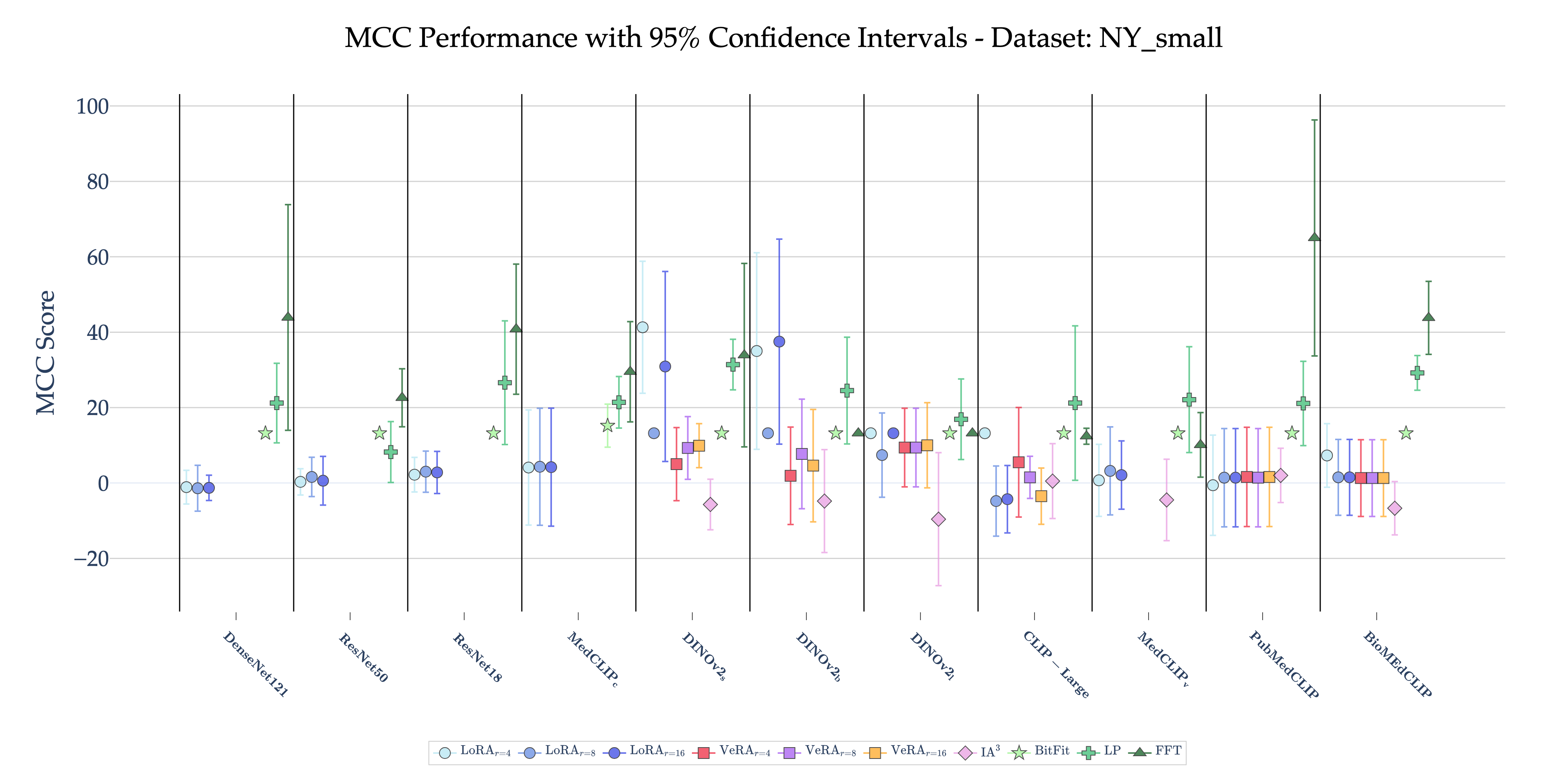}
    \caption{
        \textbf{MCC performance with 95\% confidence intervals. Task: NY\textsubscript{small}.}
        Results obtained for smaller-scale NY dataset showing variation across fine-tuning methods and backbone models.
    }
    \label{fig:mcc_ci_ny_small}
\end{figure*}

\begin{figure*}[htbp]
    \centering
    \includegraphics[width=\textwidth]{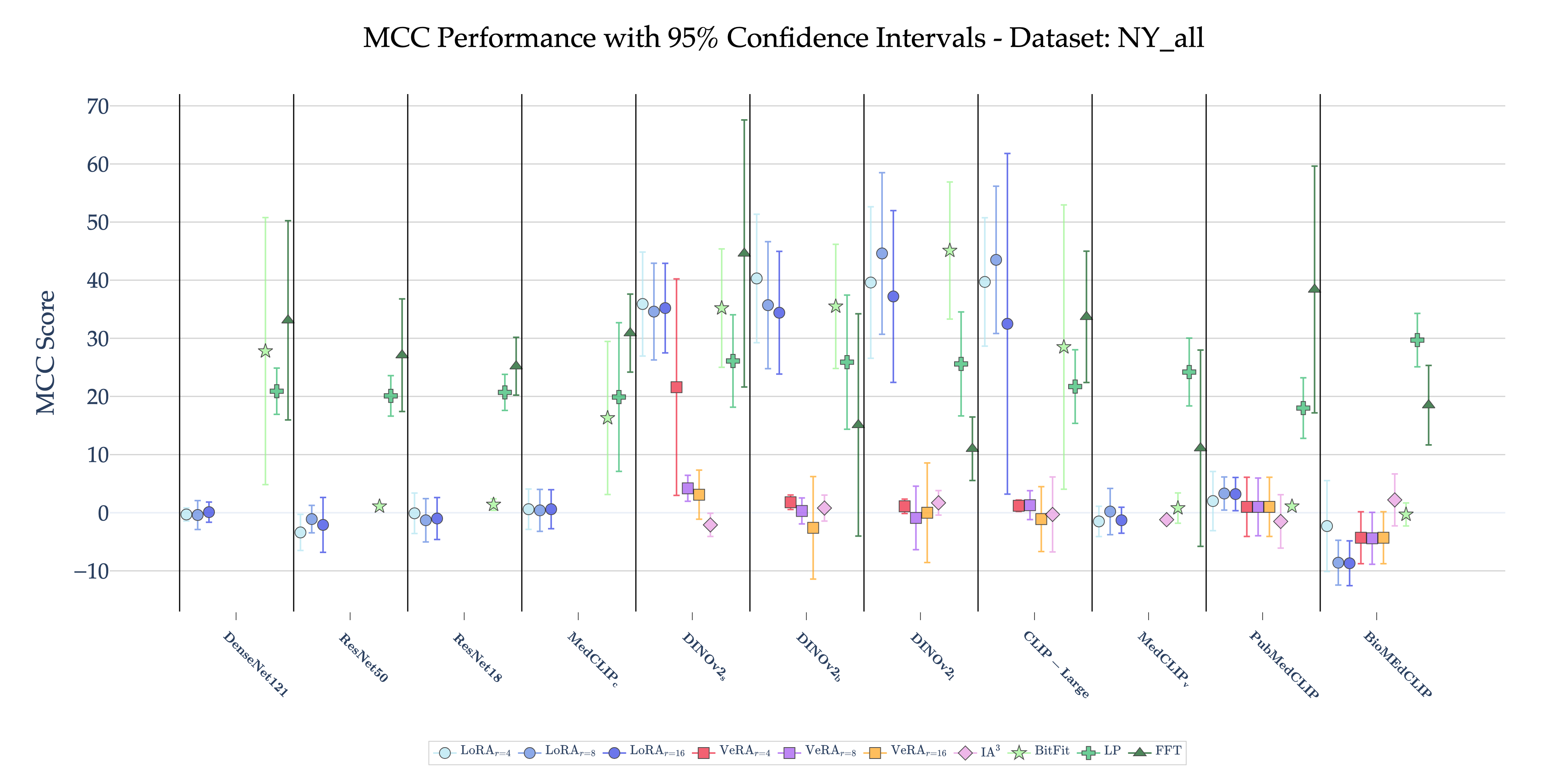}
    \caption{
        \textbf{MCC performance with 95\% confidence intervals. Task: NY\textsubscript{all}.}
        Comparison of fine-tuning methods across backbone architectures for the full NY dataset. 
        Error bars represent 95\% confidence intervals.
    }
    \label{fig:mcc_ci_ny_all}
\end{figure*}

In Figures~\ref{fig:mcc_ci_afc}, ~\ref{fig:mcc_ci_afc_m}, ~\ref{fig:mcc_ci_car}, ~\ref{fig:mcc_ci_cc}, ~\ref{fig:mcc_ci_ny_all}, ~\ref{fig:mcc_ci_ny_small}; we present the detailed experimental MCC results visualized with their corresponding $95\%$ confidence intervals.
Each plot illustrates the variability and statistical reliability of the reported performance metrics across datasets and model configurations.
These visualizations complement the main text by providing a clearer view of model stability and the degree of overlap between methods, thereby supporting a more rigorous comparative analysis.

\section{Model Repositories and Accessibility}
\label{appendix:model_links}
\begin{table}[h]
\centering
\resizebox{\columnwidth}{!}{
\begin{tabular}{l|l}
\toprule
\textbf{Model} & \textbf{Official Repository / Access Link} \\
\midrule
ResNet18/50 & \url{https://pytorch.org/hub/pytorch_vision_resnet} \\
DenseNet121 & \url{https://docs.pytorch.org/vision/main/models/densenet.html} \\
DINOv2$_{s,m,l}$ & \url{https://github.com/facebookresearch/dinov2} \\
CLIP-Large & \url{https://github.com/mlfoundations/open_clip} \\
MedCLIP$_{c,v}$ & \url{https://github.com/RyanWangZf/MedCLIP} \\
BioMedCLIP & \url{https://huggingface.co/microsoft/BiomedCLIP-PubMedBERT_256-vit_base_patch16_224} \\
PubMedCLIP & \url{https://huggingface.co/flaviagiammarino/pubmed-clip-vit-base-patch32} \\
\bottomrule
\end{tabular}}
\caption{Official repositories or Hugging Face links for all pretrained models used in this study.}
\label{tab:foundation_models_links}
\end{table}

For transparency and reproducibility, Table~\ref{tab:foundation_models_links} reports the official repositories, Hugging Face or torch links from which all pretrained model weights are obtained. All listed resources are publicly available and maintained by their respective authors.
\bb

\end{document}

%% file: tables/results_MCC.tex
\begin{table}[!ht]
    \setlength{\extrarowheight}{0.4pt} 
    \renewcommand{\arraystretch}{1.0} 
    \centering
    \resizebox{\columnwidth}{!}{
    \begin{tabular}{l|l|cccccccccc}
        \toprule
        \textbf{Model} & \backslashbox{\textbf{Task}}{\textbf{FT}} & $\mathbf{LoRA_{r=4}}$ & $\mathbf{LoRA_{r=8}}$ & $\mathbf{LoRA_{r=16}}$ & $\mathbf{VeRA_{r=4}}$ & $\mathbf{VeRA_{r=8}}$ & $\mathbf{VeRA_{r=16}}$ & $\mathbf{BitFit}$ & $\mathbf{IA^3}$ & $\mathbf{LP}$ & $\mathbf{FFT}$ \\
        \midrule
\multirow{6}{*}{$\mathbf{ResNet18}$} 
& AFC& 29.2 $\pm$ 8.9 &27.1 $\pm$ 7.3 &27.9 $\pm$ 5.7 &- &- &- &30.0 $\pm$ 8.1 &- &30.0 $\pm$ 8.1 &\textbf{38.6 $\pm$ 15.5} \\ 
& \textcolor[HTML]{dead08}{$\mathbf{AFC_{m}}$}& 0.6 $\pm$ 12.6 &1.7 $\pm$ 12.8 &0.4 $\pm$ 12.2 &- &- &- &12.8 $\pm$ 3.8 &- &5.9 $\pm$ 10.1 & \textcolor[HTML]{dead08}{\textbf{45.6 $\pm$ 25.3}} \\ 
& CAR& 2.7 $\pm$ 8.2 &7.1 $\pm$ 15.2 &7.1 $\pm$ 15.2 &- &- &- &24.7 $\pm$ 0.0 &- &11.1 $\pm$ 21.2 & \textbf{45.7 $\pm$ 39.3} \\ 
& \textcolor[HTML]{dead08}{$\mathbf{CC}$}& -1.4 $\pm$ 14.8 &1.1$\pm$ 9.2  &2.1 $\pm$ 7.2 &- &- &- &6.5 $\pm$ 0.8 &- &3.0 $\pm$ 12.3 & \textcolor[HTML]{dead08}{\textbf{46.8 $\pm$ 25.2}} \\ 
& NY\textsubscript{small}& 2.2 $\pm$ 3.7 &3.0 $\pm$ 4.4 &2.8 $\pm$ 4.5 &- &- &- &13.2 $\pm$ 0.0 &- &26.6 $\pm$ 13.2 & \textbf{40.8 $\pm$ 13.9} \\ 
& NY\textsubscript{all}& -0.1 $\pm$ 2.8 &-1.3 $\pm$ 3.0 &-1.0 $\pm$ 2.9 &- &- &- &1.4 $\pm$ 0.8 &- &20.7 $\pm$ 2.5 & \textbf{25.2 $\pm$ 4.0} \\ 
        \midrule
\multirow{6}{*}{$\mathbf{ResNet50}$} 
& AFC& 29.2 $\pm$ 5.5 &30.0 $\pm$ 8.0 &29.1 $\pm$ 11.5 &- &- &- &28.4 $\pm$ 8.4 &- &28.4 $\pm$ 8.4 & \textbf{30.8 $\pm$ 6.8} \\ 
& AFC$_m$& -3.7 $\pm$ 3.5 &-3.1 $\pm$ 5.6 &-3.5 $\pm$ 8.7 &- &- &- &12.8 $\pm$ 3.8 &- &10.4 $\pm$ 8.2 & \textbf{25.0 $\pm$ 14.3} \\ 
& CAR& 12.5 $\pm$ 26.7 &10.8 $\pm$ 23.5 &8.6 $\pm$ 32.2 &- &- &- &24.7 $\pm$ 0.0 &- &4.4 $\pm$ 25.8 & \textbf{40.2 $\pm$ 25.9} \\ 
& CC& 2.2 $\pm$ 19.3 &-2.7 $\pm$ 11.5 & \textbf{8.9 $\pm$ 16.0} &- &- &- &6.5 $\pm$ 0.8 &- &3.4 $\pm$ 20.9 & 3.7 $\pm$ 3.8 \\ 
& NY\textsubscript{small}& 0.3 $\pm$ 2.8 &1.6 $\pm$ 4.2 &0.6 $\pm$ 5.2 &- &- &- &13.2 $\pm$ 0.0 &- &8.2 $\pm$ 6.5 & \textbf{22.6 $\pm$ 6.2} \\ 
& NY\textsubscript{all}& -3.4 $\pm$ 2.5 &-1.1 $\pm$ 1.9 &-2.1 $\pm$ 3.8 &- &- &- &1.1 $\pm$ 0.0 &- &20.1 $\pm$ 2.8 & \textbf{27.1 $\pm$ 7.8} \\ 
        \midrule
\multirow{6}{*}{$\mathbf{DenseNet121}$} 
& AFC& 35.2 $\pm$ 5.8 &31.3 $\pm$ 5.5 &30.6 $\pm$ 5.4 &- &- &- &32.1 $\pm$ 7.3 &- &32.1 $\pm$ 7.3 & \textbf{42.1 $\pm$ 14.9} \\ 
&  \textcolor[HTML]{2eb86c}{$\mathbf{AFC_{m}}$}& 4.1 $\pm$ 12.5 &4.5 $\pm$ 12.1 &3.8 $\pm$ 12.4 &- &- &- &12.8 $\pm$ 3.8 &- &20.1 $\pm$ 4.6 &  \textcolor[HTML]{2eb86c}{\textbf{51.8 $\pm$ 31.1}} \\ 
& \textcolor[HTML]{dead08}{$\mathbf{CAR}$}& 4.1 $\pm$ 22.5 &-3.6 $\pm$ 24.8 &2.8 $\pm$ 21.4 &- &- &- &24.7 $\pm$ 0.0 &- &18.7 $\pm$ 26.2 & \textcolor[HTML]{dead08}{\textbf{70.4 $\pm$ 36.8}} \\ 
& CC& -3.4 $\pm$ 10.9 &-0.1 $\pm$ 8.7 &2.3 $\pm$ 15.1 &- &- &- &8.0 $\pm$ 2.9 &- &-1.5 $\pm$ 9.4 & \textbf{32.5 $\pm$ 44.0} \\ 
& NY\textsubscript{small}& -1.1 $\pm$ 3.6 &-1.4 $\pm$ 4.9 &-1.3 $\pm$ 2.7 &- &- &- &13.2 $\pm$ 0.0 &- &21.2 $\pm$ 8.5 & \textbf{43.9 $\pm$ 24.1} \\ 
& NY\textsubscript{all}& -0.3 $\pm$ 0.9 &-0.4 $\pm$ 2.0 &0.1 $\pm$ 1.4 &- &- &- &27.8 $\pm$ 18.5 &- &20.9 $\pm$ 3.2 & \textbf{33.1 $\pm$ 13.8} \\ 
        \midrule
        \midrule

\multirow{6}{*}{$\mathbf{MedCLIP_{c}}$} 
& AFC& 26.8 $\pm$ 10.7 &2.6 $\pm$ 17.9 &33.8 $\pm$ 10.5 &- &- &- &33.7 $\pm$ 9.3 &- &33.7 $\pm$ 8.8 & \textbf{39.4 $\pm$ 8.7} \\ 
& AFC$_m$& -1.9 $\pm$ 12.9 &-1.5 $\pm$ 12.8 &-1.2 $\pm$ 12.9 &- &- &- &13.4 $\pm$ 4.7 &- &17.5 $\pm$ 11.2 & \textbf{18.1 $\pm$ 8.0} \\ 
& CAR& 14.8 $\pm$ 22.2 &31.9 $\pm$ 22.0 &9.3 $\pm$ 23.8 &- &- &- &28.9 $\pm$ 9.5 &- &19.2 $\pm$ 20.0 & \textbf{49.0 $\pm$ 24.2} \\ 
& CC& 1.6 $\pm$ 10.9 &1.5 $\pm$ 12.0 &3.1 $\pm$ 12.4 &- &- &- &4.0 $\pm$ 16.8 &- &7.4 $\pm$ 2.4 & \textbf{8.5 $\pm$ 4.3} \\ 
& NY\textsubscript{small}& 4.1 $\pm$ 12.3 &4.3 $\pm$ 12.5 &4.2 $\pm$ 12.6 &- &- &- &15.2 $\pm$ 4.6 &- &21.4 $\pm$ 5.5 & \textbf{29.5 $\pm$ 10.7} \\ 
& NY\textsubscript{all}& 0.6 $\pm$ 2.8 &0.4 $\pm$ 2.9 &0.6 $\pm$ 2.7 &- &- &- &16.3 $\pm$ 10.6 &- &19.9 $\pm$ 10.3 & \textbf{30.9 $\pm$ 5.4} \\ 
        \midrule
        
\multirow{6}{*}{$\mathbf{BioMEdCLIP}$} 
& AFC& \textbf{38.9 $\pm$ 12.0} &38.5 $\pm$ 12.6 &37.1 $\pm$ 13.3 &38.5 $\pm$ 12.3 &37.4 $\pm$ 15.6 &37.4 $\pm$ 15.6 &36.1 $\pm$ 13.5 &36.8 $\pm$ 12.2 &38.7 $\pm$ 11.8 &34.3 $\pm$ 10.7 \\ 
& AFC$_m$& 8.3 $\pm$ 7.9 &6.9 $\pm$ 9.6 &6.9 $\pm$ 9.6 &-11.8 $\pm$ 6.9 &1.4 $\pm$ 13.3 &-11.9 $\pm$ 6.8 & \textbf{12.8 $\pm$ 3.8} &9.9 $\pm$ 8.4 &10.6 $\pm$ 14.9 & \textbf{12.8 $\pm$ 3.8} \\ 
& CAR& -11.5 $\pm$ 35.6 &9.9 $\pm$ 21.7 &9.9 $\pm$ 21.7 &-11.5 $\pm$ 35.6 &-11.5 $\pm$ 35.6 &-11.5 $\pm$ 35.6 &24.7 $\pm$ 0.0 & \textbf{40.8 $\pm$ 2.8} &13.9 $\pm$ 26.1 &22.8 $\pm$ 11.8 \\ 
& CC& -4.7 $\pm$ 21.4 &-4.7 $\pm$ 21.4 &-4.7 $\pm$ 21.4 &-3.3 $\pm$ 18.1 &-3.3 $\pm$ 18.1 &-3.3 $\pm$ 18.1 &3.2 $\pm$ 12.1 &-0.2 $\pm$ 15.2 &-0.8 $\pm$ 9.3 & \textbf{6.8 $\pm$ 1.3} \\ 
& \textcolor[HTML]{dead08}{$\mathbf{NY_{small}}$}& 7.3 $\pm$ 6.8 &1.5 $\pm$ 8.1 &1.5 $\pm$ 8.1 &1.3 $\pm$ 8.2 &1.3 $\pm$ 8.2 &1.3 $\pm$ 8.2 &13.2 $\pm$ 0.0 &-6.7 $\pm$ 5.7 &29.2 $\pm$ 3.7 & \textcolor[HTML]{dead08}{\textbf{43.8 $\pm$ 7.8}}\\ 
& NY\textsubscript{all}& -2.3 $\pm$ 6.3 &-8.6 $\pm$ 3.1 &-8.7 $\pm$ 3.1 &-4.3 $\pm$ 3.6 &-4.4 $\pm$ 3.6 &-4.3 $\pm$ 3.6 &-0.3 $\pm$ 1.6 &2.2 $\pm$ 3.6 & \textbf{29.7 $\pm$ 3.7} & 18.5 $\pm$ 5.5 \\ 
        \midrule
\multirow{6}{*}{$\mathbf{CLIP\textbf{-}Large}$} 
& AFC& 30.1 $\pm$ 5.8 &31.9 $\pm$ 10.3 &\textbf{33.9 $\pm$ 8.3} &31.3 $\pm$ 6.9 &31.9 $\pm$ 9.3 &31.9 $\pm$ 9.3 &33.5 $\pm$ 9.0 &30.1 $\pm$ 8.2 &33.5 $\pm$ 9.0 &17.5 $\pm$ 13.2 \\ 
& AFC$_m$& 12.8 $\pm$ 3.8 &10.2 $\pm$ 4.7 &5.4 $\pm$ 10.3 &9.6 $\pm$ 9.6 &5.1 $\pm$ 10.9 &4.3 $\pm$ 11.1 &12.8 $\pm$ 3.8 &4.3 $\pm$ 7.5 & \textbf{13.0 $\pm$ 16.1} &12.8 $\pm$ 3.8 \\ 
& CAR& 4.3 $\pm$ 29.8 &27.6 $\pm$ 6.6 &5.2 $\pm$ 18.3 &2.0 $\pm$ 28.1 &5.3 $\pm$ 30.8 &5.3 $\pm$ 30.8 &24.7 $\pm$ 0.0 &-5.8 $\pm$ 29.2 &24.6 $\pm$ 14.4 & \textbf{28.2 $\pm$ 25.3} \\ 
& CC& -6.4 $\pm$ 15.1 &-4.6 $\pm$ 10.7 &-7.5 $\pm$ 10.4 &-7.1 $\pm$ 6.2 &-1.0 $\pm$ 13.7 &-1.1 $\pm$ 13.0 &3.5 $\pm$ 6.4 &3.3 $\pm$ 12.6 &2.6 $\pm$ 19.8 & \textbf{6.5 $\pm$ 0.8} \\ 
& NY\textsubscript{small}& 13.2 $\pm$ 0.0 &-4.8 $\pm$ 7.5 &-4.3 $\pm$ 7.2 &5.5 $\pm$ 11.7 &1.5 $\pm$ 4.5 &-3.5 $\pm$ 6.0 &13.2 $\pm$ 0.0 &0.5 $\pm$ 8.0 & \textbf{21.2 $\pm$ 16.5} &1 2.4 $\pm$ 1.7 \\ 
& NY\textsubscript{all}& 39.7 $\pm$ 8.9 & \textbf{43.5 $\pm$ 10.2} &32.5 $\pm$ 23.6 &1.2 $\pm$ 0.8 &1.3 $\pm$ 2.0 &-1.1 $\pm$ 4.5 &28.5 $\pm$ 19.7 &-0.3 $\pm$ 5.2 &21.7 $\pm$ 5.1 &33.7 $\pm$ 9.1 \\ 
        \midrule

\multirow{6}{*}{$\mathbf{DINOv2_{s}}$} 
& \textcolor[HTML]{dead08}{$\mathbf{AFC}$} & 36.5 $\pm$ 6.1 &33.3 $\pm$ 10.1 &36.3 $\pm$ 6.7 &36.5 $\pm$ 9.2 &36.0 $\pm$ 6.9 &36.0 $\pm$ 6.9 &34.2 $\pm$ 10.6 &34.3 $\pm$ 7.2 &34.2 $\pm$ 10.6 & \textcolor[HTML]{dead08}{\textbf{49.6 $\pm$ 13.8}} \\ 
& AFC$_m$& \textbf{32.5 $\pm$ 20.3} &16.7 $\pm$ 15.4 &26.0 $\pm$ 23.3 &8.5 $\pm$ 8.5 &12.8 $\pm$ 3.8 &5.0 $\pm$ 12.3 &22.0 $\pm$ 10.2 &3.6 $\pm$ 11.2 &15.3 $\pm$ 13.8 &28.2 $\pm$ 13.7 \\ 
&  \textcolor[HTML]{2eb86c}{$\mathbf{CAR}$}&  \textcolor[HTML]{2eb86c}{\textbf{77.2 $\pm$ 16.4}} &27.6 $\pm$ 6.6 &68.7 $\pm$ 28.8 &14.8 $\pm$ 22.1 &14.8 $\pm$ 22.1 &14.8 $\pm$ 22.1 &11.5 $\pm$ 19.5 &14.8 $\pm$ 22.1 &12.3 $\pm$ 36.4 &28.6 $\pm$ 40.8 \\ 
& CC& -2.6 $\pm$ 8.6 &37.7 $\pm$ 27.6 &-3.5 $\pm$ 11.3 &1.1 $\pm$ 11.1 &8.1 $\pm$ 5.6 &7.2 $\pm$ 6.8 & \textbf{39.2 $\pm$ 23.8} &0.9 $\pm$ 11.8 &1.8 $\pm$ 5.6 &2.2 $\pm$ 9.3 \\ 
& NY\textsubscript{small}& \textbf{41.3 $\pm$ 14.1} &13.2 $\pm$ 0.0 &30.9 $\pm$ 20.3 &5.0 $\pm$ 7.8 &9.3 $\pm$ 6.7 &9.9 $\pm$ 4.7 &13.2 $\pm$ 0.0 &-5.7 $\pm$ 5.4 &31.4 $\pm$ 5.4 &33.9 $\pm$ 19.6 \\ 
& \textcolor[HTML]{dead08}{$\mathbf{NY_{all}}$} & 35.9 $\pm$ 7.2 &34.6 $\pm$ 6.7 &35.2 $\pm$ 6.2 &21.6 $\pm$ 15.0 &4.2 $\pm$ 1.8 &3.1 $\pm$ 3.4 &35.2 $\pm$ 8.2 &-2.1 $\pm$ 1.6 &26.1 $\pm$ 6.4 & \textcolor[HTML]{dead08}{\textbf{44.6 $\pm$ 18.5}} \\ 
        \midrule

\multirow{6}{*}{$\mathbf{DINOv2_{b}}$} 
& AFC& 34.7 $\pm$ 8.2 & \textbf{36.1 $\pm$ 8.7} &34.0 $\pm$ 5.8 &34.1 $\pm$ 9.5 &32.6 $\pm$ 7.5 &32.6 $\pm$ 7.5 &35.0 $\pm$ 5.9 &33.0 $\pm$ 9.8 &35.0 $\pm$ 5.9 &35.9 $\pm$ 11.3 \\ 
& AFC$_m$& \textbf{30.6 $\pm$ 25.0} &28.2 $\pm$ 19.6 &29.2 $\pm$ 22.1 &14.7 $\pm$ 7.8 &7.7 $\pm$ 7.3 &13.1 $\pm$ 6.0 &19.8 $\pm$ 10.5 &5.7 $\pm$ 8.5 &15.6 $\pm$ 12.8 &12.8 $\pm$ 3.8 \\ 
& CAR& 45.4 $\pm$ 19.1 &24.7 $\pm$ 0.0 &18.6 $\pm$ 19.9 &21.8 $\pm$ 29.4 &28.5 $\pm$ 32.0 &22.9 $\pm$ 22.0 & \textbf{47.3 $\pm$ 46.5} &22.9 $\pm$ 22.0 &18.1 $\pm$ 29.4 &29.7 $\pm$ 9.1 \\ 
&  \textcolor[HTML]{2eb86c}{$\mathbf{CC}$} & 14.9 $\pm$ 15.3 &34.0 $\pm$ 26.6 &10.8 $\pm$ 16.0 &7.4 $\pm$ 9.4 &4.9 $\pm$ 8.5 &0.3 $\pm$ 9.0 &  \textcolor[HTML]{2eb86c}{\textbf{46.9 $\pm$ 29.6}} &6.2 $\pm$ 9.1 &3.7 $\pm$ 12.1 &3.1 $\pm$ 7.7 \\ 
& NY\textsubscript{small}& 35.0 $\pm$ 21.0 &13.2 $\pm$ 0.0 & \textbf{37.5 $\pm$ 21.9} &1.9 $\pm$ 10.4 &7.7 $\pm$ 11.7 &4.6 $\pm$ 12.0 &13.2 $\pm$ 0.0 &-4.8 $\pm$ 11.0 &24.5 $\pm$ 11.4 &13.2 $\pm$ 0.0 \\ 
& NY\textsubscript{all}& \textbf{40.3 $\pm$ 8.9} &35.7 $\pm$ 8.8 &34.4 $\pm$ 8.5 &1.8 $\pm$ 1.0 &0.3 $\pm$ 1.8 &-2.6 $\pm$ 7.1 &35.5 $\pm$ 8.6 &0.8 $\pm$ 1.8 &25.9 $\pm$ 9.3 &15.1 $\pm$ 15.4 \\ 
        \midrule
\multirow{6}{*}{$\mathbf{DINOv2_{l}}$} 
& AFC& 36.6 $\pm$ 11.0 &31.8 $\pm$ 13.7 &34.0 $\pm$ 5.8 &32.0 $\pm$ 14.5 & \textbf{38.8 $\pm$ 11.5} &32.6 $\pm$ 7.5 &35.0 $\pm$ 5.9 &33.0 $\pm$ 9.8 &35.0 $\pm$ 5.9 &18.5 $\pm$ 8.2 \\ 
& AFC$_m$& 14.9 $\pm$ 3.7 & \textbf{15.5 $\pm$ 4.7} &15.2 $\pm$ 4.1 &11.1 $\pm$ 3.5 &12.8 $\pm$ 3.8 &2.9 $\pm$ 9.9 &12.8 $\pm$ 3.8 &-2.9 $\pm$ 12.3 &13.2 $\pm$ 12.1 &11.8 $\pm$ 5.6 \\ 
& CAR& \textbf{37.2 $\pm$ 20.7} &32.2 $\pm$ 16.8 &34.2 $\pm$ 14.5 &32.2 $\pm$ 16.8 &32.2 $\pm$ 16.8 &32.2 $\pm$ 16.8 &24.7 $\pm$ 0.0 &32.2 $\pm$ 16.8 &8.6 $\pm$ 23.7 &6.8 $\pm$ 33.1 \\ 
& CC& 8.3 $\pm$ 5.7 &6.5 $\pm$ 0.8 &10.8 $\pm$ 16.0 &10.1 $\pm$ 5.5 &11.0 $\pm$ 5.9 &11.0 $\pm$ 5.9 &6.8 $\pm$ 6.7 &4.0 $\pm$ 10.8 & \textbf{15.5 $\pm$ 12.9} &4.3 $\pm$ 5.5 \\ 
& NY\textsubscript{small}& 13.2 $\pm$ 0.0 &7.4 $\pm$ 9.0 &13.2 $\pm$ 0.0 &9.4 $\pm$ 8.4 &9.4 $\pm$ 8.4 &10.0 $\pm$ 9.1 &13.2 $\pm$ 0.0 &-9.6 $\pm$ 14.2 & \textbf{16.9 $\pm$ 8.6} &13.2 $\pm$ 0.0 \\ 
& \textcolor[HTML]{2eb86c}{$\mathbf{NY_{all}}$}& 39.6 $\pm$ 10.5 &44.6 $\pm$ 11.2 &37.2 $\pm$ 11.9 &1.1 $\pm$ 1.0 &-0.9 $\pm$ 4.4 &-0.0 $\pm$ 6.9 &  \textcolor[HTML]{2eb86c}{\textbf{45.1 $\pm$ 9.5}} &1.7 $\pm$ 1.7 &25.6 $\pm$ 7.2 &11.0 $\pm$ 4.4 \\ 
        \midrule

\multirow{6}{*}{$\mathbf{MedCLIP_{v}}$} 
& AFC& 28.7 $\pm$ 4.1 &4.2 $\pm$ 12.5 & \textbf{34.8 $\pm$ 4.8} &- &- &- &27.7 $\pm$ 16.3 &31.7 $\pm$ 17.0 &27.3 $\pm$ 16.4 &18.3 $\pm$ 9.2 \\ 
& AFC$_m$& 4.7 $\pm$ 9.5 &7.0 $\pm$ 7.1 &8.3 $\pm$ 7.5 &- &- &- &12.8 $\pm$ 3.8 &-8.6 $\pm$ 9.5 & \textbf{16.9 $\pm$ 20.4} &12.8 $\pm$ 3.8 \\ 
& CAR& 14.8 $\pm$ 22.1 &14.8 $\pm$ 22.1 &10.3 $\pm$ 19.8 &- &- &- &24.7 $\pm$ 0.0 &-11.2 $\pm$ 21.5 & \textbf{35.2 $\pm$ 10.4} &29.1 $\pm$ 24.5 \\ 
& CC& \textbf{7.7 $\pm$ 2.3} &6.1 $\pm$ 0.9 &1.4 $\pm$ 7.2 &- &- &- &5.1 $\pm$ 10.3 &-6.7 $\pm$ 8.3 &1.5 $\pm$ 7.0 &-3.6 $\pm$ 9.2 \\ 
& NY\textsubscript{small}& 0.7 $\pm$ 7.7 &3.2 $\pm$ 9.4 &2.1 $\pm$ 7.3 &- &- &- &13.2 $\pm$ 0.0 &-4.5 $\pm$ 8.7 & \textbf{22.1 $\pm$ 11.3} &10.1 $\pm$ 6.9 \\ 
& NY\textsubscript{all}& -1.5 $\pm$ 2.1 &0.2 $\pm$ 3.2 &-1.3 $\pm$ 1.8 &- &- &- &0.8 $\pm$ 2.1 &-1.2 $\pm$ 0.6 & \textbf{24.2 $\pm$ 4.7} &11.1 $\pm$ 13.6 \\ 
        \midrule
\multirow{6}{*}{$\mathbf{PubMedCLIP}$} 
&  \textcolor[HTML]{2eb86c}{$\mathbf{AFC}$}& 33.6 $\pm$ 6.4 &27.0 $\pm$ 12.5 &33.9 $\pm$ 6.3 &33.9 $\pm$ 5.7 &29.5 $\pm$ 5.6 &28.4 $\pm$ 6.5 &34.6 $\pm$ 6.6 &30.9 $\pm$ 7.7 &34.6 $\pm$ 6.6 &  \textcolor[HTML]{2eb86c}{\textbf{47.4 $\pm$ 13.7}} \\ 
& AFC$_m$& 2.9 $\pm$ 8.2 &6.2 $\pm$ 10.2 &5.9 $\pm$ 10.1 &2.6 $\pm$ 11.8 &0.7 $\pm$ 13.6 &2.6 $\pm$ 11.8 &12.8 $\pm$ 3.8 &-6.7 $\pm$ 9.9 &8.2 $\pm$ 11.4 &\textbf{44.9 $\pm$ 27.1} \\ 
& CAR& -4.2 $\pm$ 26.4 &-1.7 $\pm$ 25.0 &-1.7 $\pm$ 25.0 &-0.5 $\pm$ 24.7 &-4.2 $\pm$ 26.4 &-4.2 $\pm$ 26.4 &24.7 $\pm$ 0.0 &-7.2 $\pm$ 31.1 &24.7 $\pm$ 0.0 &\textbf{25.3 $\pm$ 31.1} \\ 
& CC& -0.3 $\pm$ 7.8 &1.2 $\pm$ 6.2 &1.2 $\pm$ 6.2 &1.2 $\pm$ 6.2 &1.2 $\pm$ 6.2 &1.2 $\pm$ 6.2 &6.5 $\pm$ 0.8 &-7.1 $\pm$ 10.3 &7.6 $\pm$ 11.6 &\textbf{17.8 $\pm$ 16.4} \\ 
&  \textcolor[HTML]{2eb86c}{$\mathbf{NY_{small}}$} & -0.6 $\pm$ 10.7 &1.4 $\pm$ 10.5 &1.4 $\pm$ 10.5 &1.6 $\pm$ 10.6 &1.4 $\pm$ 10.5 &1.6 $\pm$ 10.6 &13.2 $\pm$ 0.0 &2.0 $\pm$ 5.8 &21.1 $\pm$ 9.0 &  \textcolor[HTML]{2eb86c}{\textbf{65.0 $\pm$ 25.2}} \\ 
& NY\textsubscript{all}& 2.0 $\pm$ 4.1 &3.3 $\pm$ 2.3 &3.2 $\pm$ 2.3 &1.0 $\pm$ 4.1 &1.0 $\pm$ 4.0 &1.0 $\pm$ 4.1 &1.1 $\pm$ 0.0 &-1.5 $\pm$ 3.7 &18.0 $\pm$ 4.2 &\textbf{38.4 $\pm$ 17.1} \\ 

        \bottomrule
    \end{tabular}
    }
     \caption{This table presents the $\mathbf{MCC}$ scores for all the experiments divided in two main sections by a double line, to discriminate the results obtained from CNNs (top) and FMs (bottom).
    Each row displays a model's performance across multiple tasks under different fine-tuning approaches, with the best fine-tuning approach per task highlighted in bold.
    Cells highlighted in \textcolor[HTML]{2eb86c}{green} and \textcolor[HTML]{dead08}{orange} indicate the top and second-best results, respectively, for each dataset task across all model and fine-tuning combinations. }
    \label{tab:results_MCC}
\end{table}

%% file: tables/results_AUC.tex
\begin{table}[!ht]
    \setlength{\extrarowheight}{0.2pt} 
    \renewcommand{\arraystretch}{1.0} 
    \centering
    \resizebox{\columnwidth}{!}{
    \begin{tabular}{{l|l|cccccccccc}}
        \toprule
        \textbf{{Model}} & 
        \backslashbox{\textbf{Task}}{\textbf{FT}} & 
        $\mathbf{{LoRA_{{r=4}}}}$ & 
        $\mathbf{{LoRA_{{r=8}}}}$ & 
        $\mathbf{{LoRA_{{r=16}}}}$ & 
        $\mathbf{{VeRA_{{r=4}}}}$ &
        $\mathbf{{VeRA_{{r=8}}}}$ & 
        $\mathbf{{VeRA_{{r=16}}}}$ & 
        $\mathbf{{BitFit}}$ &
        $\mathbf{{(\ IA\ )^3}}$ & 
        $\mathbf{{LP}}$ & 
        $\mathbf{{FFT}}$  \\
    
        \midrule
        
\multirow{6}{*}{$\mathbf{ResNet18}$} & AFC & 71.9 $\pm$ 9.0 & 71.5 $\pm$ 8.7 & 71.8 $\pm$ 7.8 & - & - & - & 72.0 $\pm$ 7.6 & - & 72.0 $\pm$ 7.6 & \textbf{77.3 $\pm$ 6.8} \\
                                  & \textcolor[HTML]{2eb86c}{$\mathbf{AFC_{m}}$} & 23.2 $\pm$ 10.3 & 23.5 $\pm$ 10.6 & 23.5 $\pm$ 10.8 & - & - & - & 24.0 $\pm$ 11.5 & - & 38.1 $\pm$ 10.4 & \textbf{\textcolor[HTML]{2eb86c}{67.2 $\pm$ 19.3}} \\
                                  & CAR & 43.4 $\pm$ 6.7 & 40.3 $\pm$ 7.6 & 42.7 $\pm$ 5.5 & - & - & - & 41.1 $\pm$ 6.5 & - & 42.2 $\pm$ 14.9 & \textbf{81.1 $\pm$ 20.6} \\
                                  & \textcolor[HTML]{dead08}{$\mathbf{CC}$} & 42.7 $\pm$ 6.0 & 41.9$ \pm$ 5.1 & 42.2 $\pm$ 6.1 & - & - & - & 40.2 $\pm$ 3.9 & - & 43.2 $\pm$ 7.3 & \textcolor[HTML]{dead08}{\textbf{72.7 $\pm$ 21.1}} \\
                                  & NY\textsubscript{small} & 15.5 $\pm$ 2.9 & 15.4 $\pm$ 3.0 & 15.4 $\pm$ 2.9 & - & - & - & 15.4 $\pm$ 2.9 & - & 40.9 $\pm$ 6.7 & \textbf{58.4 $\pm$ 13.4} \\
                                  & NY\textsubscript{all} & 36.0 $\pm$ 2.0 & 35.7 $\pm$ 1.9 & 35.7 $\pm$ 1.9 & - & - & - & 45.6 $\pm$ 6.4 & - & 51.5 $\pm$ 2.1 & \textbf{56.1 $\pm$ 3.2} \\

        \midrule
\multirow{6}{*}{$\mathbf{ResNet50}$} & AFC & 71.9 $\pm$ 7.6 & 72.2 $\pm$ 7.8 & \textbf{73.0 $\pm$ 7.2} & - & - & - & 72.3 $\pm$ 7.8 & - & 72.3 $\pm$ 7.8 & 72.9 $\pm$ 6.9 \\
                                  &  AFC$_m$  & 19.0 $\pm$ 4.7 & 18.7 $\pm$ 4.6 & 19.1 $\pm$ 5.1 & - & - & - & 19.8 $\pm$ 5.1 & - & 38.4 $\pm$ 13.9 & \textbf{48.4 $\pm$ 10.7} \\
                                  & CAR & 33.4 $\pm$ 8.0 & 35.0 $\pm$ 11.4 & 33.8 $\pm$ 9.9 & - & - & - & 34.4 $\pm$ 10.1 & - & 41.9 $\pm$ 20.6 & \textbf{58.7 $\pm$ 22.2} \\
                                  & CC & 43.7 $\pm$ 5.8 & 43.8 $\pm$ 5.9 & 45.5 $\pm$ 9.8 & - & - & - & 43.9 $\pm$ 6.4 & - & \textbf{46.5 $\pm$ 10.7} & 41.8 $\pm$ 5.0 \\
                                  & NY\textsubscript{small} & 13.6 $\pm$ 2.3 & 13.8 $\pm$ 2.3 & 13.7 $\pm$ 2.1 & - & - & - & 13.8 $\pm$ 2.1 & - & 36.5 $\pm$ 6.8 & \textbf{44.6 $\pm$ 2.7} \\
                                  & NY\textsubscript{all} & 35.2 $\pm$ 2.3 & 35.4 $\pm$ 0.9 & 35.3 $\pm$ 2.0 & - & - & - & 35.6 $\pm$ 2.5 & - & 53.3 $\pm$ 3.8 & \textbf{57.7 $\pm$ 7.0} \\
             
        \midrule
                                \multirow{6}{*}{$\mathbf{DenseNet121}$} & AFC & 72.1 $\pm$ 9.6 & 71.9 $\pm$ 9.6 & 71.9 $\pm$ 9.5 & - & - & - & 71.8 $\pm$ 10.1 & - & 71.8 $\pm$ 10.1 & \textbf{81.9 $\pm$ 4.8} \\
                                  & \textcolor[HTML]{dead08}{$\mathbf{AFC_{m}}$} & 20.9 $\pm$ 6.7 & 20.9 $\pm$ 6.7 & 21.0 $\pm$ 6.9 & - & - & - & 26.4 $\pm$ 8.6 & - & 40.0 $\pm$ 11.1 & \textcolor[HTML]{dead08}{\textbf{67.0 $\pm$ 24.7}} \\
                                  & \textcolor[HTML]{dead08}{$\mathbf{CAR}$} & 39.2 $\pm$ 19.0 & 39.0 $\pm$ 19.4 & 39.7 $\pm$ 20.9 & - & - & - & 39.0 $\pm$ 18.0 & - & 47.4 $\pm$ 17.7 & \textcolor[HTML]{dead08}{\textbf{83.3 $\pm$ 28.6}} \\
                                  & CC & 37.6 $\pm$ 5.4 & 38.0 $\pm$ 6.5 & 43.4 $\pm$ 6.7 & - & - & - & 37.8 $\pm$ 6.1 & - & 39.6 $\pm$ 6.7 & \textbf{69.3 $\pm$ 24.6} \\
                                  & \textcolor[HTML]{dead08}{$\mathbf{NY_{small}}$} & 13.8 $\pm$ 2.1 & 13.8 $\pm$ 2.1 & 13.9 $\pm$ 2.2 & - & - & - & 16.6 $\pm$ 2.1 & - & 37.6 $\pm$ 2.9 & \textcolor[HTML]{dead08}{\textbf{62.6 $\pm$ 17.9}} \\
                                  & NY\textsubscript{all} & 33.7 $\pm$ 1.7 & 35.5 $\pm$ 1.4 & 33.8 $\pm$ 1.6 & - & - & - & 60.2 $\pm$ 10.2 & - & 51.2 $\pm$ 2.9 & \textbf{61.7 $\pm$ 11.4} \\
                                  
        \midrule       
        \midrule
        
\multirow{6}{*}{$\mathbf{MedCLIP_c}$} & AFC & 77.6 $\pm$ 13.1 & 58.3 $\pm$ 23.1 & 74.7 $\pm$ 9.4 & - & - & - & 74.4 $\pm$ 8.9 & - & 74.5 $\pm$ 9.0 & \textbf{77.1 $\pm$ 9.3} \\
                                  & AFC$_m$ & 23.0 $\pm$ 11.9 & 23.4 $\pm$ 12.5 & 22.9 $\pm$ 11.8 & - & - & - & 23.6 $\pm$ 12.4 & - & 33.2 $\pm$ 14.7 & \textbf{43.1 $\pm$ 11.0} \\
                                  & CAR & 38.9 $\pm$ 14.2 & 56.3 $\pm$ 17.0 & 38.7 $\pm$ 14.4 & - & - & - & 40.7 $\pm$ 15.2 & - & 42.4 $\pm$ 13.1 & \textbf{75.8 $\pm$ 26.0} \\
                                  & CC & 42.8 $\pm$ 6.6 & 42.5 $\pm$ 5.6 & 43.2 $\pm$ 6.6 & - & - & - & 43.0 $\pm$ 6.4 & - & 38.8 $\pm$ 6.3 & \textbf{44.1 $\pm$ 5.3} \\
                                  & NY\textsubscript{small} & 19.9 $\pm$ 12.0 & 20.2 $\pm$ 12.4 & 20.2 $\pm$ 12.4 & - & - & - & 20.3 $\pm$ 12.5 & - & 41.6 $\pm$ 8.2 & \textbf{43.8 $\pm$ 12.9} \\
                                  & NY\textsubscript{all} & 43.5 $\pm$ 9.6 & 43.5 $\pm$ 9.8 & 43.7 $\pm$ 9.6 & - & - & - & 50.6 $\pm$ 6.7 & - & 58.4 $\pm$ 2.5 & \textbf{59.9 $\pm$ 4.3} \\

        \midrule

        \multirow{6}{*}{$\mathbf{BioMedCLIP}$} & AFC & \textbf{78.3 $\pm$ 10.2} & 77.4 $\pm$ 9.7 & 76.8 $\pm$ 10.0 & 76.9 $\pm$ 9.2 & 76.9 $\pm$ 10.7 & 76.9 $\pm$ 10.7 & 76.9 $\pm$ 9.6 & 76.6 $\pm$ 10.2 & 76.7 $\pm$ 9.9 & 73.1 $\pm$ 9.5 \\
                                  & AFC$_m$ & 20.3 $\pm$ 6.0 & 19.6 $\pm$ 4.8 & 19.4 $\pm$ 4.6 & 18.0 $\pm$ 2.5 & 20.4 $\pm$ 6.3 & 18.0 $\pm$ 2.4 & 20.0 $\pm$ 3.5 & 25.5 $\pm$ 7.6 & \textbf{40.9 $\pm$ 10.2} & 24.4 $\pm$ 5.3 \\
                                  & CAR & 35.6 $\pm$ 7.8 & 34.6 $\pm$ 12.0 & 34.8 $\pm$ 11.9 & 35.5 $\pm$ 7.8 & 35.5 $\pm$ 7.8 & 35.8 $\pm$ 7.5 & 35.4 $\pm$ 12.6 & \textbf{57.8 $\pm$ 7.2} & 50.9 $\pm$ 13.8 & 43.9 $\pm$ 9.3 \\
                                  & CC & 40.0 $\pm$ 12.8 & 40.0 $\pm$ 12.8 & 39.9 $\pm$ 12.7 & 40.5 $\pm$ 12.6 & 40.5 $\pm$ 12.7 & 40.4 $\pm$ 12.6 & \textbf{43.7 $\pm$ 3.0} & 38.3 $\pm$ 7.5 & \textbf{43.7 $\pm$ 3.0} & 37.7 $\pm$ 10.9 \\
                                  & NY\textsubscript{small} & 14.6 $\pm$ 3.4 & 12.9 $\pm$ 2.0 & 12.9 $\pm$ 2.0 & 13.0 $\pm$ 2.1 & 13.0 $\pm$ 2.1 & 13.0 $\pm$ 2.1 & 12.9 $\pm$ 2.1 & 13.1 $\pm$ 2.3 & 39.2 $\pm$ 2.6 & \textbf{54.8 $\pm$ 10.3} \\
                                  & NY\textsubscript{all} & 34.2 $\pm$ 3.8 & 31.7 $\pm$ 1.5 & 31.7 $\pm$ 1.5 & 32.1 $\pm$ 2.3 & 32.1 $\pm$ 2.3 & 32.1 $\pm$ 2.3 & 32.4 $\pm$ 2.1 & 37.2 $\pm$ 2.3 & \textbf{54.4 $\pm$ 2.8} & 49.1 $\pm$ 3.3 \\

        \midrule
        \multirow{6}{*}{$\mathbf{CLIP\textbf{-}Large}$} & AFC & 74.7 $\pm$ 7.6 & 75.0 $\pm$ 7.5 & 74.8 $\pm$ 7.1 & \textbf{75.1 $\pm$ 7.6} & 75.1 $\pm$ 7.4 & \textbf{75.1 $\pm$ 7.4} & \textbf{75.1 $\pm$ 7.5} & 75.0 $\pm$ 7.3 & \textbf{75.1 $\pm$ 7.5} & 67.4 $\pm$ 8.1 \\
                                  & AFC$_m$ & 24.7 $\pm$ 11.1 & 28.6 $\pm$ 7.4 & 23.5 $\pm$ 11.2 & 26.2 $\pm$ 10.4 & 23.1 $\pm$ 8.8 & 22.9 $\pm$ 11.5 & 32.7 $\pm$ 10.7 & 26.4 $\pm$ 10.6 & \textbf{39.8 $\pm$ 6.0} & 23.5 $\pm$ 9.8 \\
                                  & CAR & 35.7 $\pm$ 13.0 & 56.1 $\pm$ 12.9 & 35.5 $\pm$ 8.4 & 39.4 $\pm$ 14.0 & 36.7 $\pm$ 13.7 & 36.3 $\pm$ 14.0 & 38.2 $\pm$ 23.2 & 27.8 $\pm$ 8.9 & 61.2 $\pm$ 16.4 & \textbf{64.0 $\pm$ 27.7} \\
                                  & CC & 43.6 $\pm$ 8.7 & 42.2 $\pm$ 8.2 & 42.5 $\pm$ 7.2 & 42.4 $\pm$ 8.6 & 42.9 $\pm$ 9.1 & 42.4 $\pm$ 8.5 & \textbf{47.8 $\pm$ 2.9} & 41.6 $\pm$ 8.0 & 41.3 $\pm$ 13.7 & 41.0 $\pm$ 8.3 \\
                                  & NY\textsubscript{small} & 18.8 $\pm$ 6.7 & 12.5 $\pm$ 1.2 & 12.6 $\pm$ 1.4 & 17.8 $\pm$ 7.3 & 12.7 $\pm$ 1.3 & 13.6 $\pm$ 0.9 & 13.1 $\pm$ 3.0 & 16.3 $\pm$ 7.5 & \textbf{43.2 $\pm$ 3.8} & 29.9 $\pm$ 6.9 \\
                                  & NY\textsubscript{all} & 66.2 $\pm$ 7.5 & \textbf{66.8 $\pm$ 9.1} & 56.4 $\pm$ 17.7 & 42.4 $\pm$ 2.9 & 40.8 $\pm$ 2.3 & 34.9 $\pm$ 2.8 & 61.2 $\pm$ 9.2 & 37.6 $\pm$ 4.1 & 53.0 $\pm$ 4.5 & 62.8 $\pm$ 8.7 \\

        \midrule
        \multirow{6}{*}{$\mathbf{DINOv2_s}$} & \textcolor[HTML]{dead08}{$\mathbf{AFC}$} & 76.2 $\pm$ 8.1 & 75.7 $\pm$ 8.8 & 76.5 $\pm$ 7.7 & 76.6 $\pm$ 7.8 & 76.3 $\pm$ 7.7 & 76.3 $\pm$ 7.7 & 75.5 $\pm$ 8.8 & 76.4 $\pm$ 7.9 & 75.5 $\pm$ 8.8 &     \textcolor[HTML]{dead08}{\textbf{84.6 $\pm$ 7.0}} \\
                                  & AFC$_m$ & 53.2 $\pm$ 11.3 & 49.9 $\pm$ 16.8 & \textbf{55.0 $\pm$ 17.2} & 26.9 $\pm$ 13.5 & 24.6 $\pm$ 9.0 & 26.8 $\pm$ 14.9 & 53.1 $\pm$ 22.6 & 23.2 $\pm$ 10.4 & 41.8 $\pm$ 7.9 & 46.0 $\pm$ 14.5 \\
                                  & \textcolor[HTML]{2eb86c}{$\mathbf{CAR}$} & 80.7 $\pm$ 16.2 & 54.5 $\pm$ 15.8 & \textbf{\textcolor[HTML]{2eb86c}{86.6 $\pm$ 20.9}} & 43.6 $\pm$ 21.1 & 40.9 $\pm$ 14.1 & 42.0 $\pm$ 15.1 & 54.9 $\pm$ 13.4 & 41.5 $\pm$ 15.7 & 61.1 $\pm$ 13.4 & 69.9 $\pm$ 30.3 \\
                                  & CC & 38.5 $\pm$ 5.3 & \textbf{66.0 $\pm$ 17.4} & 39.3 $\pm$ 4.8 & 43.4 $\pm$ 9.5 & 44.2 $\pm$ 8.9 & 44.1 $\pm$ 8.4 & 65.7 $\pm$ 14.3 & 40.9 $\pm$ 10.0 & 41.9 $\pm$ 6.8 & 36.6 $\pm$ 6.8 \\
                                  & NY\textsubscript{small} & \textbf{52.0 $\pm$ 8.6} & 20.4 $\pm$ 7.3 & 50.9 $\pm$ 9.7 & 20.6 $\pm$ 8.8 & 18.5 $\pm$ 5.6 & 20.4 $\pm$ 6.4 & 20.4 $\pm$ 5.1 & 13.1 $\pm$ 1.3 & 45.0 $\pm$ 7.5 & 47.4 $\pm$ 15.9 \\
                                  & \textcolor[HTML]{2eb86c}{$\mathbf{NY_{all}}$} & 62.1 $\pm$ 6.3 & 61.4 $\pm$ 6.1 & 60.3 $\pm$ 5.1 & 53.0 $\pm$ 9.9 & 41.8 $\pm$ 1.8 & 41.0 $\pm$ 2.6 & 61.3 $\pm$ 6.5 & 32.9 $\pm$ 0.8 & 54.7 $\pm$ 5.7 & \textbf{\textcolor[HTML]{2eb86c}{71.5 $\pm$ 13.4}} \\   
        \midrule
        \multirow{6}{*}{$\mathbf{DINOv2_b}$} & AFC & 75.1 $\pm$ 8.7 & 74.7 $\pm$ 8.6 & 75.0 $\pm$ 8.3 & 75.2 $\pm$ 8.3 & 75.3 $\pm$ 8.3 & 75.3 $\pm$ 8.3 & 75.5 $\pm$ 8.3 & 74.8 $\pm$ 8.7 & 75.5 $\pm$ 8.3 & \textbf{77.7 $\pm$ 4.4} \\
                                  & AFC$_m$ & 51.0 $\pm$ 15.3 & \textbf{56.9 $\pm$ 13.5} & 56.5 $\pm$ 17.8 & 24.8 $\pm$ 8.7 & 26.2 $\pm$ 11.2 & 23.9 $\pm$ 6.8 & 52.9 $\pm$ 8.6 & 21.7 $\pm$ 7.5 & 44.9 $\pm$ 7.4 & 24.1 $\pm$ 6.7 \\
                                  & CAR & 60.9 $\pm$ 14.5 & 39.2 $\pm$ 14.4 & 40.1 $\pm$ 11.5 & 48.8 $\pm$ 18.2 & 50.6 $\pm$ 19.4 & 40.9 $\pm$ 10.0 & \textbf{65.4 $\pm$ 29.3} & 42.1 $\pm$ 8.9 & 55.2 $\pm$ 14.7 & 44.9 $\pm$ 22.6 \\
                                  & \textcolor[HTML]{2eb86c}{$\mathbf{CC}$} & 51.9 $\pm$ 6.5 & 63.9 $\pm$ 21.7 & 49.1 $\pm$ 6.3 & 43.9 $\pm$ 5.8 & 43.7 $\pm$ 5.0 & 43.5 $\pm$ 5.0 & \textbf{\textcolor[HTML]{2eb86c}{73.3 $\pm$ 22.6}} & 40.4 $\pm$ 4.4 & 53.1 $\pm$ 5.6 & 49.2 $\pm$ 13.9 \\
                                  & NY\textsubscript{small} & 52.5 $\pm$ 13.1 & 20.0 $\pm$ 5.9 & \textbf{55.9 $\pm$ 11.1} & 14.3 $\pm$ 4.2 & 17.0 $\pm$ 3.7 & 14.3 $\pm$ 3.9 & 14.5 $\pm$ 3.2 & 16.1 $\pm$ 2.2 & 43.3 $\pm$ 7.4 & 20.2 $\pm$ 6.0 \\
                                  & NY\textsubscript{all} & \textbf{65.5 $\pm$ 7.5} & 64.0 $\pm$ 6.8 & 62.9 $\pm$ 6.8 & 36.1 $\pm$ 3.3 & 34.4 $\pm$ 2.6 & 34.8 $\pm$ 3.8 & 62.2 $\pm$ 6.4 & 40.5 $\pm$ 2.8 & 56.0 $\pm$ 7.1 & 50.6 $\pm$ 6.7 \\

        \midrule
        \multirow{6}{*}{$\mathbf{DINOv2_l}$} & AFC & 75.3 $\pm$ 8.5 & 75.8 $\pm$ 9.1 & 75.0 $\pm$ 8.3 & 75.1 $\pm$ 9.0 & \textbf{76.3 $\pm$ 8.3} & 75.3 $\pm$ 8.3 & 75.5 $\pm$ 8.3 & 74.8 $\pm$ 8.7 & 75.5 $\pm$ 8.3 & 63.1 $\pm$ 8.8 \\
                                  & AFC$_m$ & 38.6 $\pm$ 13.7 & 37.5 $\pm$ 10.7 & 38.2 $\pm$ 11.6 & 19.9 $\pm$ 4.8 & 19.9 $\pm$ 6.0 & 25.6 $\pm$ 9.7 & 30.9 $\pm$ 3.8 & 21.5 $\pm$ 9.2 & \textbf{41.2 $\pm$ 6.4} & 31.0 $\pm$ 6.9 \\
                                  & CAR & \textbf{64.8 $\pm$ 27.7} & 39.3 $\pm$ 13.6 & 60.7 $\pm$ 23.5 & 42.0 $\pm$ 17.5 & 41.8 $\pm$ 17.6 & 41.8 $\pm$ 17.7 & 39.3 $\pm$ 10.0 & 41.7 $\pm$ 17.7 & 39.1 $\pm$ 20.8 & 35.1 $\pm$ 16.1 \\
                                  & CC & 48.4 $\pm$ 5.4 & 47.2 $\pm$ 11.2 & 49.1 $\pm$ 6.3 & 43.2 $\pm$ 9.8 & 43.3 $\pm$ 9.9 & 43.3 $\pm$ 9.9 & 43.8 $\pm$ 9.5 & 39.0 $\pm$ 7.8 & \textbf{50.1 $\pm$ 10.8} & 38.7 $\pm$ 6.6 \\
                                  & NY\textsubscript{small} & 30.5 $\pm$ 9.4 & 16.9 $\pm$ 8.6 & 30.5 $\pm$ 9.5 & 18.1 $\pm$ 8.1 & 17.5 $\pm$ 8.4 & 17.8 $\pm$ 8.4 & 16.9 $\pm$ 8.3 & 18.0 $\pm$ 7.1 & \textbf{41.9 $\pm$ 9.0} & 26.5 $\pm$ 7.7 \\
                                  & \textcolor[HTML]{dead08}{$\mathbf{NY_{all}}$} & 66.3 $\pm$ 10.4 & 67.8 $\pm$ 8.7 & 63.4 $\pm$ 9.7 & 35.0 $\pm$ 4.0 & 32.2 $\pm$ 3.9 & 34.7 $\pm$ 6.3 & \textcolor[HTML]{dead08}{\textbf{68.6 $\pm$ 8.9}} & 38.3 $\pm$ 3.4 & 54.9 $\pm$ 4.0 & 46.3 $\pm$ 1.5 \\

        \midrule
\multirow{6}{*}{$\mathbf{MedCLIP_{v}}$} 
& AFC & \textbf{74.8 $\pm$ 13.7} & 56.5 $\pm$ 19.9 & 73.7 $\pm$ 8.7 & - & - & - & 72.5 $\pm$ 9.3 & 69.4 $\pm$ 15.1 & 72.2 $\pm$ 9.7 & 63.4 $\pm$ 7.9 \\
& AFC$_m$ & 21.8 $\pm$ 4.3 & 21.0 $\pm$ 3.9 & 20.2 $\pm$ 4.1 & - & - & - & 19.6 $\pm$ 6.2 & 19.5 $\pm$ 4.7 & \textbf{36.8 $\pm$ 10.2} & 31.8 $\pm$ 7.0 \\
& CAR & 36.3 $\pm$ 24.0 & 36.6 $\pm$ 24.0 & 25.8 $\pm$ 5.6 & - & - & - & 23.8 $\pm$ 2.7 & 38.8 $\pm$ 21.3 & \textbf{61.5 $\pm$ 15.8} & 59.4 $\pm$ 25.5 \\
& CC & \textbf{43.6 $\pm$ 10.0} & 43.3 $\pm$ 10.2 & 39.1 $\pm$ 5.2 & - & - & - & 42.6 $\pm$ 8.5 & 36.1 $\pm$ 4.9 & 37.8 $\pm$ 2.5 & 39.4 $\pm$ 7.2 \\
& NY\textsubscript{small} & 13.8 $\pm$ 3.0 & 14.4 $\pm$ 3.9 & 14.8 $\pm$ 4.2 & - & - & - & 16.6 $\pm$ 5.6 & 14.7 $\pm$ 2.1 & \textbf{34.5 $\pm$ 15.5} & 29.5 $\pm$ 4.5 \\
& NY\textsubscript{all} & 34.8 $\pm$ 1.6 & 35.8 $\pm$ 2.5 & 34.8 $\pm$ 1.6 & - & - & - & 35.3 $\pm$ 1.4 & 40.4 $\pm$ 2.8 & \textbf{52.0 $\pm$ 3.6} & 49.3 $\pm$ 7.0 \\

\midrule
\multirow{6}{*}{$\mathbf{PubMedCLIP}$} 
& \textcolor[HTML]{2eb86c}{$\mathbf{AFC}$} & 74.6 $\pm$ 8.4 & 73.8 $\pm$ 8.9 & 74.6 $\pm$ 8.3 & 74.5 $\pm$ 8.4 & 74.9 $\pm$ 8.6 & 73.7 $\pm$ 9.1 & 74.6 $\pm$ 8.3 & 74.9 $\pm$ 8.2 & 74.6 $\pm$ 8.3 & \textbf{\textcolor[HTML]{2eb86c}{84.9 $\pm$ 2.3}} \\
& AFC$_m$ & 21.9 $\pm$ 8.2 & 22.1 $\pm$ 7.6 & 21.9 $\pm$ 7.6 & 22.7 $\pm$ 8.5 & 21.3 $\pm$ 2.9 & 22.6 $\pm$ 8.1 & 27.3 $\pm$ 10.2 & 20.2 $\pm$ 7.7 & 38.5 $\pm$ 8.9 & \textbf{64.2 $\pm$ 22.1} \\
& CAR & 33.9 $\pm$ 10.4 & 34.1 $\pm$ 10.7 & 34.1 $\pm$ 10.7 & 33.5 $\pm$ 10.3 & 33.2 $\pm$ 10.0 & 33.4 $\pm$ 10.2 & 33.1 $\pm$ 11.5 & 28.2 $\pm$ 7.3 & 53.0 $\pm$ 8.2 & \textbf{54.0 $\pm$ 21.5} \\
& CC & 39.1 $\pm$ 5.4 & 38.8 $\pm$ 5.3 & 39.0 $\pm$ 5.5 & 39.5 $\pm$ 5.6 & 39.4 $\pm$ 5.7 & 39.4 $\pm$ 5.6 & 39.3 $\pm$ 5.2 & 40.0 $\pm$ 5.6 & 43.9 $\pm$ 4.6 & \textbf{63.2 $\pm$ 14.9} \\
& \textcolor[HTML]{2eb86c}{$\mathbf{NY_{small}}$} & 16.6 $\pm$ 3.3 & 16.2 $\pm$ 4.1 & 16.3 $\pm$ 4.1 & 16.4 $\pm$ 4.3 & 16.2 $\pm$ 4.1 & 16.3 $\pm$ 4.3 & 16.3 $\pm$ 4.3 & 16.4 $\pm$ 4.3 & 39.3 $\pm$ 5.7 & \textbf{\textcolor[HTML]{2eb86c}{76.9 $\pm$ 24.6}} \\
& NY\textsubscript{all} & 38.8 $\pm$ 5.6 & 40.2 $\pm$ 3.9 & 40.1 $\pm$ 3.9 & 38.7 $\pm$ 5.7 & 38.7 $\pm$ 5.7 & 38.7 $\pm$ 5.7 & 40.1 $\pm$ 4.8 & 35.8 $\pm$ 2.3 & 51.1 $\pm$ 4.1 & \textbf{65.1 $\pm$ 13.9} \\

        \bottomrule
    \end{tabular}
    }
    \caption{This table presents the $\mathbf{PR\text{-}AUC}$ scores for all the experiments divided in two main sections by a double line, to discriminate the results obtained from CNNs (top) and FMs (bottom).
    Each row displays a model's performance across multiple tasks under different fine-tuning (FT) approaches, with the best fine-tuning approach per task highlighted in bold.
    Cells highlighted in \textcolor[HTML]{2eb86c}{green} and \textcolor[HTML]{dead08}{orange} indicate the top and second-best results, respectively, for each dataset task across all model and fine-tuning combinations. }
    \label{tab:results_AUC}

\end{table}

%% file: tables/CCN_vs_FM_stat.tex
\begin{table}[!ht]
\centering
\begin{tabular}{l|cccccc}
\toprule
vs. & LoRA\_r=4 & LoRA\_r=8 & LoRA\_r=16 & BitFit & LP & FFT \\
\midrule
LoRA\_r=4  & --     & 4/6     & 4/6     & 0/6     & 1/6     & 0/6     \\
LoRA\_r=8  & 2/6     & --     & 3/6     & 0/6     & 1/6     & 0/6     \\
LoRA\_r=16 & 2/6     & 3/6     & --     & 1/6     & 1/6     & 0/6     \\
BitFit     & 6/6*    & 6/6*    & 5/6*    & --     & 3/6     & 1/6     \\
LP         & 5/6*    & 5/6*    & 5/6*    & 3/6     & --     & 0/6     \\
FFT        & 6/6*    & 6/6*    & 6/6*    & 5/6*    & 6/6*    & --     \\
\bottomrule
\end{tabular}
\caption{
Pairwise statistical comparison of the fine-tuning techniques applied to both CNN and FM architectures.
Each cell reports the number of datasets (out of six) in which the method on the row achieved a higher \emph{average MCC rank} than the method on the column.
Values are derived from the rank distributions obtained by averaging the performance of all CNN and FM models across datasets.
Asterisks (*) denote statistically significant differences between rank distributions ($p < 0.05$), as determined by the one-tailed Wilcoxon signed-rank test.
}

\label{tab:CNN_vs_FM_stat}
\end{table}

%% file: tables/FMs_stat.tex
\begin{table}[!t]
\centering
\scriptsize
\resizebox{\textwidth}{!}{
\begin{tabular}{lcccccccccc}
\toprule
 & LoRA\_r=4 & LoRA\_r=8 & LoRA\_r=16 & VeRA\_r=4 & VeRA\_r=8 & VeRA\_r=16 & IA3 & BitFit & LP & FFT \\
\midrule
LoRA\_r=4   & --     & 5/6     & 5/6     & 5/6*    & 4/6     & 5/6*    & 6/6*    & 4/6     & 3/6     & 1/6     \\
LoRA\_r=8   & 1/6     & --     & 2/6     & 5/6     & 4/6     & 5/6     & 5/6*    & 1/6     & 1/6     & 1/6     \\
LoRA\_r=16  & 1/6     & 4/6     & --     & 5/6*    & 5/6     & 5/6*    & 6/6*    & 1/6     & 3/6     & 1/6     \\
VeRA\_r=4   & 1/6     & 1/6     & 1/6     & --     & 2/6     & 4/6     & 5/6     & 1/6     & 1/6     & 1/6     \\
VeRA\_r=8   & 2/6     & 2/6     & 1/6     & 4/6     & --     & 6/6*    & 5/6*    & 1/6     & 1/6     & 1/6     \\
VeRA\_r=16  & 1/6     & 1/6     & 1/6     & 2/6     & 0/6     & --     & 4/6     & 0/6     & 0/6     & 1/6     \\
IA3         & 0/6     & 1/6     & 0/6     & 1/6     & 1/6     & 2/6     & --     & 0/6     & 0/6     & 1/6     \\
BitFit      & 2/6     & 5/6*    & 5/6     & 5/6*    & 5/6*    & 6/6*    & 6/6*    & --     & 3/6     & 3/6     \\
LP          & 3/6     & 5/6     & 3/6     & 5/6*    & 5/6*    & 6/6*    & 6/6*    & 3/6     & --     & 1/6     \\
FFT         & 5/6     & 5/6*    & 5/6     & 5/6     & 5/6     & 5/6*    & 5/6*    & 3/6     & 5/6     & --     \\
\bottomrule
\end{tabular}
}
\caption{
Pairwise statistical comparison among foundation model fine-tuning strategies.
Each cell reports the number of datasets (out of six) in which the fine-tuning method on the row achieved a higher \emph{average MCC rank} than the method on the column, computed across all foundation models.
Asterisks (*) indicate statistically significant differences between methods ($p < 0.05$), obtained through the one-tailed Wilcoxon signed-rank test.}

\label{tab:FM_stat}
\end{table}